\documentclass{article}
\def\VenusArxivVersion{1}
\ifdefined\VenusArxivVersion\else
\documentclass{article} % For LaTeX2e
\fi
\newif\ifvenusarxiv
\ifdefined\VenusArxivVersion\venusarxivtrue\else\venusarxivfalse\fi
\ifvenusarxiv
  \newcommand*{\HarnessName}{VenusREM-Harness}
  \newcommand*{\HarnessShortName}{VRH}
  \newcommand*{\ViroBenchmarkName}{VenusViroHub}
  \newcommand*{\ViroBenchmarkShortName}{VVH}
  
  \newcommand*{\GitHubURL}{https://github.com/ai4protein/VenusREM-Harness}
  \newcommand*{\HuggingFaceURL}{https://huggingface.co/datasets/AI4Protein/VenusREM-Harness/}
\else
  \newcommand*{\HarnessName}{REM-Harness}
  \newcommand*{\HarnessShortName}{RH}
  \newcommand*{\ViroBenchmarkName}{ViroHub}
  \newcommand*{\ViroBenchmarkShortName}{VH}
  
  \newcommand*{\GitHubURL}{https://anonymous.4open.science/r/REM-Harness-9C1C}
  \newcommand*{\HuggingFaceURL}{https://huggingface.co/datasets/anonymous-researcher-123/REM-Harness}
\fi
\usepackage{iclr2027_conference,times}

\usepackage{amsmath,amsfonts,bm}

\def\eqref#1{equation~\ref{#1}}
\def\1{\bm{1}}

\DeclareMathAlphabet{\mathsfit}{\encodingdefault}{\sfdefault}{m}{sl}
\SetMathAlphabet{\mathsfit}{bold}{\encodingdefault}{\sfdefault}{bx}{n}

\usepackage{graphicx}
\usepackage{booktabs}
\usepackage{multirow}
\usepackage{array}
\usepackage{amssymb}
\usepackage{wrapfig}
\usepackage{placeins}
\usepackage{float}
\usepackage{flafter}
\usepackage{longtable}
\usepackage{capt-of}
\usepackage{hyperref}
\usepackage{xurl}

\usepackage{xcolor}
\newif\ifreviewchanges
\reviewchangesfalse
\DeclareRobustCommand{\reviewchange}[1]{\ifreviewchanges{\color{red}#1}\else#1\fi}
\usepackage[bottom]{footmisc}
\title{A General Harness for Protein Foundation Model Fitness Prediction}

\ifvenusarxiv
\iclrfinalcopy
\author{% Author order, affiliations, and contribution roles supplied by the authors.
{\normalsize\bfseries
Yang Tan\textsuperscript{1,2,*}\quad
Qijia Tian\textsuperscript{1,*}\quad
Gangyu Sun\textsuperscript{1}\quad
Bozitao Zhong\textsuperscript{1}\\[2pt]
Mingchen Li\textsuperscript{1}\quad
Yuanxi Yu\textsuperscript{1}\quad
Nanqing Dong\textsuperscript{2,3,\ensuremath{\dagger}}\quad
Liang Hong\textsuperscript{1,3,\ensuremath{\dagger}}\par}
\vspace{4pt}
{\small
\textsuperscript{1}Shanghai Jiao Tong University\quad
\textsuperscript{2}Shanghai Innovation Institute\\
\textsuperscript{3}Shanghai Artificial Intelligence Laboratory\par}
\vspace{3pt}
{\footnotesize
\textsuperscript{*}Equal contribution.\quad
\textsuperscript{\ensuremath{\dagger}}Corresponding authors.\\
\texttt{nanqing.dong@sii.edu.cn};\quad\texttt{hongl3liang@sjtu.edu.cn}\par}
}
\hypersetup{pdfauthor={Yang Tan, Qijia Tian, Gangyu Sun, Bozitao Zhong, Mingchen Li, Yuanxi Yu, Nanqing Dong, Liang Hong}}
\makeatletter
\renewcommand{\@maketitle}{%
  \lhead{Preprint}%
  \vspace*{-12pt}%
  \vbox{\hsize\textwidth
    {\LARGE\scshape\@title\par}%
    \vskip 7pt
    {\centering\setlength{\parskip}{0pt}\@author\par}%
    \vskip 7pt}}
\makeatother

\renewenvironment{abstract}{%
  \vskip.075in\centerline{\large\scshape Abstract}\vspace{0.5ex}%
  \begingroup\setlength{\leftmargini}{1.5em}\begin{quote}%
}{\par\end{quote}\endgroup\vskip 1ex}

\else
  \author{Anonymous authors\\Paper under double-blind review}
\fi

\begin{document}

\maketitle

\begin{abstract}
Accurate fitness prediction is central to protein engineering and understanding sequence--function relationships. With advances in deep learning, protein foundation models (PFMs) have become widely used for this task. Recent analyses, however, show that these models share preferences reflecting their training data, while unreliable inputs can further distort fitness predictions. Family-specific evolutionary evidence and structural context provide complementary constraints on these scores. This motivates \textsc{\HarnessName{}} (\textsc{\HarnessShortName{}}), a general, model-agnostic, training-free Retrieval-Enhanced Mutation harness. It combines frozen PFM scores with multiple sequence alignment (MSA) evidence according to model uncertainty, then reduces shared amino-acid preferences and weights scores using structural confidence and solvent exposure. Across $1{,}211$ assays and $3.1$ million measured variants from ProteinGym, VenusMutHub, and \ViroBenchmarkName{}\reviewchange{, a viral benchmark we curate here}, all $71$ configurations improve Spearman correlation on all $3$ benchmarks by $0.073$ on average, with broad gains across $5$ metrics. We also introduce \mbox{\textsc{VenusREM2}}, a \textsc{ProSST} ensemble built with the harness. \reviewchange{Against official leaderboard baselines, it leads all function, taxon, MSA-depth, and mutation-depth categories, reaching ProteinGym Average Spearman $0.556$ ($+0.038$).} \reviewchange{Extended analyses compare PFM and MSA preferences, assess gains across protein domains and immune-escape cases, and measure computational speedups.}
\end{abstract}

\section{Introduction}

\begin{wrapfigure}{r}{0.49\textwidth}
\centering
\includegraphics[width=\linewidth]{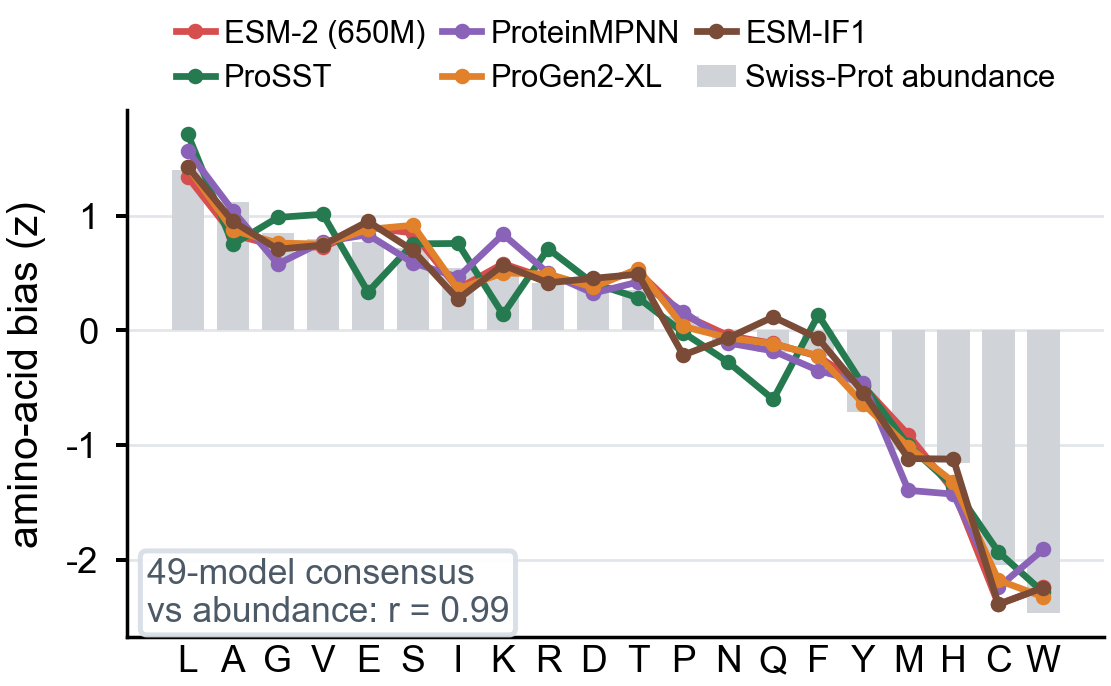}
\caption{Shared amino-acid preferences across representative PFMs.}
\label{fig:teaser}
\end{wrapfigure}

Single amino acid substitutions can alter stability, catalysis, recognition, and cellular fitness, making prediction central to variant interpretation and protein engineering \citep{notin2024proteingym,zhou2024mlife}. Deep mutational scanning (DMS) measures these effects but covers few natural proteins and variants. Computational models rank large numbers of mutations from sequence or structure without assay labels \citep{meier2021esm1v,frazer2021eve,lin2023esm2}.

\leavevmode{}Among these approaches, protein foundation models (PFMs) learn general patterns from large protein sequence collections \citep{rives2021esm1b}, but their amino-acid preferences also reflect the composition of the training data (Figure~\ref{fig:teaser}). \reviewchange{These preferences can affect how accurately PFMs predict mutation effects} \citep{gordon2025preference}. \reviewchange{Predicting a specific mutation requires knowing which amino acids are tolerated at that position in the protein. Homologous sequences show which amino acids have been retained at corresponding positions during evolution} \citep{hopf2017evmutation}. \reviewchange{A multiple sequence alignment (MSA) summarizes these observations as amino-acid frequencies at each position. We find that these frequencies often differ from average PFM preferences and UniProt amino-acid frequencies} (Appendix~\ref{app:profile_separation}). \textsc{Tranception} uses this family information to improve frozen PFM predictions \citep{notin2022tranception}. \reviewchange{Similarly, \textsc{VenusREM} combines MSA amino-acid frequencies with scores from \textsc{ProSST}, a sequence--structure Transformer with disentangled attention} \citep{li2024prosst,tan2025venusrem}.

Homologous sequences describe evolutionary tolerance, while a substitution's physical effects also depend on its structural environment \citep{zhou2024protlgn}. Mutations in buried cores are often more destabilizing than surface substitutions, whereas exposed surfaces participate in molecular recognition \citep{zhang2024s3f,beltran2025domainome}. Structural information can therefore help prediction, provided it is reliable \citep{tan2024sesadapter}. Predicted structures vary in accuracy across residues, and a structure may not represent the binding partners or conditions of an assay \citep{jumper2021alphafold2,terwilliger2024alphafold}. \textsc{RSALOR} adjusts conservation scores by solvent accessibility, while \textsc{S3F} uses sequence scores where structural confidence is low \citep{zhang2024s3f,tsishyn2025rsalor}.

\reviewchange{Together, these observations suggest refining general PFM preferences with family and structural information about the target protein.} \textit{How can a general harness integrate evolutionary and structural evidence to reduce model bias and account for uncertainty in protein fitness prediction?}

Inspired by large language model (LLM) harnesses that coordinate inference for reliable deployment \citep{deepseek-harness2026}, we introduce the training-free Retrieval-Enhanced Mutation harness (\textsc{\HarnessName{}}; \textsc{\HarnessShortName{}}), \textbf{a unified and efficient inference layer} around frozen PFMs for fitness prediction. \reviewchange{Our contributions are as follows:}
\begin{itemize}
\item \reviewchange{\textbf{\HarnessName{}.} We combine adaptive MSA fusion, gated background correction, and structural weighting in a framework for frozen PFMs. Across ProteinGym \citep{notin2024proteingym}, VenusMutHub \citep{zhang2025venusmuthub}, and \ViroBenchmarkName{} ($1{,}211$ assays and $3.1$ million measured variants), all $71$ configurations improve Spearman correlation, with an average gain of $0.073$.}
\item \reviewchange{\textbf{\ViroBenchmarkName{}.} We provide a new viral protein dataset for community benchmarking and studies of immune escape and other viral phenotypes. It contains $616{,}690$ substitutions from $89$ assays, covering $18$ virus types, $52$ wild-type (WT) sequences, and seven phenotypes.}
\item \reviewchange{\textbf{VenusREM2.} Building on \textsc{VenusREM}, we apply the harness to six checkpoints of \textsc{ProSST}. Their ensemble sets a new state of the art in the official ProteinGym comparison: \textbf{Average Spearman $\mathbf{0.556}$} ($+0.038$). Replacing \textsc{VenusREM}'s loops with array operations makes MSA counting and mutation scoring $62.3\times$ and $20.7\times$ faster, respectively, after model inference.}
\end{itemize}

\begin{figure}[!t]
\centering
\includegraphics[width=\textwidth]{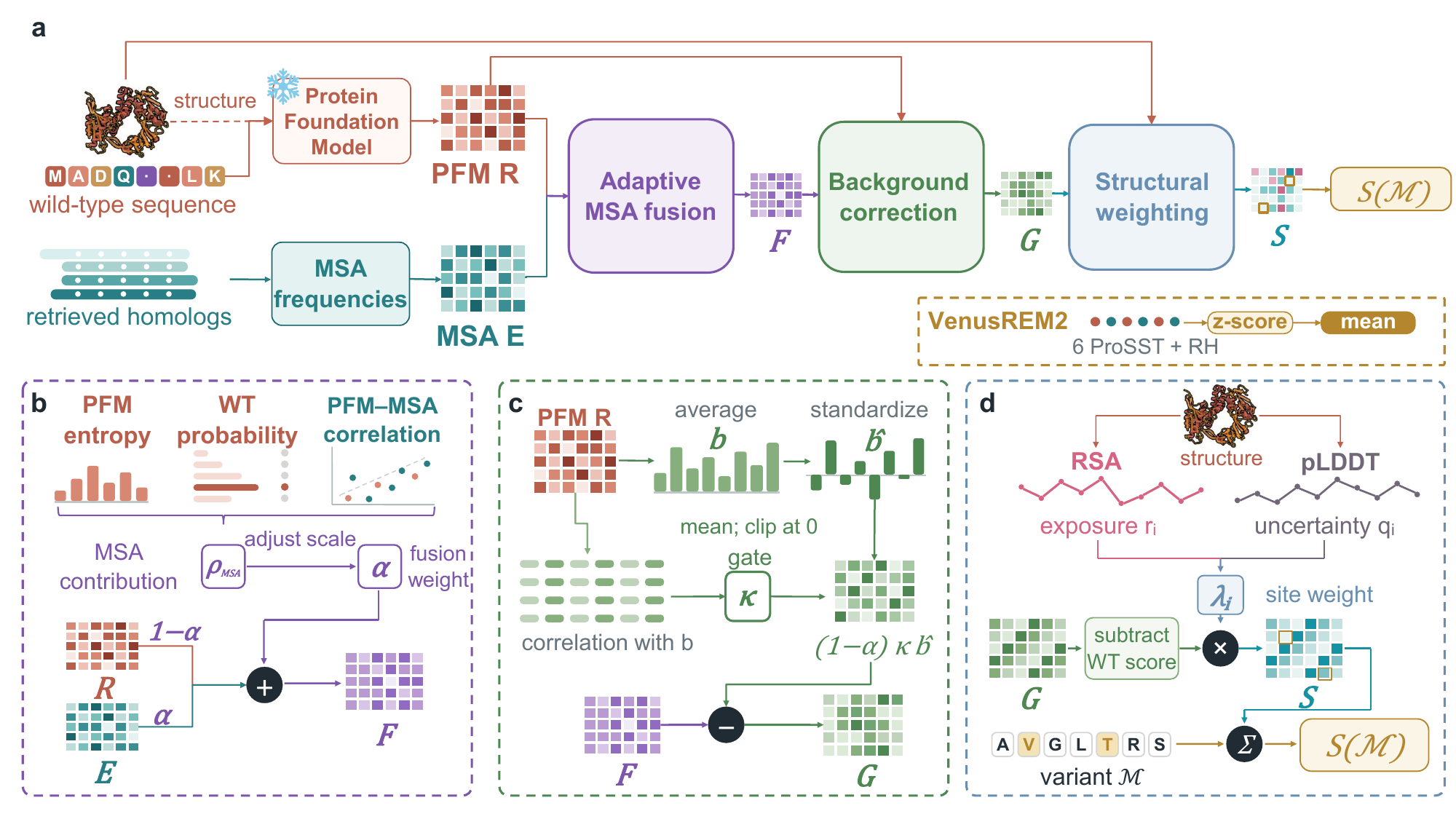}
\caption{Framework of \textsc{\HarnessName{}}. \textbf{(a)} \reviewchange{The harness adjusts frozen} PFM scores using \reviewchange{evolutionary information from MSAs} and structural \reviewchange{information}. \textbf{(b)} Adaptive MSA fusion \reviewchange{sets} the \reviewchange{MSA weight for each protein} and model. \textbf{(c)} \reviewchange{Gated background correction reduces shared} model \reviewchange{preferences} while preserving \reviewchange{the MSA contribution}. \textbf{(d)} \reviewchange{Structural confidence and solvent} exposure \reviewchange{adjust substitution scores, which are then summed for} each variant.}
\label{fig:framework}
\end{figure}

\section{REM-Harness: A General Framework for Fitness Prediction}
\label{sec:method}

\reviewchange{Evolutionary constraints, learned PFM preferences, and structural reliability motivate three components in \textsc{\HarnessName{}} (Figure~\ref{fig:framework}). MSA fusion adds family-specific information to PFM scores. Background correction reduces amino-acid preferences shared across positions in the PFM contribution. Structural weighting adjusts the resulting mutation effects using solvent exposure and confidence in predicted structures.}

\subsection{\reviewchange{PFM and MSA scores}}
\label{sec:interface}

\reviewchange{For the PFM scores, we use the WT sequence $x=(x_1,\ldots,x_L)$ of length $L$, with its structure when required. Position $i$ contains WT residue $x_i$, and $a$ indexes the 20 standard amino acids. We store PFM log probabilities or logits in $R\in\mathbb{R}^{L\times20}$, with row $R_{i,\cdot}$ containing the 20 scores at position $i$.}

\reviewchange{We obtain amino-acid probabilities in three ways: \textbf{(i) Masked scoring} replaces each WT residue $x_i$ in turn with a mask token to form $x_{\setminus i}$ and predicts amino-acid probabilities at that position. \textbf{(ii) WT scoring} predicts probabilities at every position from the full unmasked WT sequence in one pass. \textbf{(iii) Teacher-forced scoring (tf)} predicts them from preceding WT residues $x_{<i}$ in the decoding order. For each input, let $z_{\theta,i}$ be the logits from the frozen PFM with parameters $\theta$. Softmax gives probabilities $P_{\theta,i}$ over its vocabulary. The log-probability scores are:}
\begin{equation}
\begin{gathered}
\reviewchange{P_{\theta,i}=\operatorname{softmax}(z_{\theta,i}),}\\
\begin{aligned}
R^{\mathrm{mask}}_{i,a}&=\log \reviewchange{P_{\theta,i}}(a;x_{\setminus i}\reviewchange{,\mathcal C}), &
R^{\mathrm{wt}}_{i,a}&=\log \reviewchange{P_{\theta,i}}(a;x\reviewchange{,\mathcal C}), &
R^{\mathrm{tf}}_{i,a}&=\log \reviewchange{P_{\theta,i}}(a;x_{<i}\reviewchange{,\mathcal C}).
\end{aligned}
\end{gathered}
\label{eq:readouts}
\end{equation}
\reviewchange{Here, $\mathcal C$ denotes optional structural inputs, omitted when not required. Appendix~\ref{app:baselines}, Table~\ref{tab:baselines}, specifies the scoring interface for each configuration.}

\reviewchange{For the MSA scores, we retrieve homologous sequences and align them to the WT \reviewchange{using MMseqs2 via ColabFold \citep{steinegger2017mmseqs2,mirdita2022colabfold}}. For an MSA covering all $L$ WT positions, let $N$ be the number of aligned sequences and $n_{i,a}$ the number with amino acid $a$ at position $i$. We convert these counts to frequencies $f_{i,a}$ and apply log-softmax to form the MSA score matrix $E\in\mathbb{R}^{L\times20}$:}
\begin{equation}
\reviewchange{f_{i,a}=\frac{n_{i,a}}{N},\qquad
E_{i,a}=\bigl[\log\operatorname{softmax}(f_{i,\cdot})\bigr]_a.}
\label{eq:msa_frequency}
\end{equation}
\reviewchange{Appendix~\ref{app:score_construction} details PFM outputs, gap counting, and normalization.}

\subsection{\reviewchange{Adaptive MSA fusion}}
\label{sec:fusion}

\reviewchange{Recent PFM-based methods combine PFM and MSA scores \citep{notin2022tranception,tan2025venusrem}. Using $R$ and $E$ from Equations~\ref{eq:readouts} and~\ref{eq:msa_frequency} and MSA weight $\alpha$, the fused matrix is:}

\begin{equation}
\reviewchange{F=(1-\alpha)R+\alpha E.}
\label{eq:fusion}
\end{equation}

\reviewchange{\textsc{VenusREM} fixes $\alpha=0.8$ using validation data. We instead calculate $\alpha$ for each protein and PFM. We hypothesize that MSA information helps more when the PFM assigns low probabilities to WT residues or favors different amino acids from the MSA. We combine these criteria using entropy, then adjust for the score ranges.}

\reviewchange{\textbf{(i) WT probabilities.} We first measure how much probability the PFM assigns to residues already present in the WT. Softmax normalizes each row of $R$ over the 20 amino acids, giving probabilities $p_i(a)$ and WT probability $p_i(x_i)$. Lower WT probabilities increase $\rho_{\mathrm{WT}}$:}

\begin{equation}
\reviewchange{p_i(a)=\bigl[\operatorname{softmax}(R_{i,\cdot})\bigr]_a},
\qquad
\reviewchange{\rho_{\mathrm{WT}}=\operatorname{clip}\!\left(
-\frac{1}{L}\sum_{i=1}^{L}\log_{20}p_i(x_i),0,1\right).}
\label{eq:wt_weight}
\end{equation}
\reviewchange{Base-20 logarithms give one for uniform probabilities. Clipping keeps $\rho_{\mathrm{WT}}$ in $[0,1]$.}

\reviewchange{\textbf{(ii) PFM--MSA correlation.} \reviewchange{Lower correlation indicates greater disagreement between learned PFM preferences and family-specific evolutionary preferences, so we give the MSA more influence.} We calculate the Pearson correlation between their 20 scores at each position, then average to obtain $\bar r_{R,E}$ \reviewchange{ and define $\rho_{\mathrm{corr}}\in[0,1]$:}}

\begin{equation}
\reviewchange{\bar r_{R,E}
=\frac{1}{L}\sum_{i=1}^{L}\operatorname{corr}(R_{i,\cdot},E_{i,\cdot}),
\qquad
\rho_{\mathrm{corr}}=\frac{1-\bar r_{R,E}}{2}.}
\label{eq:correlation_weight}
\end{equation}

\reviewchange{\textbf{(iii) Entropy weighting.} A low WT probability can mean that the PFM strongly favors other residues or assigns similar probabilities to many residues. Entropy helps distinguish these cases by measuring how spread out all 20 probabilities are.}

\reviewchange{We average normalized entropy across positions to obtain $\bar H\in[0,1]$. At low entropy, we give more weight to the WT criterion $\rho_{\mathrm{WT}}$ (Equation~\ref{eq:wt_weight}). At high entropy, low WT probabilities are expected, so we give more weight to the PFM--MSA comparison $\rho_{\mathrm{corr}}$ (Equation~\ref{eq:correlation_weight}). We combine these weights linearly to obtain $\rho_{\mathrm{MSA}}$:}

\begin{equation}
\reviewchange{\bar H=-\frac{1}{L}\sum_{i=1}^{L}\sum_a p_i(a)\log_{20}p_i(a),}
\qquad
\rho_{\mathrm{MSA}}=(1-\bar H)\reviewchange{\rho_{\mathrm{WT}}}+\bar H\reviewchange{\rho_{\mathrm{corr}}}.
\label{eq:pref_map}
\end{equation}

\reviewchange{\textbf{(iv) Scale adjustment.} The same mixing weight gives more influence to scores that vary more. We therefore adjust $\rho_{\mathrm{MSA}}$ from Equation~\ref{eq:pref_map} for the variation in PFM and MSA scores.}

\reviewchange{Mutation scores compare each replacement with the WT residue. Across all possible substitutions, the sample standard deviation of these score differences is $s_R$ for the PFM and $s_E$ for the MSA. We divide $1-\rho_{\mathrm{MSA}}$ by $s_R$ and $\rho_{\mathrm{MSA}}$ by $s_E$, then normalize the two weights to sum to one. For $s_R,s_E>0$, this gives:}
\begin{equation}
\reviewchange{\alpha
=\frac{\rho_{\mathrm{MSA}}/s_E}
{(1-\rho_{\mathrm{MSA}})/s_R+\rho_{\mathrm{MSA}}/s_E}.}
\label{eq:native_alpha}
\end{equation}

\reviewchange{Using $\alpha$ from Equation~\ref{eq:native_alpha} in Equation~\ref{eq:fusion} gives the fused score matrix $F$. Section~\ref{sec:results_scales} compares this adaptive fusion with fixed coefficients, and Appendix~\ref{app:scale_derivation} gives the derivation.}

\subsection{\reviewchange{Model background correction}}
\label{sec:bias}
\label{sec:calibration}

\reviewchange{MSA fusion adds family information but retains shared PFM preferences, which can reflect training-data composition (Figure~\ref{fig:teaser}; \citealp{gordon2025preference}). We subtract this background from the PFM contribution to emphasize preferences at individual sites.}

\reviewchange{For log-probability scores, $b_a$ is the logarithm of the mean probability assigned to amino acid $a$ across all $L$ positions. The 20 values form the background $\mathbf b$. We obtain standardized values $\widehat b_a$ using their mean $\bar b$ and sample standard deviation $s_b$:}
\begin{equation}
b_a=\log\!\left(\frac{1}{L}\sum_{i=1}^{L}e^{R_{i,a}}\right),
\qquad
\widehat b_a=\frac{b_a-\bar b}{s_b}.
\label{eq:background}
\end{equation}

\reviewchange{We apply stronger correction when positions repeatedly favor the same amino acids as the background. We measure this using the mean Pearson correlation $c$ between each PFM row $R_{i,\cdot}$ and $\mathbf b$ from Equation~\ref{eq:background}. The gate $\kappa(R)$ equals $c$ when positive and zero otherwise:}
\begin{equation}
c=\frac{1}{L}\sum_{i=1}^{L}\reviewchange{\operatorname{corr}(R_{i,\cdot},\mathbf b)},
\qquad
\kappa(R)=\reviewchange{\max(0,c)}.
\label{eq:gate}
\end{equation}
\reviewchange{We multiply the standardized background by the gate in Equation~\ref{eq:gate} and the PFM weight $1-\alpha$. Subtracting this correction from $F$ gives $G$ (Appendix~\ref{app:calibration_identities}):}

\begin{equation}
\reviewchange{G_{i,a}=F_{i,a}-(1-\alpha)\kappa(R)\widehat b_a.}
\label{eq:corrected_field}
\end{equation}

\begin{figure}[!t]
\centering
\includegraphics[width=\textwidth]{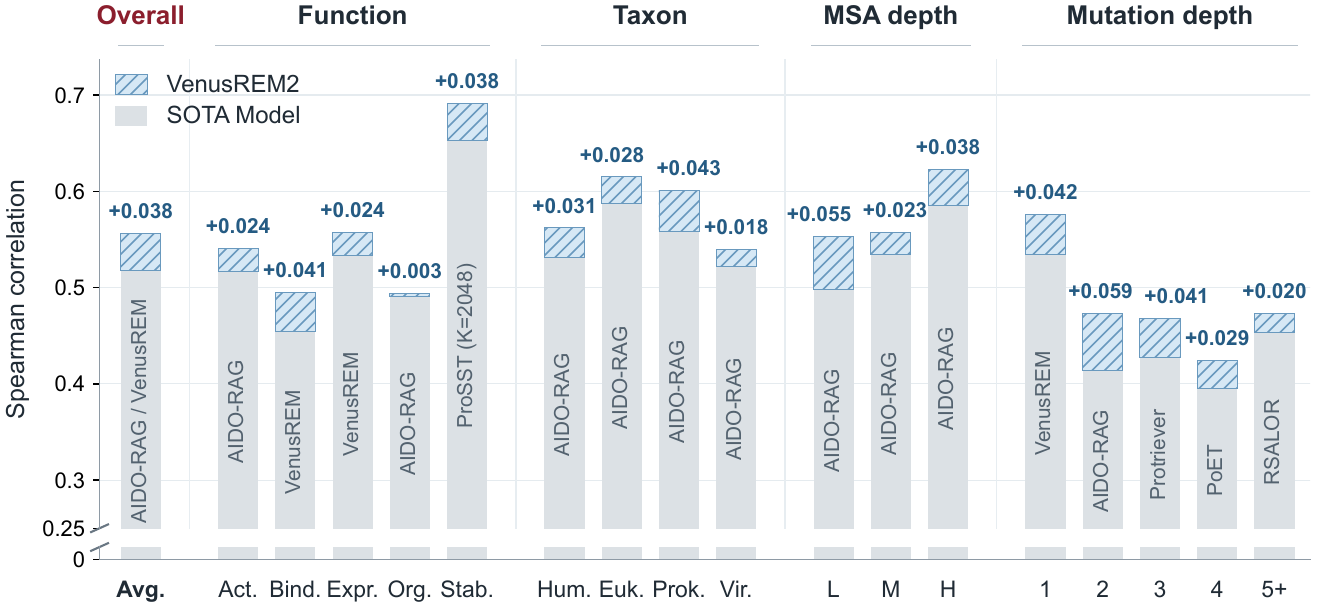}
\caption{\reviewchange{Spearman comparison with official ProteinGym baselines}, overall and across 17 categories (Appendix Tables~\ref{tab:leaderboard}--\ref{tab:leaderboard_full_mutation_depth}). Gray: strongest \reviewchange{comparison model} (SOTA Model); blue hatching and labels: \textsc{VenusREM2} gains. Vertical labels identify \reviewchange{the comparison models}. Avg.: overall; Act./Bind./Expr./Org./Stab.: activity/binding/expression/organismal fitness/stability; Hum./Euk./Prok./Vir.: human/other eukaryotes/prokaryotes/viruses; L/M/H: low/medium/high MSA depth. Mutation depth counts substitutions per variant ($5+$: at least five).}
\label{fig:leaderboard}
\end{figure}

\subsection{\reviewchange{Structural weighting and variant scores}}
\label{sec:shrinkage}
\label{sec:ranking}

\reviewchange{After fusion and background correction, we use two measurements from the WT structure predicted by AlphaFold2 \citep{jumper2021alphafold2}: \textbf{(i) pLDDT} (predicted local distance difference test score) estimates confidence in the local structure. Low pLDDT indicates less reliable structural information \citep{zhang2024s3f}. \textbf{(ii) RSA} (relative solvent accessibility) measures residue exposure to solvent. Exposed residues often tolerate substitutions better than buried residues \citep{tsishyn2025rsalor}.}

\reviewchange{We represent greater mutational tolerance by smaller predicted changes from WT. We also reduce effect estimates when structural confidence is low. Both adjustments shrink scores at sites with above-average exposure or uncertainty. Let $r_i$ be RSA clipped to $[0,1]$ and $q_i=1-\mathrm{pLDDT}_i/100$ measure uncertainty. With protein averages $\bar r$ and $\bar q$, we define the two weights and their product $\lambda_i$:}
\begin{equation}
\begin{gathered}
\reviewchange{w_i^{\mathrm{RSA}}=1-\max\{0,r_i-\bar r\},\qquad
w_i^{\mathrm{pLDDT}}=1-\max\{0,q_i-\bar q\},}\\
\reviewchange{\lambda_i=w_i^{\mathrm{RSA}}\,w_i^{\mathrm{pLDDT}}.}
\end{gathered}
\label{eq:lambda}
\end{equation}
\reviewchange{Using $G$ from Equation~\ref{eq:corrected_field}, we subtract the WT score from the replacement score. We multiply this difference by $\lambda_i$ (Equation~\ref{eq:lambda}) to obtain $S_{i,a}$ for substitution $x_i\to a$. A variant's substitution set $\mathcal M$ contains pairs $(i,a)$ with $a\ne x_i$. We sum their scores:}
\begin{equation}
S(\mathcal M)=\sum_{(i,a)\in\mathcal M}S_{i,a}
\reviewchange{=\sum_{(i,a)\in\mathcal M}\lambda_i\bigl(G_{i,a}-G_{i,x_i}\bigr).}
\label{eq:aggregate}
\end{equation}
\reviewchange{Appendix~\ref{app:structure_extension} details the averages and handling of missing structural data.}

\begin{table}[!t]
\caption{\reviewchange{Scope and scale of ProteinGym \citep{notin2024proteingym}, VenusMutHub \citep{zhang2025venusmuthub}, and \ViroBenchmarkName{}. Escape/entry denotes separate immune-escape and cell-entry groups.}}
\label{tab:benchmark_comparison}
\centering
\small
\setlength{\tabcolsep}{4pt}
\reviewchange{%
\begin{tabular}{@{}>{\raggedright\arraybackslash}p{.32\textwidth}>{\centering\arraybackslash}p{.18\textwidth}>{\centering\arraybackslash}p{.18\textwidth}>{\centering\arraybackslash}p{\dimexpr.32\textwidth-6\tabcolsep\relax}@{}}
\toprule
\textbf{Feature} & \textbf{ProteinGym} & \textbf{VenusMutHub} & \textbf{\ViroBenchmarkName{} (ours)} \\
\midrule
Data scope & Broad DMS & \shortstack{Biochemical\\measurements} & \shortstack{\textbf{Viral DMS}\\(\textbf{18 virus types})} \\
Assays / WT sequences & 217 / 187 & 905 / 527 & 89 / 52 \\
Substitution records & 2,465,767 & 27,594 & 616,690 \\
Phenotype groups & 5 & 4 & 7 \\
\midrule
Separate escape / entry groups & Not separate & Not separate & $\checkmark$ (\textbf{21 / 18 assays}) \\
\bottomrule
\end{tabular}%
}
\end{table}

\subsection{\reviewchange{VenusREM2: ProSST ensemble}}
\label{sec:venusrem2}

\reviewchange{Combining models from the same family is a common way to improve fitness prediction, as in \textsc{ESM-1v} and \textsc{ProtSSN} \citep{meier2021esm1v,tan2025protssn}. Building on \textsc{VenusREM}, \textsc{VenusREM2} combines six \textsc{ProSST} checkpoints for their strong mutation prediction performance. They differ in the number $k$ of structure-token types used to encode local 3D environments, with $k\in\mathcal K=\{20,128,512,1024,2048,4096\}$ (Appendix~\ref{app:score_construction}).}

\reviewchange{Each checkpoint uses WT scoring (Equation~\ref{eq:readouts}) and its own fusion weight $\alpha_k$ and background correction. The MSA, RSA, and pLDDT are shared. For variant $v$ with substitution set $\mathcal M$, Equation~\ref{eq:aggregate} gives checkpoint $k$'s score $S_k(v)$.}

\reviewchange{The checkpoints have different score ranges. We standardize their predictions within assay $A$ using each one's mean $\mu_{k,A}$ and population standard deviation $\sigma_{k,A}$, then average:}
\begin{equation}
S_{\mathrm{VenusREM2}}(v;A)
=\frac{1}{6}\reviewchange{\sum_{k\in\mathcal K}}
\frac{S_k(v)-\mu_{k,A}}{\sigma_{k,A}}.
\label{eq:prosst_ensemble}
\end{equation}
\reviewchange{Standardization uses no measured fitness values, but changing the variant set can change the ensemble ranking (Appendix~\ref{app:ensemble_definition}).}

\section{Results}
\label{sec:benchmarks}

\subsection{Experimental setup}
\label{sec:setup}

\paragraph{Benchmarks and paired evaluation.}
We evaluate ProteinGym ($217$ substitution assays), VenusMutHub ($905$ assays; $527$ WT clusters), and \ViroBenchmarkName{} ($89$ assays). \reviewchange{Spearman correlation is the primary metric. Normalized discounted cumulative gain (NDCG) and top-variant recall measure ranking quality. Area under the receiver operating characteristic curve (AUC) and Matthews correlation coefficient (MCC) measure binary classification performance.} We use ProteinGym's official Average Spearman and VenusMutHub's mean over assays. For \ViroBenchmarkName{}, we first average assays for each protein and phenotype, then proteins within each phenotype, and finally the seven phenotypes. Each configuration specifies a checkpoint and scoring method. The $71$ configurations span four model families (Appendix~\ref{app:baselines}). For each configuration, we compare performance before and after applying the harness, keeping model parameters, inputs, and scoring method fixed. Raw denotes original predictions; \textsc{\HarnessShortName{}} denotes harness predictions.

\subsection{\reviewchange{\ViroBenchmarkName{}: a community dataset for viral protein research}}
\label{sec:virohub}

\paragraph{\reviewchange{Scope and research use.}}
\reviewchange{\ViroBenchmarkName{} provides a dedicated benchmark for viral protein mutation effects. It complements ProteinGym's broad DMS coverage and VenusMutHub's small-scale biochemical measurements \citep{notin2024proteingym,zhang2025venusmuthub} (Table~\ref{tab:benchmark_comparison}). The collection contains $616{,}690$ single-amino-acid substitutions from $89$ assays across $18$ virus types and $52$ WT sequences. Its seven phenotype groups include $21$ immune-escape and $18$ cell-entry assays, compared with $14$ and $11$ in ViroGym \citep{zhou2026virogym}. This coverage supports comparisons across viral function, immune escape, and host-cell entry, including antibody-related case studies (Section~\ref{sec:immune_escape}).}

\paragraph{\reviewchange{Data curation.}}
\reviewchange{We reconstruct each assay from published mutant-level measurements and its WT sequence. Residue numbering is mapped to the experimental construct, and score directions are standardized. We remove missing scores, unmappable mutations, WT-residue mismatches, and duplicate measurements within the collection or shared with ProteinGym or VenusMutHub. WT controls are retained in the data files but excluded from every evaluation metric. We release WT sequences, sources, processing records, and evaluation code to support reuse. Appendix~\ref{app:virogym_data_processing} documents these steps, and Appendix~\ref{app:metric_defs} specifies aggregation and uncertainty estimation.}

\subsection{Consistent gains across architectures and benchmarks}
\label{sec:transfer}

\begin{figure}[!t]
\centering
\includegraphics[width=\textwidth]{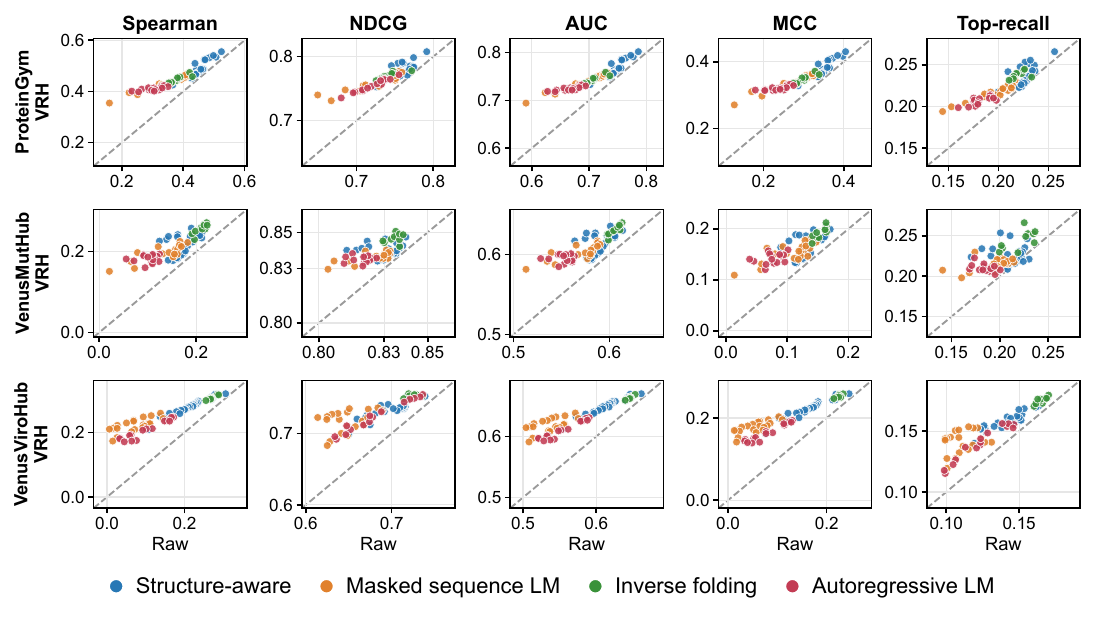}
\caption{Performance before and after applying \textsc{\HarnessName{}} across five metrics on ProteinGym, VenusMutHub, and \ViroBenchmarkName{}. Raw denotes original predictions; \textsc{\HarnessShortName{}} denotes harness predictions. Each point represents one of 71 configurations.}
\label{fig:scatter}
\end{figure}

\paragraph{Gains across architectures and evaluation metrics.}
\textsc{\HarnessName{}} improves Spearman for all $71$ configurations on all three benchmarks, with gains also in classification and identification of top variants. Mean Spearman increases by $0.068$ on ProteinGym, $0.056$ on VenusMutHub, and $0.094$ on \ViroBenchmarkName{}, under their respective aggregation protocols. Across all five metrics, performance improves in all $355$ comparisons on ProteinGym and \ViroBenchmarkName{}, while $353/355$ show gains or ties at reported precision on VenusMutHub (Figure~\ref{fig:scatter}; Appendix~\ref{app:per_config_metrics}). Gains are largest for autoregressive models on ProteinGym and VenusMutHub and masked sequence models on \ViroBenchmarkName{}, while structure-aware and inverse-folding models also benefit (Appendix~\ref{app:family_results}).

\paragraph{State-of-the-art performance on ProteinGym.}
\reviewchange{\textsc{VenusREM2} achieves a new state of the art in the official ProteinGym comparison: Average Spearman $0.556$, up $0.038$. It leads all $17$ function, taxon, MSA-depth, and mutation-depth categories against these baselines} (Figure~\ref{fig:leaderboard}; Appendix Tables~\ref{tab:leaderboard}--\ref{tab:leaderboard_full_mutation_depth}). \reviewchange{Adding the published \textsc{PoET-2} results leaves \textsc{VenusREM2} first overall and in 15 of 17 categories (Appendix Table~\ref{tab:poet2_comparison}).} Against our matched six-model ensemble without the harness, the gain is $0.031$, separating the harness benefit from model averaging. All $71$ configurations also improve in every ProteinGym functional category ($355/355$ comparisons). Gains occur in all MSA-depth groups and are larger for families with fewer aligned sequences (Appendix~\ref{app:stratified}).

\begin{table}[!t]
\caption{\textsc{ProSST}-2048 ablation. \reviewchange{PG validation (ProteinGym) and VMH (VenusMutHub) use mean assay Spearman. Full PG uses official Average Spearman.} \ViroBenchmarkShortName{} (\ViroBenchmarkName{}): \reviewchange{Spearman for mutants, averaged over proteins within phenotypes and then phenotypes}. Raw denotes original PFM scores; $\Delta_{\mathrm{Raw}}$ is the change from this baseline (three decimals). Fixed MSA fusion uses $\alpha=0.8$; adaptive fusion uses Equation~\ref{eq:native_alpha}. Ungated correction fixes $\kappa=1$; gated correction uses $\kappa(R)$. \reviewchange{Paired component statistics are reported in Appendix~\ref{app:prosst_component_statistics}.}}
\label{tab:ablation}
\centering
\scriptsize
\setlength{\tabcolsep}{1.8pt}
\resizebox{\textwidth}{!}{%
\begin{tabular}{@{}llcccccccc@{}}
\toprule
\multirow[c]{2}{*}{Component} & \multirow[c]{2}{*}{Setting} & \multicolumn{2}{c}{PG Validation (21)} & \multicolumn{2}{c}{PG Full (217)} & \multicolumn{2}{c}{VMH Full (905)} & \multicolumn{2}{c}{\ViroBenchmarkShortName{} (89)} \\
\cmidrule(lr){3-4}\cmidrule(lr){5-6}\cmidrule(lr){7-8}\cmidrule(l){9-10}
& & Score & $\Delta_{\mathrm{Raw}}$ & Score & $\Delta_{\mathrm{Raw}}$ & Score & $\Delta_{\mathrm{Raw}}$ & Score & $\Delta_{\mathrm{Raw}}$ \\
\midrule
Raw & PFM scores & 0.5573 & --- & 0.5069 & --- & 0.2126 & --- & 0.2709 & --- \\
\midrule
\multirow[c]{2}{*}{MSA fusion} & Fixed weight & 0.5742 & +0.017 & 0.5205 & +0.014 & \textbf{0.2341} & \textbf{+0.022} & 0.2897 & +0.019 \\
 & Adaptive weight & \textbf{0.5774} & \textbf{+0.020} & \textbf{0.5246} & \textbf{+0.018} & 0.2336 & +0.021 & \textbf{0.2901} & \textbf{+0.019} \\
\midrule
\multirow[c]{2}{*}{Model correction} & Without gate & 0.5641 & +0.007 & 0.5219 & +0.015 & 0.2326 & +0.020 & 0.2841 & +0.013 \\
 & With gate & \textbf{0.5788} & \textbf{+0.022} & \textbf{0.5284} & \textbf{+0.022} & \textbf{0.2369} & \textbf{+0.024} & \textbf{0.2942} & \textbf{+0.023} \\
\midrule
\multirow[c]{2}{*}{Structural weighting} & RSA only & 0.5826 & +0.025 & \textbf{0.5329} & \textbf{+0.026} & \textbf{0.2499} & \textbf{+0.037} & \textbf{0.3085} & \textbf{+0.038} \\
 & pLDDT only & \textbf{0.5833} & \textbf{+0.026} & \textbf{0.5329} & \textbf{+0.026} & 0.2352 & +0.023 & 0.2994 & +0.028 \\
\midrule
\textsc{\HarnessName{}} & Full method & \textbf{0.5849} & \textbf{+0.028} & \textbf{0.5343} & \textbf{+0.027} & 0.2491 & +0.037 & \textbf{0.3102} & \textbf{+0.039} \\
\bottomrule
\end{tabular}%
}
\end{table}

\subsection{Complementarity between PFM and MSA scores}
\label{sec:results_scales}

\paragraph{Complementarity and fusion gains.}
To test when MSA information helps, we analyze $148$ ProteinGym proteins scored with unmasked \textsc{ESM-2}. At each site, we compare the 20 MSA amino-acid frequencies with two references: average PFM preferences across the protein and log amino-acid frequencies in UniProt. We count the fraction of sites with Pearson $r<0.3$ for at least one reference. Proteins with a larger fraction gain more from MSA fusion (Spearman $\rho=0.338$; Appendix Figure~\ref{fig:separation_gain}). Proteins in the highest quartile gain $0.058$ more than the lowest. This association persists after adjusting for sequence length, MSA depth, the fraction of nongap MSA entries, and original model performance (Appendix~\ref{app:profile_separation}). This supports complementing average PFM preferences with position-specific information from homologous sequences.

\paragraph{Adaptive fusion.}
On the $21$-assay \textsc{ProSST}-2048 validation split, MSA scores alone reach Spearman $0.4093$, the original model $0.5573$, and adaptive MSA fusion $0.5774$. Adaptive fusion matches or exceeds performance with fixed $\alpha\in[0.6,0.9]$, and this pattern holds when averaging over all ten folds (Appendix~\ref{app:alpha_sweep}). \reviewchange{Appendix~\ref{app:saprot_transfer} repeats the component comparisons with \textsc{SaProt}-650M-AF2.}

\subsection{Contributions of the three components}
\label{sec:components}

\paragraph{Gains beyond MSA fusion.}
On the $21$ validation assays, MSA fusion supplies $0.0201$ of the $0.0276$ \textsc{ProSST}-2048 gain, with background correction and structural weighting adding the remaining $0.0075$ (Table~\ref{tab:ablation}). \reviewchange{Each stage also improves performance on the remaining $196$ ProteinGym assays, raising official Average Spearman from $0.5072$ to $0.5337$.} Across all $71$ configurations, MSA fusion accounts for approximately $50\%$, $55\%$, and $36\%$ of gains on ProteinGym, VenusMutHub, and \ViroBenchmarkName{}, respectively (Appendix~\ref{app:staged_audits}). \reviewchange{Paired component comparisons, confidence intervals (CIs), and $p$-values are reported in Appendix~\ref{app:prosst_component_statistics}.}

\paragraph{Gating improves background correction.}
On ProteinGym, ungated correction improves on MSA fusion in $43/71$ configurations, versus $71/71$ with gating, with mean gains of $0.003$ and $0.012$, respectively (Appendix Table~\ref{tab:gate_ablation}). Gating also exceeds ungated correction in $37/71$ VenusMutHub and $68/71$ \ViroBenchmarkName{} configurations. \reviewchange{Excluding the scored site from background estimation changes the gate little} (Appendix Figure~\ref{fig:coherence_mechanism}; Appendix~\ref{app:staged_audits}).

\paragraph{Exposure and confidence provide complementary context.}
After background correction, combining RSA and pLDDT gives the strongest \textsc{ProSST}-2048 Spearman on the full $217$-assay ProteinGym benchmark ($0.5343$) and \ViroBenchmarkName{} ($0.3102$; Table~\ref{tab:ablation}). Exposure accounts for most of the structural gain on VenusMutHub, where RSA alone reaches $0.2499$ and the combination $0.2491$.

\subsection{Gains across viral phenotypes}
\label{sec:immune_escape}

\begin{figure}[!t]
\centering
\includegraphics[width=\textwidth]{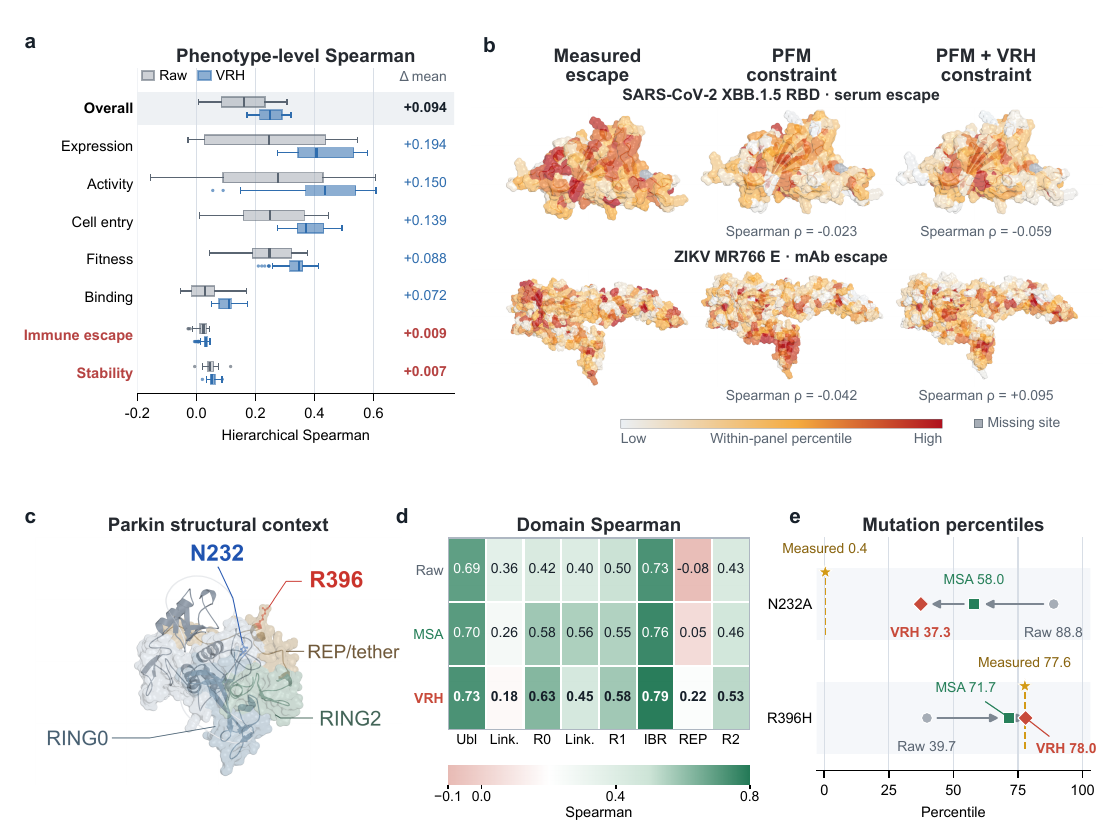}
\caption{\textbf{(a)} Spearman distributions before and after applying the harness, grouped by phenotype across $71$ configurations (medians, interquartile ranges (IQRs), $1.5\times$ IQR whiskers, and outliers). Numbers indicate mean gains; red highlights immune escape and stability. \textbf{(b)} Measured escape and predicted constraint (lower predicted fitness) for SARS-CoV-2 XBB.1.5 receptor-binding domain (RBD) and Zika virus (ZIKV) MR766 envelope protein (E); colors show percentiles within each panel and gray marks missing sites. \textbf{(c)} Parkin N232 and R396 in structural context. \textbf{(d)} Domain-wise Spearman. \textbf{(e)} N232A and R396H move closer to their experimentally measured ranks.}
\label{fig:viral_parkin}
\end{figure}

\paragraph{Strong gains in protein expression and activity.}
Across $71$ configurations, mean gains on \ViroBenchmarkName{} are largest for expression ($+0.194$) and activity ($+0.150$), and smaller for immune escape ($+0.009$) and stability ($+0.007$; Figure~\ref{fig:viral_parkin}a). We analyze the $21$ escape assays across $12$ viral protein sequences. The highest Spearman averaged across proteins is $0.046$, achieved by unmasked \textsc{S3F} with the harness (Table~\ref{tab:leaderboard_vvh_category}).

\paragraph{Escape assays reveal distinct selective pressures.}
Across the $21$ escape assays, mean Spearman over the $71$ configurations with the harness ranges from $-0.30$ to $+0.21$. \reviewchange{Assays with negative correlations involve exposed sites in the spike receptor-binding domain (RBD) of severe acute respiratory syndrome coronavirus 2 (SARS-CoV-2), and antibody-contact sites in the Zika virus (ZIKV) envelope (E) protein.} Frozen models score many escape substitutions as deleterious (Figure~\ref{fig:viral_parkin}b)\reviewchange{, so higher measured escape can coincide with lower predicted fitness}. \reviewchange{These locations help explain the negative correlations}: \reviewchange{mutations predicted to impair the protein} can \reviewchange{still allow escape from} antibody recognition.

\subsection{Ranking gains across Parkin domains}
\label{sec:parkin_case}

\paragraph{Improvements across Parkin domains}.
In the Parkin abundance measurements \citep{clausen2024parkin}, unmasked \textsc{ESM-2} 650M, MSA fusion, and the full harness reach Spearman $0.503$, $0.585$, and $0.645$ across $8{,}756$ substitutions at $462$ sites. The full harness improves folded domains, including RING0 ($0.425$ to $0.627$) and RING2 ($0.427$ to $0.529$; Figure~\ref{fig:viral_parkin}d). Across ProteinGym, regions with pLDDT at least $50$ improve in $83\%$ of $214$ qualifying assays (mean Spearman gain $+0.056$). Appendix~\ref{app:parkin_details} compares linkers and regions with low pLDDT.

\paragraph{Corrections address both overestimation and underestimation.}
N232A, at the start of RING1, is measured at the $0.4$th percentile but placed at the $88.8$th percentile by the original model. MSA fusion and the full harness lower it to $58.0$ and $37.3$. R396H in the REP/tether region instead rises from $39.7$ to $71.7$ and $78.0$, approaching its measured $77.6$th percentile (Figure~\ref{fig:viral_parkin}c,e). AUC for classifying variants by abundance also improves, from $0.769$ for the original model to $0.815$ with MSA fusion and $0.857$ with the full harness (Appendix~\ref{app:parkin_details}).

\subsection{Scoring methods, efficiency, and mutation interactions}
\label{sec:efficient_inference}

\paragraph{Scoring methods.}
Masked and WT scoring both benefit from the harness across eight checkpoints on all three benchmarks. Before applying the harness, \mbox{\textsc{ProteinMPNN}} teacher forcing \reviewchange{on the WT sequence improves on the official scoring method, raising} mean assay Spearman from $0.271$--$0.290$ to $0.376$--$0.401$ across eight checkpoints (Appendices~\ref{app:interface_audit}--\ref{app:pmpnn_audit}).

\paragraph{Caching speeds up variant scoring.}
For \textsc{ProteinMPNN}, \reviewchange{caching one score matrix per protein is} $219$--$601\times$ \reviewchange{faster than rebuilding mutant sequences. After \textsc{ProSST} inference, cached mutation lookup is} $20.7\times$ \reviewchange{faster and MSA amino-acid counting with array operations is} $62.3\times$ \reviewchange{faster than the corresponding loops} (Appendix Figure~\ref{fig:runtime}; Appendix~\ref{app:amortized}).

\paragraph{\reviewchange{Single- and double-mutant ranking.}}
\reviewchange{On the GB1 binding measurements of \citet{olson2014epistasis}, \textsc{ProteinMPNN} teacher forcing improves on its official scoring method, raising single-mutant Spearman from $0.105$ to $0.306$ and double-mutant Spearman from $0.165$ to $0.284$. Rescoring substitutions with their partner mutations still gives near-zero correlations with measured interaction effects. The gains mainly reflect improved individual mutation effects (Appendix~\ref{app:epistasis}; Table~\ref{tab:pmpnn_epistasis}).}

\FloatBarrier

\section{Related Work}
\label{sec:related_work}

\paragraph{\reviewchange{Protein foundation models.}}
\reviewchange{PFMs differ in how they use sequence and structure. \emph{Structure-aware models}, including \textsc{SaProt} and \textsc{ProSST}, combine amino-acid and structure tokens \citep{su2023saprot,li2024prosst}. \emph{Masked sequence language models}, including \textsc{ESM-1b}, \textsc{ESM-1v}, \textsc{ESM-2}, and \textsc{CARP}, predict missing residues from sequence context \citep{rives2021esm1b,meier2021esm1v,lin2023esm2,carp}. \emph{Inverse-folding models}, including \textsc{ESM-IF1}, \textsc{ProteinMPNN}, and \textsc{MIF-ST}, predict sequences compatible with a protein backbone \citep{hsu2022esm-if1,dauparas2022proteinmpnn,mifst}. \emph{Autoregressive language models}, such as \textsc{RITA} and \textsc{ProGen2}, predict each residue from preceding residues \citep{hesslow2022rita,nijkamp2023progen2}.}

\paragraph{\reviewchange{Post-hoc calibration and post-training.}}
\reviewchange{\textsc{MSA Transformer} and \textsc{PoET} use homologous sequences as inputs \citep{rao2021msa,truong2023poet}. \emph{Post-training} learns or updates parameters through fitness-guided or unsupervised fine-tuning \citep{lee2023finetuning_plm_rank,zhou2024fsfp,gordon2025preference}, structural adapters \citep{tan2024sesadapter}, fine-tuning on retrieved MSAs (\textsc{AIDO-RAG}) \citep{li2024aidorag}, or learned homology retrieval \citep{notin2025protriever}. \emph{Post-hoc calibration} adjusts scores at inference. \textsc{Tranception}, \textsc{TranceptEVE}, \textsc{S3F-MSA}, and \textsc{VenusREM} add family information \citep{notin2022tranception,notin2022trancepteve,zhang2024s3f,tan2025venusrem}. \textsc{S3F} uses structural confidence and \textsc{RSALOR} solvent accessibility \citep{zhang2024s3f,tsishyn2025rsalor}.}

\paragraph{\reviewchange{Mutation fitness benchmarks.}}
\reviewchange{The \textsc{DeepSequence} study compiled DMS datasets for evaluating mutation-effect predictions \citep{riesselman2018deepsequence}. MaveDB archives variant-effect assays \citep{esposito2019mavedb}, while \textsc{FLIP} tests generalization from limited data \citep{dallago2021flip}. FireProtDB and SoluProtMutDB collect stability and solubility measurements, respectively \citep{stourac2021fireprotdb,velecky2022soluprotmutdb}. ProteinGym standardizes DMS evaluation \citep{notin2024proteingym}, and VenusMutHub covers small-scale biochemical measurements \citep{zhang2025venusmuthub}. ViroGym evaluates viral protein mutation effects \citep{zhou2026virogym}.}

\section{Discussion and Conclusion}
\label{sec:discussion}

While LLM harnesses are increasingly used in practice, harnesses for biological foundation models remain less explored. \textsc{\HarnessName{}} improves fitness prediction across frozen PFM families by combining evolutionary evidence, background correction, and structural weighting without additional training. However, the limited accuracy on immune-escape assays highlights a broader challenge: intrinsic protein constraints alone cannot fully capture phenotypes.

\leavevmode These findings highlight the distinction between refining existing model scores and capturing information absent from their inputs. Future work could incorporate protein--protein and protein--small-molecule interactions to improve mutation-effect prediction across diverse biological settings.

\clearpage
\ifvenusarxiv
% Funding acknowledgements for the public preprint only.
% Loaded conditionally by 0main.tex when compiling 0main_arxiv.tex.
\subsection*{Acknowledgement}
This work was supported by the National Key Research and Development Program of China (2024YFA0917603); Shanghai Municipal Science and Technology Major Project; the AI for Science Program, Shanghai Municipal Commission of Economy and Informatization (2025-GZL-RGZN-BTBX-02009); the Computational Biology Key Program of Shanghai Science and Technology Commission (23JS1400600); Shanghai Jiao Tong University Scientific and Technological Innovation Funds (21X010200843); Science and Technology Innovation Key R\&D Program of Chongqing (CSTB2022TIAD-STX0017, CSTB2024TIAD-STX0032); the Student Innovation Center at Shanghai Jiao Tong University, and Shanghai Artificial Intelligence Laboratory.

\fi
\subsection*{AI Use Statement}
OpenAI Codex assisted with research design, coding, data processing and analysis, derivation checks, figures, literature review, writing, and translation. It was not used to generate synthetic datasets. The authors reviewed all AI-assisted work, verified results and citations against their sources, and take full responsibility for the paper, code, and figures.

\subsection*{Ethics Statement}
This study uses public protein data, with no human participants or personal data. Predictions require experimental validation. Reuse must follow the original data and model licenses.

\subsection*{Reproducibility Statement}
The repositories below provide code, data, setup instructions, evaluation scripts, and experiment configurations. Section~\ref{sec:method} and Appendix~\ref{app:evaluation_details} describe the method and evaluation. Appendices~\ref{app:baselines}--\ref{app:additional_tables} provide implementation details and complete results, and Appendix~\ref{app:usage} explains usage.

{\urlstyle{same}
\noindent GitHub: %
\ifx\GitHubURL\empty
  \textit{[anonymous repository URL to be added]}%
\else
  \expandafter\url\expandafter{\GitHubURL}%
\fi\\
Hugging Face: %
\ifx\HuggingFaceURL\empty
  \textit{[anonymous dataset URL to be added]}%
\else
  \expandafter\url\expandafter{\HuggingFaceURL}%
\fi\par}

\FloatBarrier

\bibliography{0references,references_venusvirohub}
\bibliographystyle{iclr2027_conference}

\clearpage
\appendix

\section{Evaluation, validation, and data sources}
\label{app:evaluation_details}

Appendix~\ref{app:metric_defs} defines the metric-specific evaluation sets and aggregation. Appendix~\ref{app:input_distributions} describes sequence, alignment, and structural inputs; Appendix~\ref{app:virogym_data_processing} documents viral assay sources; Appendix~\ref{app:virogym} summarizes the resulting benchmark and results across models.

\subsection{Metric definitions and benchmark aggregation}
\label{app:metric_defs}
For an assay with $n$ retained pairs of experimental effects and predicted scores $(y_i,s_i)$, both oriented so that larger values are preferred, let $u_i=\operatorname{rank}(y_i)$ and $v_i=\operatorname{rank}(s_i)$ use average ranks for ties. Their arithmetic means are $\bar u$ and $\bar v$. Assay-level Spearman correlation is
\begin{equation}
\rho_{\mathrm{S}}(y,s)=
\frac{\sum_{i=1}^{n}(u_i-\bar u)(v_i-\bar v)}
{\sqrt{\sum_{i=1}^{n}(u_i-\bar u)^2}\sqrt{\sum_{i=1}^{n}(v_i-\bar v)^2}}.
\label{eq:metric_spearman}
\end{equation}

ProteinGym and \ViroBenchmarkName{} use NDCG@10\% with linear min--max gains $g_i=(y_i-\min_j y_j)/(\max_j y_j-\min_j y_j)$ and $k=\lfloor0.1n\rfloor$. Here $r_i^{(s)}$ and $r_i^{(y)}$ are ordinal descending ranks under $s$ and $y$; the evaluator calculates these ranks as \texttt{argsort(argsort(-score))+1} in NumPy, using its default sorting order for ties. The predicted and ideal discounted gains are
\begin{equation}
\begin{gathered}
\mathrm{DCG}_{10}=\sum_{i:r_i^{(s)}\le k}\frac{g_i}{\log_2(r_i^{(s)}+1)},
\qquad
\mathrm{IDCG}_{10}=\sum_{i:r_i^{(y)}\le k}\frac{g_i}{\log_2(r_i^{(y)}+1)},\\
\mathrm{NDCG}_{10}=\frac{\mathrm{DCG}_{10}}{\mathrm{IDCG}_{10}}.
\end{gathered}
\label{eq:metric_ndcg}
\end{equation}
VenusMutHub retains its full-list NDCG definition. With the same normalized relevance $g_i$, it uses exponential gain $2^{g_i}-1$ at all $n$ positions:
\begin{equation}
\mathrm{NDCG}_{\mathrm{full}}
=\frac{\sum_{i=1}^{n}(2^{g_i}-1)/\log_2(r_i^{(s)}+1)}
{\sum_{i=1}^{n}(2^{g_i}-1)/\log_2(r_i^{(y)}+1)}.
\label{eq:metric_ndcg_full}
\end{equation}
Its prediction sort is stable, preserving input order for ties. Thus NDCG values are compared before and after applying the harness within a benchmark, with the benchmark-specific definition held fixed.

Let $Q_{0.9}(\cdot)$ be the empirical 90th percentile with linear interpolation, and define $T_y=\{i:y_i\ge Q_{0.9}(y)\}$ and $T_s=\{i:s_i\ge Q_{0.9}(s)\}$. Ties at the threshold are retained, so these sets may exceed $10\%$ of variants. Top-variant recall@10\% is
\begin{equation}
\mathrm{Recall}_{10}=\frac{|T_y\cap T_s|}{|T_y|}.
\label{eq:metric_toprecall}
\end{equation}

For binary classification, ProteinGym uses its supplied \texttt{DMS\_score\_bin} labels. VenusMutHub and \ViroBenchmarkName{} use $z_i=\mathbf{1}\{y_i\ge\operatorname{median}(y)\}$, where $\mathbf{1}\{\cdot\}$ is an indicator. With $n_+=\sum_i z_i$ and $n_-=n-n_+$, AUC uses continuous prediction scores and half credit for tied positive--negative pairs:
\begin{equation}
\mathrm{AUC}=\frac{1}{n_+n_-}\sum_{i:z_i=1}\sum_{j:z_j=0}
\left[\mathbf{1}\{s_i>s_j\}+\tfrac12\mathbf{1}\{s_i=s_j\}\right].
\label{eq:metric_auc}
\end{equation}
For MCC, predicted labels are $\hat z_i=\mathbf{1}\{s_i\ge\operatorname{median}(s)\}$, giving
\begin{equation}
\mathrm{MCC}=\frac{\mathrm{TP}\,\mathrm{TN}-\mathrm{FP}\,\mathrm{FN}}
{\sqrt{(\mathrm{TP}+\mathrm{FP})(\mathrm{TP}+\mathrm{FN})(\mathrm{TN}+\mathrm{FP})(\mathrm{TN}+\mathrm{FN})}}.
\label{eq:metric_mcc}
\end{equation}
Here $\mathrm{TP},\mathrm{TN},\mathrm{FP},\mathrm{FN}$ count true positives, true negatives, false positives, and false negatives under $(z_i,\hat z_i)$. Spearman requires nonconstant rankings; NDCG requires nonzero relevance range, positive ideal gain, and, for the truncated version, $k\ge1$; AUC requires both experimental classes. The \ViroBenchmarkName{} evaluator excludes non-finite pairs, rejects assays with fewer than three pairs or constant scores, and records a zero-denominator MCC as undefined. ProteinGym and VenusMutHub likewise require both experimental classes for AUC/MCC, but their evaluator returns MCC $0$ for constant predicted labels. Undefined values are omitted from subsequent means. All metrics are computed per assay before benchmark aggregation; VenusMutHub reports the mean over assays. For metric $q$, let $\mathcal A^{(q)}_{pb}$ be the assays with a defined value for phenotype $p$ and viral protein $b$, let $\mathcal B^{(q)}_p=\{b:\mathcal A^{(q)}_{pb}\neq\varnothing\}$, and let $\mathcal P^{(q)}=\{p:\mathcal B^{(q)}_p\neq\varnothing\}$. Writing the per-assay value as $m^{(q)}_{pba}$ gives
\begin{equation}
M^{(q)}_{\mathrm{\ViroBenchmarkName{}}}=\frac{1}{|\mathcal P^{(q)}|}\sum_{p\in\mathcal P^{(q)}}
\frac{1}{|\mathcal B^{(q)}_p|}\sum_{b\in\mathcal B^{(q)}_p}
\frac{1}{|\mathcal A^{(q)}_{pb}|}\sum_{a\in\mathcal A^{(q)}_{pb}}m^{(q)}_{pba}.
\label{eq:venusvirohub_aggregation}
\end{equation}
This prevents densely assayed viral proteins from dominating. Spearman, NDCG, and top-variant recall use all $89$ assays and $52$ phenotype--protein groups; AUC and MCC use $82$ assays and $45$ groups because seven CVB3 fitness assays have single-class labels. ProteinGym leaderboard Spearman follows the official 217-assay evaluator; extended metrics retain the benchmark-specific definitions above.

Official Average Spearman, mean assay Spearman for VenusMutHub, and \ViroBenchmarkName{} hierarchical aggregation follow Section~\ref{sec:benchmarks}. Uncertainty uses 20{,}000 paired bootstrap draws over 186 ProteinGym UniProt IDs, 527 VenusMutHub WT-sequence clusters, or the metric-valid \ViroBenchmarkName{} phenotype--protein groups within each of seven phenotypes. For \ViroBenchmarkName{}, seed 0 and the same metric-specific draws are used across all configurations for comparisons before and after applying the harness. Configurations are held fixed, so intervals quantify uncertainty over benchmark units rather than a model population. Appendix~\ref{app:bias_details} gives the site-profile and separation protocols; Appendix~\ref{app:staged_audits} gives component selection and leave-one-site-out checks.

\paragraph{\reviewchange{Statistical comparisons of components.}}
\label{app:component_statistics}
\reviewchange{Tables~\ref{tab:prosst_component_statistics} and~\ref{tab:saprot_component_statistics} report paired score changes $\Delta$ (evaluated minus reference) between frozen configurations. ProteinGym validation uses the unrounded mean over 21 assays with distinct UniProt IDs. Full ProteinGym rounds assay Spearman to three decimals, then averages within UniProt--function pairs, over UniProt IDs within functions, and equally over represented functions. Other benchmarks retain the aggregation above.}

\reviewchange{For these comparisons, 95\% percentile confidence intervals use 20{,}000 paired group-bootstrap draws (seed 0), with groups defined above. Two-sided $p$-values use 20{,}000 permutations (seed 1), each independently swapping configuration labels for entire groups and recomputing $\Delta$. Under the null of exchangeable labels,}
\begin{equation*}
\reviewchange{p=\frac{1+\sum_{b=1}^{20{,}000}\mathbf{1}\!\left\{|\Delta_b^{\mathrm{perm}}|\geq|\Delta|\right\}}{20{,}001}.}
\end{equation*}

\subsection{Dataset and input distributions}
\label{app:input_distributions}

\begin{figure}[!ht]
\centering
\includegraphics[width=\textwidth]{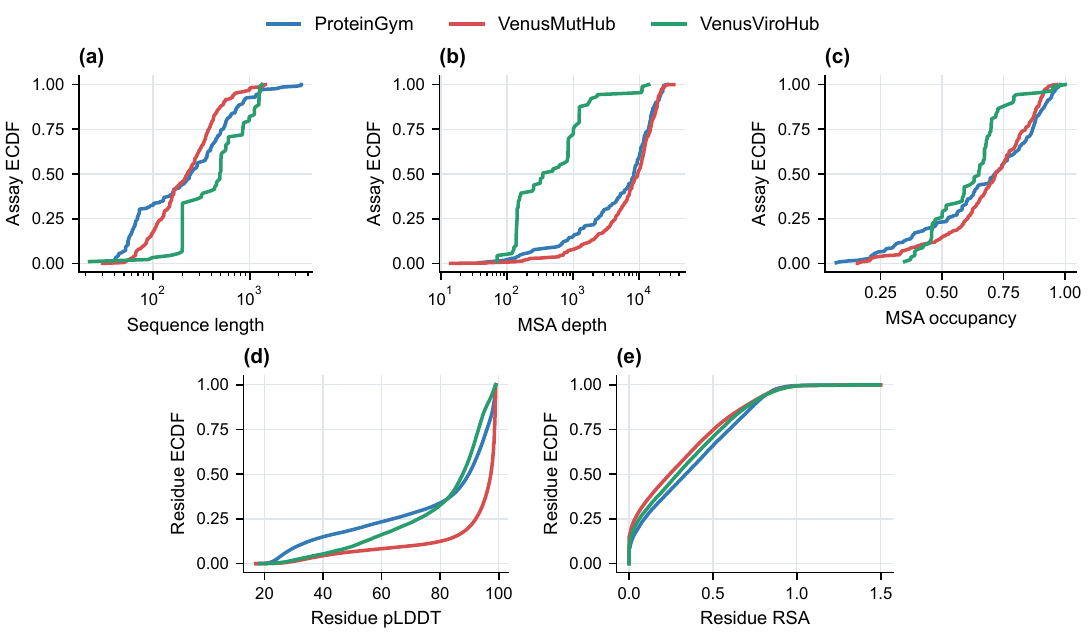}
\caption{Input distributions for ProteinGym (217 assays), VenusMutHub (905), and \ViroBenchmarkName{} (89). \textbf{(a)} Sequence length. \textbf{(b)} MSA depth. \textbf{(c)} MSA occupancy. \textbf{(d)} Residue pLDDT. \textbf{(e)} Residue RSA. Structure panels pool unique WT sequences; all selected structures fully cover the query.}
\label{fig:dataset_stats}
\end{figure}

\paragraph{Sequence and alignment distributions.}
ProteinGym contains 217 assays spanning 186 unique UniProt IDs and 187 unique wild-type (WT) sequences, VenusMutHub contains 905 assays across 527 unique WT clusters, and \ViroBenchmarkName{} contributes 89 assays across 52 unique WT sequences after duplicate removal. Figure~\ref{fig:dataset_stats}a--c and Table~\ref{tab:dataset_stats} compare their inputs. \ViroBenchmarkName{} sequences are longer on average (559 residues versus 397 and 288) but have shallower MSAs (1{,}217 sequences versus 8{,}650 and 9{,}791); mean MSA occupancy is 0.62, compared with 0.67 for ProteinGym and 0.69 for VenusMutHub.

\begin{table}[!ht]
\caption{Assay-level input statistics for ProteinGym, VenusMutHub, and \ViroBenchmarkName{}, reported as mean $\pm$ standard deviation. $N$ is the number of assays; MSA occupancy is the fraction of non-gap match-state entries; pLDDT and RSA are assay-level means.}
\label{tab:dataset_stats}
\centering
\scriptsize
\setlength{\tabcolsep}{2.0pt}
\resizebox{\textwidth}{!}{%
\begin{tabular}{lrrrrrr}
\toprule
Dataset & $N$ assays & Sequence length & MSA depth & MSA occupancy & Mean pLDDT & Mean RSA \\
\midrule
ProteinGym & 217 & 397 $\pm$ 502 & 8,650 $\pm$ 6,713 & 0.67 $\pm$ 0.23 & 86.6 $\pm$ 10.8 & 0.37 $\pm$ 0.09 \\
VenusMutHub & 905 & 288 $\pm$ 239 & 9,791 $\pm$ 6,452 & 0.69 $\pm$ 0.18 & 92.1 $\pm$ 7.5 & 0.32 $\pm$ 0.08 \\
\ViroBenchmarkName{} & 89 & 559 $\pm$ 383 & 1,217 $\pm$ 2,610 & 0.62 $\pm$ 0.14 & 83.9 $\pm$ 9.8 & 0.33 $\pm$ 0.07 \\
\bottomrule
\end{tabular}%
}
\end{table}

\paragraph{Structure distributions.}
pLDDT is read from C$\alpha$ B-factors and RSA from Shrake--Rupley accessibility normalized by residue-specific maximum ASA. Selected AlphaFold2 models cover every query residue in all three benchmarks. \ViroBenchmarkName{} has the lowest assay-level mean pLDDT (83.9), with 41.1\% of residues at pLDDT $\geq90$ and 9.7\% below 50, while its mean RSA (0.33) lies between ProteinGym (0.37) and VenusMutHub (0.32). Figure~\ref{fig:dataset_stats}d--e and Tables~\ref{tab:plddt_ratios}--\ref{tab:rsa_ratios} report the full distributions.

\begin{table}[!ht]
\caption{AlphaFold2 confidence ranges over unique wild-type sequences. All selected structures cover the full query; residue percentages are pooled within each benchmark.}
\label{tab:plddt_ratios}
\centering
\small
\setlength{\tabcolsep}{6pt}
\begin{tabular}{lrrrrrr}
\toprule
Dataset & Unique WT & Structures & pLDDT $<50$ & 50--70 & 70--90 & $\geq90$ \\
\midrule
ProteinGym & 187 & 187 & 18.9\% & 9.0\% & 22.6\% & 49.5\%\\
VenusMutHub & 527 & 527 & 6.7\% & 3.3\% & 10.6\% & 79.5\%\\
\ViroBenchmarkName{} & 52 & 52 & 9.7\% & 13.2\% & 36.0\% & 41.1\%\\
\bottomrule
\end{tabular}
\end{table}

\begin{table}[!ht]
\caption{RSA ranges over unique wild-type sequences. Residue percentages are pooled within each benchmark.}
\label{tab:rsa_ratios}
\centering
\small
\begin{tabular}{lrrr}
\toprule
Dataset & RSA $<0.2$ & RSA $0.2$--$0.5$ & RSA $\geq0.5$ \\
\midrule
ProteinGym & 36.7\% & 29.4\% & 33.9\%\\
VenusMutHub & 45.8\% & 28.9\% & 25.3\%\\
\ViroBenchmarkName{} & 40.6\% & 30.2\% & 29.2\%\\
\bottomrule
\end{tabular}
\end{table}

\FloatBarrier
\subsection{\ViroBenchmarkName{} data collection and sources}
\label{app:virogym_data_processing}

\paragraph{Relation to existing benchmarks.}
Section~\ref{sec:virohub} describes how \ViroBenchmarkName{} complements existing benchmarks. We checked for duplicate measurements against ProteinGym and VenusMutHub after mapping mutation coordinates to corresponding WT sequences. No retained assay repeats a measurement from the same study, protein background, and phenotype in either benchmark. Individual substitutions or protein backgrounds can still be shared across experiments that measure different phenotypes or use different conditions.

\paragraph{Data collection.}
We identified candidate assays from public data portals, repositories, cited studies, and recent literature, then reconstructed each assay from its original materials. We retained quantitative measurements of single-amino-acid substitutions with an accessible mutant-level table, a recoverable wild-type sequence, mappable residue numbering, and approximately 400 or more substitutions after quality control. Assays were distinguished by study, virus strain, protein or scanned region, and phenotype. Mapped coordinates and sequences were used to identify and exclude duplicate measurements.

\paragraph{Data processing.}
Each assay was converted to a wild-type sequence, single-substitution identifier, and experimental score. Residue coordinates followed the experimental construct, using documented offsets or sequence alignment when needed. We removed missing scores, unmappable substitutions, reference-residue mismatches, and duplicate records. Replicates or experimental conditions were combined using the assay-specific rule recorded in the processing manifest. Score directions were harmonised so that larger values represent a more favourable outcome for the reported phenotype, while the original numerical scales were retained. After quality control, the files retain $642{,}737$ rows, comprising $616{,}690$ substitutions and $26{,}047$ wild-type controls (XnX). Controls may remain in data and score files but are excluded before computing any evaluation metric; all $89$ assays are retained.

\paragraph{Benchmark composition.}
Table~\ref{tab:venusvirohub_composition} gives the assay and substitution counts for all seven phenotype groups. The manifest records each assay's source, virus strain, experimental construct, coordinate convention, wild-type sequence hash, processing rule, and retained mutation count. Evaluation follows the metric-specific hierarchy in Equation~\ref{eq:venusvirohub_aggregation}.

\begin{table}[!ht]
\caption{Phenotype composition of \ViroBenchmarkName{} after quality control. Evaluated substitutions exclude wild-type control rows.}
\label{tab:venusvirohub_composition}
\begin{center}
\small
\setlength{\tabcolsep}{7pt}
\begin{tabular}{lrr}
\toprule
Harmonised phenotype & Assays & Evaluated substitutions \\
\midrule
Immune escape & 21 & 148{,}468 \\
Cell entry & 18 & 189{,}048 \\
Binding & 17 & 90{,}245 \\
Expression & 15 & 58{,}120 \\
Fitness & 15 & 114{,}287 \\
Stability & 2 & 11{,}187 \\
Activity & 1 & 5{,}335 \\
\midrule
Total & 89 & 616{,}690 \\
\bottomrule
\end{tabular}
\end{center}
\end{table}

\paragraph{Assay sources.}
Tables~\ref{tab:venusvirohub_provenance}--\ref{tab:vvh_prov_stability} link every retained assay to what it measures and its original study or public data repository, grouped by measurement type. Exact assay identifiers, construct coordinates, retained mutation counts, coordinate transformations, and assay-specific combination rules are provided in the processing manifest.

Antibody escape is recorded separately for monoclonal antibodies (Table~\ref{tab:venusvirohub_provenance}) and polyclonal sera (Table~\ref{tab:vvh_prov_escape_serum}). This distinction preserves the experimental context of the immune-escape comparisons.

\begin{table}[H]
\centering
\caption{\ViroBenchmarkName{} assays measuring \emph{Monoclonal-antibody escape} (15 assays). Distinct rows from the same study correspond to different viral constructs or measured phenotypes. \textsuperscript{\dag}Public dataset repository without an associated archival article.}
\label{tab:venusvirohub_provenance}
\small
\setlength{\tabcolsep}{5pt}
\begin{tabular}{@{}>{\raggedright\arraybackslash}p{0.42\textwidth}>{\raggedright\arraybackslash}p{0.50\textwidth}@{}}
\toprule
Virus and construct & Source \\
\midrule
HIV-1 BF520 Env & \citep{radford2025hiv} \\
HIV-1 TRO11 Env & \citep{radford2025hiv} \\
Influenza A/H3N2 HK19 HA & \citep{welsh2024h3} \\
Influenza A/H3N2 MC22 HA & \citep{yu2025h3} \\
Lassa virus Josiah GP & \citep{carr2024lassa} \\
Nipah virus F & \citep{larsen2026nipahf} \\
Nipah virus Malaysia RBP & \citep{larsen2025nipah} \\
Nipah virus Malaysia RBP & \citep{larsen2026minibinder}\textsuperscript{\dag} \\
Rabies virus G & \citep{aditham2025rabies} \\
Respiratory syncytial virus A F & \citep{simonich2026rsv} \\
SARS-CoV-2 Wuhan-Hu-1 RBD & \citep{cao2022imprinted} \\
SARS-CoV-2 Omicron BA.2 spike & \citep{dadonaite2026crowe}\textsuperscript{\dag} \\
SARS-CoV-2 KP.3.1.1 spike & \citep{dadonaite2025kp311} \\
SARS-CoV-2 XBB.1.5 spike & \citep{dadonaite2024xbb} \\
Zika virus E & \citep{sourisseau2019zikv,kikawa2023zikv} \\
\bottomrule
\end{tabular}
\end{table}

\begin{table}[H]
\centering
\caption{\ViroBenchmarkName{} assays measuring \emph{Serum-antibody escape} (6 assays).}
\label{tab:vvh_prov_escape_serum}
\small
\setlength{\tabcolsep}{5pt}
\begin{tabular}{@{}>{\raggedright\arraybackslash}p{0.42\textwidth}>{\raggedright\arraybackslash}p{0.50\textwidth}@{}}
\toprule
Virus and construct & Source \\
\midrule
HIV-1 BF520 Env & \citep{radford2023hiv} \\
Human coronavirus 229E spike & \citep{harari2026cov229e} \\
Influenza A/H5N1 HA & \citep{dadonaite2024h5} \\
Lassa virus Josiah GP & \citep{carr2026lassvsera}\textsuperscript{\dag} \\
Respiratory syncytial virus A F & \citep{simonich2026rsv} \\
SARS-CoV-2 XBB.1.5 RBD & \citep{dadonaite2024xbb} \\
\bottomrule
\end{tabular}
\end{table}

Cell-entry assays (Table~\ref{tab:vvh_prov_entry}) and receptor-binding assays (Table~\ref{tab:vvh_prov_binding}) probe related but distinct steps in host interaction; their measurements are kept separate in the manifest.

\begin{table}[H]
\centering
\caption{\ViroBenchmarkName{} assays measuring \emph{Cell entry} (18 assays).}
\label{tab:vvh_prov_entry}
\small
\setlength{\tabcolsep}{5pt}
\begin{tabular}{@{}>{\raggedright\arraybackslash}p{0.42\textwidth}>{\raggedright\arraybackslash}p{0.50\textwidth}@{}}
\toprule
Virus and construct & Source \\
\midrule
Chikungunya virus E3/E2/E1 & \citep{ju2025chikv} \\
Chikungunya virus E3/E2/E1 & \citep{ju2025chikv} \\
HIV-1 BF520 Env & \citep{radford2025hiv} \\
HIV-1 TRO11 Env & \citep{radford2025hiv} \\
Human coronavirus 229E spike & \citep{harari2026cov229e} \\
Influenza A/H3N2 MC22 HA & \citep{yu2025h3} \\
Influenza A/H5N1 HA & \citep{dadonaite2024h5} \\
Influenza A/H7N9 Anhui13 HA & \citep{yu2026h7} \\
Influenza A/H7N9 Anhui13 HA & \citep{yu2026h7} \\
Lassa virus Josiah GP & \citep{carr2024lassa} \\
Nipah virus F & \citep{larsen2026nipahf} \\
Nipah virus Malaysia RBP & \citep{larsen2025nipah} \\
Rabies virus G & \citep{aditham2025rabies} \\
Respiratory syncytial virus A F & \citep{simonich2026rsv} \\
SARS-CoV-2 XBB.1.5 RBD & \citep{dadonaite2024xbb} \\
SARS-CoV-2 Omicron BA.2 spike & \citep{dadonaite2024xbb} \\
SARS-CoV-2 KP.3.1.1 spike & \citep{dadonaite2025kp311} \\
SARS-CoV-2 XBB.1.5 spike & \citep{dadonaite2024xbb} \\
\bottomrule
\end{tabular}
\end{table}

\begin{table}[H]
\centering
\caption{\ViroBenchmarkName{} assays measuring \emph{Receptor binding} (17 assays).}
\label{tab:vvh_prov_binding}
\small
\setlength{\tabcolsep}{5pt}
\begin{tabular}{@{}>{\raggedright\arraybackslash}p{0.42\textwidth}>{\raggedright\arraybackslash}p{0.50\textwidth}@{}}
\toprule
Virus and construct & Source \\
\midrule
Chikungunya virus E3/E2/E1 & \citep{ju2025chikv} \\
Human coronavirus 229E spike & \citep{harari2026cov229e} \\
Nipah virus Malaysia RBP & \citep{larsen2025nipah} \\
SARS-CoV-2 Alpha RBD & \citep{starr2022shifting} \\
SARS-CoV-2 Beta RBD & \citep{starr2022shifting} \\
SARS-CoV-2 Delta RBD & \citep{starr2022shifting} \\
SARS-CoV-2 Eta RBD & \citep{starr2022shifting} \\
SARS-CoV-2 Omicron BA.1 RBD & \citep{starr2022omicron} \\
SARS-CoV-2 Omicron BA.2.86 RBD & \citep{taylor2024deep} \\
SARS-CoV-2 Omicron BA.2 RBD & \citep{starr2022omicron} \\
SARS-CoV-2 Omicron BQ.1.1 RBD & \citep{taylor2023deep} \\
SARS-CoV-2 Omicron EG.5 RBD & \citep{taylor2024deep} \\
SARS-CoV-2 Omicron FLip RBD & \citep{taylor2024deep} \\
SARS-CoV-2 Omicron XBB.1.5 RBD & \citep{taylor2023deep} \\
SARS-CoV-2 Omicron BA.2 spike & \citep{dadonaite2024xbb} \\
SARS-CoV-2 KP.3.1.1 spike & \citep{dadonaite2025kp311} \\
SARS-CoV-2 XBB.1.5 spike & \citep{dadonaite2024xbb} \\
\bottomrule
\end{tabular}
\end{table}

Table~\ref{tab:vvh_prov_expression} records protein-expression and abundance assays. These assays measure protein production and persistence, complementing the antibody-recognition assays.

\begin{table}[H]
\centering
\caption{\ViroBenchmarkName{} assays measuring \emph{Protein expression and abundance} (15 assays).}
\label{tab:vvh_prov_expression}
\small
\setlength{\tabcolsep}{5pt}
\begin{tabular}{@{}>{\raggedright\arraybackslash}p{0.42\textwidth}>{\raggedright\arraybackslash}p{0.50\textwidth}@{}}
\toprule
Virus and construct & Source \\
\midrule
Bat coronavirus PRD0038 RBD & \citep{starr2022sarbecovirus} \\
Bat coronavirus RmYN02 RBD & \citep{starr2022sarbecovirus} \\
Bat coronavirus RsYN04 RBD & \citep{starr2022sarbecovirus} \\
SARS-CoV-2 PLpro & \citep{wu2024plpro} \\
SARS-CoV-2 Alpha RBD & \citep{starr2022shifting} \\
SARS-CoV-2 Beta RBD & \citep{starr2022shifting} \\
SARS-CoV-2 Delta RBD & \citep{starr2022shifting} \\
SARS-CoV-2 Eta RBD & \citep{starr2022shifting} \\
SARS-CoV-2 Omicron BA.1 RBD & \citep{starr2022omicron} \\
SARS-CoV-2 Omicron BA.2.86 RBD & \citep{taylor2024deep} \\
SARS-CoV-2 Omicron BA.2 RBD & \citep{starr2022omicron} \\
SARS-CoV-2 Omicron BQ.1.1 RBD & \citep{taylor2023deep} \\
SARS-CoV-2 Omicron EG.5 RBD & \citep{taylor2024deep} \\
SARS-CoV-2 Omicron FLip RBD & \citep{taylor2024deep} \\
SARS-CoV-2 Omicron XBB.1.5 RBD & \citep{taylor2023deep} \\
\bottomrule
\end{tabular}
\end{table}

Fitness assays (Table~\ref{tab:vvh_prov_fitness}) and stability or protease-activity assays (Table~\ref{tab:vvh_prov_stability}) complete the record of assay sources. Distinct constructs from one study remain separate assay entries.

\begin{table}[H]
\centering
\caption{\ViroBenchmarkName{} assays measuring \emph{Viral fitness and growth} (15 assays).}
\label{tab:vvh_prov_fitness}
\small
\setlength{\tabcolsep}{5pt}
\begin{tabular}{@{}>{\raggedright\arraybackslash}p{0.42\textwidth}>{\raggedright\arraybackslash}p{0.50\textwidth}@{}}
\toprule
Virus and construct & Source \\
\midrule
Coxsackievirus B3 2A & \citep{alvarezrodriguez2024cvb3} \\
Coxsackievirus B3 2B & \citep{alvarezrodriguez2024cvb3} \\
Coxsackievirus B3 2C & \citep{alvarezrodriguez2024cvb3} \\
Coxsackievirus B3 3A & \citep{alvarezrodriguez2024cvb3} \\
Coxsackievirus B3 3B & \citep{alvarezrodriguez2024cvb3} \\
Coxsackievirus B3 3C & \citep{alvarezrodriguez2024cvb3} \\
Coxsackievirus B3 3D & \citep{alvarezrodriguez2024cvb3} \\
Coxsackievirus B3 VP1 & \citep{alvarezrodriguez2024cvb3} \\
Coxsackievirus B3 VP3 & \citep{alvarezrodriguez2024cvb3} \\
Enterovirus A capsid & \citep{bakhache2024enterovirus} \\
Enterovirus A replicase & \citep{bakhache2024enterovirus} \\
Hepatitis B virus reverse transcriptase & \citep{yu2024hbv} \\
Influenza A/H3N2 HK19 HA & \citep{welsh2024h3} \\
SARS-CoV-2 Omicron BA.1 spike & \citep{dadonaite2023spike} \\
SARS-CoV-2 Delta spike & \citep{dadonaite2023spike} \\
\bottomrule
\end{tabular}
\end{table}

\begin{table}[H]
\centering
\caption{\ViroBenchmarkName{} assays measuring \emph{Protein stability and protease activity} (3 assays).}
\label{tab:vvh_prov_stability}
\small
\setlength{\tabcolsep}{5pt}
\begin{tabular}{@{}>{\raggedright\arraybackslash}p{0.42\textwidth}>{\raggedright\arraybackslash}p{0.50\textwidth}@{}}
\toprule
Virus and construct & Source \\
\midrule
Influenza A/H3N2 MC22 HA & \citep{yu2025h3} \\
Influenza A/H5N1 HA & \citep{dadonaite2024h5} \\
SARS-CoV-2 PLpro & \citep{wu2024plpro} \\
\bottomrule
\end{tabular}
\end{table}

\FloatBarrier
\subsection{\ViroBenchmarkName{} results}
\label{app:virogym}

We summarize results for all 71 configurations using the data described in Appendix~\ref{app:virogym_data_processing} and the metric-specific aggregation in Appendix~\ref{app:metric_defs}.

\paragraph{Gains across models.}
Under this hierarchy, mean Spearman rises from $0.158$ to $0.252$ ($+0.094$), with positive changes for all $71$ configurations. The largest gains occur for masked sequence models ($+0.151$; $0.072$ to $0.222$) and autoregressive models ($+0.107$; $0.099$ to $0.206$). Structure-aware encoders improve from $0.215$ to $0.280$ ($+0.065$), and inverse-folding models from $0.271$ to $0.307$ ($+0.036$). The harness gives the largest gains for sequence-only models and also improves models that already use structure (Table~\ref{tab:virogym_family}).

\paragraph{Metric and component consistency.}
Averaged over the same $71$ configurations, NDCG increases from $0.684$ to $0.729$, AUC from $0.584$ to $0.634$, MCC from $0.125$ to $0.202$, and top-variant recall from $0.133$ to $0.152$. Every configuration improves on every metric, yielding $355$ of $355$ positive model--metric comparisons (Tables~\ref{tab:viro_metrics_family_1}--\ref{tab:viro_metrics_family_4}). For \textsc{ProSST}-2048, adaptive MSA fusion raises Spearman from $0.2709$ to $0.2901$, and gated background correction reaches $0.2942$. RSA-only and pLDDT-only weighting reach $0.3085$ and $0.2994$, respectively, while the full harness reaches $0.3102$. Exposure weighting accounts for more of the structural gain on \ViroBenchmarkName{} (Table~\ref{tab:ablation}).

\FloatBarrier
\section{Baseline models and implementations}
\label{app:baselines}

Table~\ref{tab:baselines} lists the evaluated models, scoring methods, and official implementations used in Figure~\ref{fig:framework}. Each configuration combines a frozen checkpoint with a scoring method from Equation~\ref{eq:readouts}. Different model sizes or scoring methods count as separate configurations. Appendix~\ref{app:bias_details} compares their shared amino-acid preferences and differences among checkpoints. The \textsc{CARP} series includes 600K, 38M, 76M, and 640M models. \textsc{ProtSSN} includes 9 graph encoders combining $k\in\{10,20,30\}$ neighbors with hidden widths $h\in\{512,768,1280\}$, alongside their ensemble. Both series score the unmasked WT sequence.

The six \textsc{ProSST} checkpoints use WT scoring and differ in their structure-token vocabularies (Section~\ref{sec:venusrem2}). Appendix~\ref{app:ensemble_definition} specifies their standardization and combination in \textsc{VenusREM2}. \textsc{S3F} contributes masked and unmasked WT configurations from one checkpoint. \textsc{SaProt} uses both methods for its 35M-AF2, 650M-AF2, and 650M-PDB checkpoints. Structures are AlphaFold2 predictions \citep{jumper2021alphafold2}, and query-matched alignments are built using MMseqs2 via ColabFold \citep{steinegger2017mmseqs2,mirdita2022colabfold}.

The comparison in Figure~\ref{fig:leaderboard} is detailed in Tables~\ref{tab:leaderboard} and~\ref{tab:leaderboard_full}, which report official overall and biological-category scores for one representative per model series, covering sequence, structure, and evolutionary baselines. Table~\ref{tab:leaderboard_full_taxon_msa} keeps those same rows and replaces the five biological-category columns with official taxon and MSA-depth bins: \textsc{VenusREM2} is strongest \reviewchange{among these official leaderboard baselines} in every reported category, including virus ($0.540$) and the low-depth bin ($0.553$). Table~\ref{tab:leaderboard_full_mutation_depth} completes the comparison by mutation depth, the number of amino-acid substitutions per variant; \textsc{VenusREM2} ranks first in all five bins.

\begin{table}[!htbp]
\caption{Evaluated models, scoring methods, and implementations. Equation~\ref{eq:readouts} defines masked scoring (mask), unmasked WT scoring (wt), and teacher forcing (tf).}
\label{tab:baselines}
\begin{center}
\small
\renewcommand{\arraystretch}{1.08}
\setlength{\tabcolsep}{2pt}
\resizebox{\textwidth}{!}{%
\begin{tabular}{@{}lllll@{}}
\toprule
Model & Family & Configurations & Scoring & Source \\
\midrule
\textsc{ESM-1b} & Masked sequence LM & 650M & mask, wt & \href{https://github.com/facebookresearch/esm}{Official repository} \\
\textsc{ESM-1v} & Masked sequence LM & 650M & mask, wt & \href{https://github.com/facebookresearch/esm}{Official repository} \\
\textsc{ESM-2} & Masked sequence LM & 8M/35M/150M/650M/3B & mask, wt & \href{https://github.com/facebookresearch/esm}{Official repository} \\
\textsc{ESMC} & Masked sequence LM & 300M, 600M & mask & \href{https://github.com/evolutionaryscale/esm}{Official repository} \\
\textsc{CARP} & Masked sequence LM & 600K/38M/76M/640M & wt & \href{https://github.com/microsoft/protein-sequence-models}{Official repository} \\
\textsc{S3F} & Structure-aware & S3F & wt, mask & \href{https://github.com/DeepGraphLearning/S3F}{Official repository} \\
\textsc{ProGen2} & Autoregressive LM & B/S/M/L/XL & tf & \href{https://github.com/salesforce/progen}{Official repository} \\
\textsc{ProGen3} & Autoregressive LM & 112M/219M/339M/762M/1B/3B & tf & \href{https://github.com/Profluent-AI/progen3}{Official repository} \\
\textsc{RITA} & Autoregressive LM & S/M/L/XL & tf & \href{https://github.com/lightonai/rita}{Official repository} \\
\textsc{ESM3} & Structure-aware & open-weight (1.4B) & mask & \href{https://github.com/evolutionaryscale/esm}{Official repository} \\
\textsc{ProSST} & Structure-aware & $K{=}20/128/512/1024/2048/4096$ & wt & \href{https://github.com/ai4protein/ProSST}{Official repository} \\
\textsc{ProtSSN} & Structure-aware & $k{=}10/20/30$, $h{=}512/768/1280$; ensemble & wt & \href{https://github.com/ai4protein/ProtSSN}{Official repository} \\
\textsc{SaProt} & Structure-aware & 35M-AF2, 650M-AF2, 650M-PDB & mask, wt & \href{https://github.com/westlake-repl/SaProt}{Official repository} \\
\textsc{ESM-IF1} & Inverse folding & --- & tf & \href{https://github.com/facebookresearch/esm}{Official repository} \\
\textsc{ProteinMPNN} & Inverse folding & v\_48\_002/010/020/030; ProteinMPNN-Soluble $\times$4 & tf & \href{https://github.com/dauparas/ProteinMPNN}{Official repository} \\
\textsc{MIF-ST} & Inverse folding & --- & tf & \href{https://github.com/microsoft/protein-sequence-models}{Official repository} \\
\textsc{VenusREM} & Retrieval-enhanced & \textsc{ProSST} ($K{=}2048$) + MSA & wt & \href{https://github.com/ai4protein/VenusREM}{Official repository} \\
\textsc{RSALOR} & MSA + structure & conservation $\times$ RSA & --- & \href{https://github.com/3BioCompBio/RSALOR}{Official repository} \\
\bottomrule
\end{tabular}}%
\end{center}
\end{table}

\FloatBarrier
\begin{table}[!t]
\caption{ProteinGym zero-shot substitution leaderboard: official 217-assay Average Spearman. Bold red, purple, and black mark the top 3 distinct scores per column at reported precision.}
\label{tab:leaderboard}
\centering
\scriptsize
\setlength{\tabcolsep}{2.6pt}
\renewcommand{\arraystretch}{1.0}
\resizebox{\textwidth}{!}{%
\begin{tabular}{@{}rlccccccccc@{}}
\toprule
Rank & Model & Seq & Str & Evo & Average & Activity & Binding & Expression & Organismal & Stability \\
\midrule
--- & \textbf{\textsc{VenusREM2} (ours)} & \checkmark & \checkmark & \checkmark & \textcolor{red!75!black}{\textbf{0.556}} & \textcolor{red!75!black}{\textbf{0.541}} & \textcolor{red!75!black}{\textbf{0.495}} & \textcolor{red!75!black}{\textbf{0.557}} & \textcolor{red!75!black}{\textbf{0.494}} & \textcolor{red!75!black}{\textbf{0.691}} \\
\midrule
1 & \textsc{AIDO Protein-RAG} (16B) \citep{sun2024aido} & --- & \checkmark & \checkmark & \textcolor[HTML]{7030A0}{\textbf{0.518}} & \textcolor[HTML]{7030A0}{\textbf{0.517}} & 0.426 & 0.522 & \textcolor[HTML]{7030A0}{\textbf{0.491}} & 0.635 \\
2 & \textsc{VenusREM} \citep{tan2025venusrem} & \checkmark & \checkmark & \checkmark & \textcolor[HTML]{7030A0}{\textbf{0.518}} & 0.495 & \textcolor[HTML]{7030A0}{\textbf{0.454}} & \textcolor[HTML]{7030A0}{\textbf{0.533}} & 0.459 & \textbf{0.650} \\
3 & \textsc{ProSST} ($K{=}2048$) \citep{li2024prosst} & \checkmark & \checkmark & --- & \textbf{0.507} & 0.476 & \textbf{0.445} & \textbf{0.530} & 0.431 & \textcolor[HTML]{7030A0}{\textbf{0.653}} \\
5 & \textsc{S3F-MSA} \citep{zhang2024s3f} & --- & \checkmark & \checkmark & 0.496 & \textbf{0.502} & 0.440 & 0.479 & 0.477 & 0.581 \\
8 & \textsc{Protriever} \citep{notin2025protriever} & --- & --- & \checkmark & 0.479 & 0.487 & 0.396 & 0.496 & \textbf{0.479} & 0.537 \\
9 & \textsc{ESCOTT} \citep{tekpinar2025prescott} & --- & \checkmark & \checkmark & 0.476 & 0.499 & 0.389 & 0.468 & 0.466 & 0.557 \\
12 & \textsc{PoET} (200M) \citep{truong2023poet} & --- & --- & \checkmark & 0.470 & 0.494 & 0.396 & 0.466 & 0.475 & 0.519 \\
14 & \textsc{ESM3} open (1.4B) \citep{hayes2025esm3} & \checkmark & \checkmark & --- & 0.466 & 0.430 & 0.400 & 0.470 & 0.389 & 0.641 \\
15 & \textsc{RSALOR} \citep{tsishyn2025rsalor} & --- & \checkmark & \checkmark & 0.465 & 0.479 & 0.416 & 0.427 & 0.426 & 0.575 \\
16 & \textsc{VespaG} \citep{marquet2024vespag} & \checkmark & --- & --- & 0.458 & 0.493 & 0.370 & 0.456 & 0.437 & 0.533 \\
17 & \textsc{SaProt} (650M AF2) \citep{su2023saprot} & \checkmark & \checkmark & --- & 0.457 & 0.458 & 0.378 & 0.488 & 0.366 & 0.592 \\
18 & \textsc{TranceptEVE-L} \citep{notin2022trancepteve} & \checkmark & --- & \checkmark & 0.456 & 0.487 & 0.376 & 0.457 & 0.459 & 0.500 \\
20 & \textsc{GEMME} \citep{laine2019gemme} & --- & --- & \checkmark & 0.455 & 0.482 & 0.383 & 0.438 & 0.452 & 0.519 \\
23 & \textsc{ProtSSN} ensemble \citep{tan2025protssn} & \checkmark & \checkmark & --- & 0.449 & 0.466 & 0.366 & 0.449 & 0.396 & 0.568 \\
40 & \textsc{ESM-IF1} \citep{hsu2022esm-if1} & --- & \checkmark & --- & 0.422 & 0.368 & 0.389 & 0.407 & 0.324 & 0.624 \\
45 & \textsc{ESM-2} (650M) \citep{lin2023esm2} & \checkmark & --- & --- & 0.414 & 0.425 & 0.337 & 0.415 & 0.368 & 0.523 \\
47 & \textsc{ESM-1v} ensemble \citep{meier2021esm1v} & \checkmark & --- & --- & 0.407 & 0.420 & 0.320 & 0.429 & 0.386 & 0.477 \\
53 & \textsc{MIF-ST} \citep{mifst} & \checkmark & \checkmark & --- & 0.400 & 0.390 & 0.321 & 0.438 & 0.366 & 0.485 \\
57 & \textsc{ESM-1b} \citep{rives2021esm1b} & \checkmark & --- & --- & 0.394 & 0.428 & 0.287 & 0.406 & 0.349 & 0.500 \\
59 & \textsc{ProGen3} (3B) \citep{bhatnagar2025progen3} & \checkmark & --- & --- & 0.392 & 0.410 & 0.287 & 0.428 & 0.400 & 0.438 \\
71 & \textsc{RITA-XL} \citep{hesslow2022rita} & \checkmark & --- & --- & 0.373 & 0.366 & 0.302 & 0.414 & 0.384 & 0.398 \\
72 & \textsc{CARP} (640M) \citep{carp} & \checkmark & --- & --- & 0.369 & 0.395 & 0.273 & 0.397 & 0.366 & 0.412 \\
91 & \textsc{ProteinMPNN} \citep{dauparas2022proteinmpnn} & --- & \checkmark & --- & 0.257 & 0.197 & 0.163 & 0.198 & 0.164 & 0.565 \\
\bottomrule
\end{tabular}}%
\vspace{0.2em}

\parbox{\textwidth}{\scriptsize\textit{Source and notes.} Ranked rows use the official ProteinGym substitutions summary; one entry per model series, matching Table~\ref{tab:leaderboard_full}. \textsc{VenusREM2} uses the six-checkpoint \textsc{ProSST} ensemble in Table~\ref{tab:stratified_taxa_msa}.}
\end{table}

\FloatBarrier
\begin{table}[!htbp]
\caption{Complete zero-shot ProteinGym substitutions comparison under the official 217-assay Average Spearman protocol, split by biological category. The same models appear by taxon and MSA depth in Table~\ref{tab:leaderboard_full_taxon_msa}. Bold red, purple, and black mark the top 3 distinct scores per column at reported precision.}
\label{tab:leaderboard_full}
\begin{center}
\fontsize{8.5}{10}\selectfont
\setlength{\tabcolsep}{1.5pt}
\renewcommand{\arraystretch}{1.08}
\resizebox{\textwidth}{!}{%
\begin{tabular}{@{}rlccccccccc@{}}
\toprule
Rank & Model & Seq & Str & Evo & Average & Activity & Binding & Expression & Organismal & Stability \\
\midrule
--- & \textbf{\textsc{VenusREM2} (ours)} & \checkmark & \checkmark & \checkmark & \textcolor{red!75!black}{\textbf{0.556}} & \textcolor{red!75!black}{\textbf{0.541}} & \textcolor{red!75!black}{\textbf{0.495}} & \textcolor{red!75!black}{\textbf{0.557}} & \textcolor{red!75!black}{\textbf{0.494}} & \textcolor{red!75!black}{\textbf{0.691}} \\
\midrule
1 & \textsc{AIDO Protein-RAG} (16B) \citep{sun2024aido} & --- & \checkmark & \checkmark & \textcolor[HTML]{7030A0}{\textbf{0.518}} & \textcolor[HTML]{7030A0}{\textbf{0.517}} & 0.426 & 0.522 & \textcolor[HTML]{7030A0}{\textbf{0.491}} & 0.635 \\
2 & \textsc{VenusREM} \citep{tan2025venusrem} & \checkmark & \checkmark & \checkmark & \textcolor[HTML]{7030A0}{\textbf{0.518}} & 0.495 & \textcolor[HTML]{7030A0}{\textbf{0.454}} & \textcolor[HTML]{7030A0}{\textbf{0.533}} & 0.459 & \textbf{0.650} \\
3 & \textsc{ProSST} ($K{=}2048$) \citep{li2024prosst} & \checkmark & \checkmark & --- & \textbf{0.507} & 0.476 & \textbf{0.445} & \textbf{0.530} & 0.431 & \textcolor[HTML]{7030A0}{\textbf{0.653}} \\
5 & \textsc{S3F-MSA} \citep{zhang2024s3f} & --- & \checkmark & \checkmark & 0.496 & \textbf{0.502} & 0.440 & 0.479 & 0.477 & 0.581 \\
8 & \textsc{Protriever} \citep{notin2025protriever} & --- & --- & \checkmark & 0.479 & 0.487 & 0.396 & 0.496 & \textbf{0.479} & 0.537 \\
9 & \textsc{ESCOTT} \citep{tekpinar2025prescott} & --- & \checkmark & \checkmark & 0.476 & 0.499 & 0.389 & 0.468 & 0.466 & 0.557 \\
12 & \textsc{PoET} (200M) \citep{truong2023poet} & --- & --- & \checkmark & 0.470 & 0.494 & 0.396 & 0.466 & 0.475 & 0.519 \\
14 & \textsc{ESM3} open (1.4B) \citep{hayes2025esm3} & \checkmark & \checkmark & --- & 0.466 & 0.430 & 0.400 & 0.470 & 0.389 & 0.641 \\
15 & \textsc{RSALOR} \citep{tsishyn2025rsalor} & --- & \checkmark & \checkmark & 0.465 & 0.479 & 0.416 & 0.427 & 0.426 & 0.575 \\
16 & \textsc{VespaG} \citep{marquet2024vespag} & \checkmark & --- & --- & 0.458 & 0.493 & 0.370 & 0.456 & 0.437 & 0.533 \\
17 & \textsc{SaProt} (650M AF2) \citep{su2023saprot} & \checkmark & \checkmark & --- & 0.457 & 0.458 & 0.378 & 0.488 & 0.366 & 0.592 \\
18 & \textsc{TranceptEVE-L} \citep{notin2022trancepteve} & \checkmark & --- & \checkmark & 0.456 & 0.487 & 0.376 & 0.457 & 0.459 & 0.500 \\
20 & \textsc{GEMME} \citep{laine2019gemme} & --- & --- & \checkmark & 0.455 & 0.482 & 0.383 & 0.438 & 0.452 & 0.519 \\
23 & \textsc{ProtSSN} ensemble \citep{tan2025protssn} & \checkmark & \checkmark & --- & 0.449 & 0.466 & 0.366 & 0.449 & 0.396 & 0.568 \\
40 & \textsc{ESM-IF1} \citep{hsu2022esm-if1} & --- & \checkmark & --- & 0.422 & 0.368 & 0.389 & 0.407 & 0.324 & 0.624 \\
45 & \textsc{ESM-2} (650M) \citep{lin2023esm2} & \checkmark & --- & --- & 0.414 & 0.425 & 0.337 & 0.415 & 0.368 & 0.523 \\
47 & \textsc{ESM-1v} ensemble \citep{meier2021esm1v} & \checkmark & --- & --- & 0.407 & 0.420 & 0.320 & 0.429 & 0.386 & 0.477 \\
53 & \textsc{MIF-ST} \citep{mifst} & \checkmark & \checkmark & --- & 0.400 & 0.390 & 0.321 & 0.438 & 0.366 & 0.485 \\
57 & \textsc{ESM-1b} \citep{rives2021esm1b} & \checkmark & --- & --- & 0.394 & 0.428 & 0.287 & 0.406 & 0.349 & 0.500 \\
59 & \textsc{ProGen3} (3B) \citep{bhatnagar2025progen3} & \checkmark & --- & --- & 0.392 & 0.410 & 0.287 & 0.428 & 0.400 & 0.438 \\
71 & \textsc{RITA-XL} \citep{hesslow2022rita} & \checkmark & --- & --- & 0.373 & 0.366 & 0.302 & 0.414 & 0.384 & 0.398 \\
72 & \textsc{CARP} (640M) \citep{carp} & \checkmark & --- & --- & 0.369 & 0.395 & 0.273 & 0.397 & 0.366 & 0.412 \\
91 & \textsc{ProteinMPNN} \citep{dauparas2022proteinmpnn} & --- & \checkmark & --- & 0.257 & 0.197 & 0.163 & 0.198 & 0.164 & 0.565 \\
\bottomrule
\end{tabular}}%
\vspace{0.2em}

\parbox{\linewidth}{\fontsize{8.5}{10}\selectfont\textit{Source and notes.} Ranked rows are from \url{https://proteingym.org/benchmarks}; one entry per model series, using its highest official Average Spearman. Seq, Str, and Evo denote sequence, structure, and evolutionary inputs.}
\end{center}
\end{table}

\FloatBarrier
\subsection{Stratified analyses by taxonomy and MSA depth}
\label{app:stratified}
Table~\ref{tab:stratified_taxa_msa} tests whether the \textsc{VenusREM2} gain is confined to one taxonomic or MSA-depth bin. Virus assays gain $0.064$, the low-depth bin $0.055$, and the high-depth bin $0.011$. The larger low-depth gain shows that the aggregate improvement is not restricted to proteins with abundant homologs. These averages show which groups improve, but do not separate the effects of MSA fusion, background correction, and structural weighting.

Table~\ref{tab:leaderboard_full_mutation_depth} separates variants by the number of substitutions. The same six-member ensemble leads each of the five depth bins, extending the overall comparison to both single- and multiple-substitution variants.

\begin{table}[!htbp]
\caption{Complete zero-shot ProteinGym substitutions comparison for the same models as Table~\ref{tab:leaderboard_full}, split by official taxon and MSA-depth bins. MSA depth uses $N_{\mathrm{eff}}/L$: Low $<1$, Medium $[1,100)$, High $\geq 100$. Bold red, purple, and black mark the top 3 distinct scores per column at reported precision.}
\label{tab:leaderboard_full_taxon_msa}
\begin{center}
\fontsize{8.5}{10}\selectfont
\setlength{\tabcolsep}{1.5pt}
\renewcommand{\arraystretch}{1.08}
\resizebox{\textwidth}{!}{%
\begin{tabular}{@{}rlcccccccccc@{}}
\toprule
Rank & Model & Seq & Str & Evo & Human & Other euk. & Prok. & Virus & Low & Medium & High \\
\midrule
--- & \textbf{\textsc{VenusREM2} (ours)} & \checkmark & \checkmark & \checkmark & \textcolor{red!75!black}{\textbf{0.562}} & \textcolor{red!75!black}{\textbf{0.615}} & \textcolor{red!75!black}{\textbf{0.601}} & \textcolor{red!75!black}{\textbf{0.540}} & \textcolor{red!75!black}{\textbf{0.553}} & \textcolor{red!75!black}{\textbf{0.557}} & \textcolor{red!75!black}{\textbf{0.623}} \\
\midrule
1 & \textsc{AIDO Protein-RAG} (16B) \citep{sun2024aido} & --- & \checkmark & \checkmark & \textcolor[HTML]{7030A0}{\textbf{0.531}} & \textcolor[HTML]{7030A0}{\textbf{0.587}} & \textcolor[HTML]{7030A0}{\textbf{0.558}} & \textcolor[HTML]{7030A0}{\textbf{0.522}} & \textcolor[HTML]{7030A0}{\textbf{0.498}} & \textcolor[HTML]{7030A0}{\textbf{0.534}} & \textcolor[HTML]{7030A0}{\textbf{0.585}} \\
2 & \textsc{VenusREM} \citep{tan2025venusrem} & \checkmark & \checkmark & \checkmark & \textbf{0.529} & \textbf{0.582} & \textbf{0.549} & 0.492 & \textbf{0.495} & \textbf{0.524} & 0.577 \\
3 & \textsc{ProSST} ($K{=}2048$) \citep{li2024prosst} & \checkmark & \checkmark & --- & 0.516 & 0.573 & \textbf{0.549} & 0.454 & 0.465 & 0.507 & \textbf{0.580} \\
5 & \textsc{S3F-MSA} \citep{zhang2024s3f} & --- & \checkmark & \checkmark & 0.502 & 0.558 & 0.521 & 0.502 & 0.469 & 0.509 & 0.547 \\
8 & \textsc{Protriever} \citep{notin2025protriever} & --- & --- & \checkmark & 0.480 & 0.542 & 0.492 & \textbf{0.516} & 0.464 & 0.498 & 0.512 \\
9 & \textsc{ESCOTT} \citep{tekpinar2025prescott} & --- & \checkmark & \checkmark & 0.486 & 0.537 & 0.500 & 0.502 & 0.462 & 0.496 & 0.524 \\
12 & \textsc{PoET} (200M) \citep{truong2023poet} & --- & --- & \checkmark & 0.482 & 0.541 & 0.464 & 0.491 & 0.478 & 0.478 & 0.510 \\
14 & \textsc{ESM3} open (1.4B) \citep{hayes2025esm3} & \checkmark & \checkmark & --- & 0.480 & 0.548 & 0.530 & 0.406 & 0.397 & 0.466 & 0.575 \\
15 & \textsc{RSALOR} \citep{tsishyn2025rsalor} & --- & \checkmark & \checkmark & 0.473 & 0.530 & 0.496 & 0.477 & 0.467 & 0.468 & 0.529 \\
16 & \textsc{VespaG} \citep{marquet2024vespag} & \checkmark & --- & --- & 0.474 & 0.532 & 0.484 & 0.421 & 0.427 & 0.472 & 0.511 \\
17 & \textsc{SaProt} (650M AF2) \citep{su2023saprot} & \checkmark & \checkmark & --- & 0.478 & 0.529 & 0.514 & 0.320 & 0.394 & 0.446 & 0.546 \\
18 & \textsc{TranceptEVE-L} \citep{notin2022trancepteve} & \checkmark & --- & \checkmark & 0.473 & 0.513 & 0.455 & 0.461 & 0.436 & 0.472 & 0.490 \\
20 & \textsc{GEMME} \citep{laine2019gemme} & --- & --- & \checkmark & 0.469 & 0.516 & 0.467 & 0.472 & 0.446 & 0.474 & 0.493 \\
23 & \textsc{ProtSSN} ensemble \citep{tan2025protssn} & \checkmark & \checkmark & --- & 0.470 & 0.528 & 0.492 & 0.370 & 0.409 & 0.454 & 0.524 \\
40 & \textsc{ESM-IF1} \citep{hsu2022esm-if1} & --- & \checkmark & --- & 0.417 & 0.502 & 0.498 & 0.389 & 0.292 & 0.427 & 0.549 \\
45 & \textsc{ESM-2} (650M) \citep{lin2023esm2} & \checkmark & --- & --- & 0.457 & 0.486 & 0.458 & 0.261 & 0.338 & 0.409 & 0.513 \\
47 & \textsc{ESM-1v} ensemble \citep{meier2021esm1v} & \checkmark & --- & --- & 0.458 & 0.464 & 0.413 & 0.294 & 0.316 & 0.409 & 0.495 \\
53 & \textsc{MIF-ST} \citep{mifst} & \checkmark & \checkmark & --- & 0.399 & 0.413 & 0.457 & 0.405 & 0.371 & 0.404 & 0.453 \\
57 & \textsc{ESM-1b} \citep{rives2021esm1b} & \checkmark & --- & --- & 0.435 & 0.491 & 0.438 & 0.260 & 0.365 & 0.396 & 0.480 \\
59 & \textsc{ProGen3} (3B) \citep{bhatnagar2025progen3} & \checkmark & --- & --- & 0.410 & 0.433 & 0.392 & 0.414 & 0.325 & 0.407 & 0.453 \\
71 & \textsc{RITA-XL} \citep{hesslow2022rita} & \checkmark & --- & --- & 0.396 & 0.399 & 0.346 & 0.394 & 0.300 & 0.390 & 0.412 \\
72 & \textsc{CARP} (640M) \citep{carp} & \checkmark & --- & --- & 0.417 & 0.395 & 0.381 & 0.284 & 0.331 & 0.374 & 0.424 \\
91 & \textsc{ProteinMPNN} \citep{dauparas2022proteinmpnn} & --- & \checkmark & --- & 0.284 & 0.394 & 0.348 & 0.262 & 0.187 & 0.271 & 0.439 \\
\bottomrule
\end{tabular}}%
\vspace{0.2em}

\parbox{\linewidth}{\fontsize{8.5}{10}\selectfont\textit{Source and notes.} Ranked rows use the official ProteinGym substitutions summary; one entry per model series, matching Table~\ref{tab:leaderboard_full}. \textsc{VenusREM2} uses the six-checkpoint \textsc{ProSST} ensemble in Table~\ref{tab:stratified_taxa_msa}.}
\end{center}
\end{table}

\begin{table}[!ht]
\caption{\textsc{VenusREM2} ProteinGym Average Spearman by taxon and MSA depth, recomputed with the official evaluator. MSA depth uses $N_{\mathrm{eff}}/L$: Low $<1$, Medium $[1,100)$, High $\geq 100$. Raw denotes the original ensemble, and \textsc{\HarnessShortName{}} denotes the ensemble after applying the harness; $\Delta$ is the improvement.}
\label{tab:stratified_taxa_msa}
\centering
\small
\setlength{\tabcolsep}{8pt}
\begin{tabular}{llccc}
\toprule
Stratification & Group & Raw & \textsc{\HarnessShortName{}} & $\Delta$ \\
\midrule
Taxon & Human & 0.524 & 0.562 & +0.038 \\
 & Other eukaryotes & 0.596 & 0.615 & +0.019 \\
 & Prokaryote & 0.579 & 0.601 & +0.022 \\
 & Virus & 0.476 & 0.540 & +0.064 \\
\midrule
MSA depth & Low & 0.498 & 0.553 & +0.055 \\
 & Medium & 0.517 & 0.557 & +0.040 \\
 & High & 0.612 & 0.623 & +0.011 \\
\bottomrule
\end{tabular}
\end{table}

\begin{table}[!htbp]
\caption{Complete zero-shot ProteinGym substitutions comparison for the same models as Table~\ref{tab:leaderboard_full}, split by official mutation-depth bins. Depth is the number of amino-acid substitutions per variant; $5+$ denotes at least five substitutions. Bold red, purple, and black mark the top 3 distinct scores per column at reported precision.}
\label{tab:leaderboard_full_mutation_depth}
\begin{center}
% Match the rendered body size of Table 16; distribute spare width without enlarging the text.
\fontsize{8.1}{9.53}\selectfont
\setlength{\tabcolsep}{1.5pt}
\renewcommand{\arraystretch}{1.08}
\begin{tabular*}{\textwidth}{@{}rlccc@{\extracolsep{\fill}}ccccc@{}}
\toprule
Rank & Model & Seq & Str & Evo & 1 & 2 & 3 & 4 & $5+$ \\
\midrule
--- & \textbf{\textsc{VenusREM2} (ours)} & \checkmark & \checkmark & \checkmark & \textcolor{red!75!black}{\textbf{0.576}} & \textcolor{red!75!black}{\textbf{0.473}} & \textcolor{red!75!black}{\textbf{0.468}} & \textcolor{red!75!black}{\textbf{0.424}} & \textcolor{red!75!black}{\textbf{0.473}} \\
\midrule
1 & \textsc{AIDO Protein-RAG} (16B) \citep{sun2024aido} & --- & \checkmark & \checkmark & \textbf{0.527} & \textcolor[HTML]{7030A0}{\textbf{0.414}} & \textbf{0.419} & \textbf{0.394} & 0.414 \\
2 & \textsc{VenusREM} \citep{tan2025venusrem} & \checkmark & \checkmark & \checkmark & \textcolor[HTML]{7030A0}{\textbf{0.534}} & \textbf{0.397} & 0.355 & 0.322 & 0.368 \\
3 & \textsc{ProSST} ($K{=}2048$) \citep{li2024prosst} & \checkmark & \checkmark & --- & 0.521 & 0.394 & 0.317 & 0.277 & 0.332 \\
5 & \textsc{S3F-MSA} \citep{zhang2024s3f} & --- & \checkmark & \checkmark & 0.499 & 0.333 & 0.378 & 0.346 & 0.383 \\
8 & \textsc{Protriever} \citep{notin2025protriever} & --- & --- & \checkmark & 0.473 & 0.320 & \textcolor[HTML]{7030A0}{\textbf{0.427}} & 0.387 & 0.427 \\
9 & \textsc{ESCOTT} \citep{tekpinar2025prescott} & --- & \checkmark & \checkmark & 0.476 & 0.314 & 0.377 & 0.385 & \textbf{0.434} \\
12 & \textsc{PoET} (200M) \citep{truong2023poet} & --- & --- & \checkmark & 0.466 & 0.299 & \textbf{0.419} & \textcolor[HTML]{7030A0}{\textbf{0.395}} & 0.418 \\
14 & \textsc{ESM3} open (1.4B) \citep{hayes2025esm3} & \checkmark & \checkmark & --- & 0.487 & 0.337 & 0.307 & 0.285 & 0.363 \\
15 & \textsc{RSALOR} \citep{tsishyn2025rsalor} & --- & \checkmark & \checkmark & 0.474 & 0.292 & 0.369 & 0.392 & \textcolor[HTML]{7030A0}{\textbf{0.453}} \\
16 & \textsc{VespaG} \citep{marquet2024vespag} & \checkmark & --- & --- & 0.456 & 0.247 & 0.346 & 0.323 & 0.367 \\
17 & \textsc{SaProt} (650M AF2) \citep{su2023saprot} & \checkmark & \checkmark & --- & 0.458 & 0.313 & 0.275 & 0.268 & 0.336 \\
18 & \textsc{TranceptEVE-L} \citep{notin2022trancepteve} & \checkmark & --- & \checkmark & 0.446 & 0.278 & 0.349 & 0.331 & 0.382 \\
20 & \textsc{GEMME} \citep{laine2019gemme} & --- & --- & \checkmark & 0.448 & 0.276 & 0.330 & 0.341 & 0.414 \\
23 & \textsc{ProtSSN} ensemble \citep{tan2025protssn} & \checkmark & \checkmark & --- & 0.459 & 0.291 & 0.292 & 0.244 & 0.298 \\
40 & \textsc{ESM-IF1} \citep{hsu2022esm-if1} & --- & \checkmark & --- & 0.440 & 0.350 & 0.301 & 0.294 & 0.358 \\
45 & \textsc{ESM-2} (650M) \citep{lin2023esm2} & \checkmark & --- & --- & 0.422 & 0.248 & 0.205 & 0.163 & 0.218 \\
47 & \textsc{ESM-1v} ensemble \citep{meier2021esm1v} & \checkmark & --- & --- & 0.403 & 0.215 & 0.173 & 0.153 & 0.218 \\
53 & \textsc{MIF-ST} \citep{mifst} & \checkmark & \checkmark & --- & 0.431 & 0.263 & 0.327 & 0.298 & 0.298 \\
57 & \textsc{ESM-1b} \citep{rives2021esm1b} & \checkmark & --- & --- & 0.383 & 0.226 & 0.172 & 0.150 & 0.270 \\
59 & \textsc{ProGen3} (3B) \citep{bhatnagar2025progen3} & \checkmark & --- & --- & 0.388 & 0.179 & 0.226 & 0.204 & 0.256 \\
71 & \textsc{RITA-XL} \citep{hesslow2022rita} & \checkmark & --- & --- & 0.357 & 0.141 & 0.148 & 0.160 & 0.233 \\
72 & \textsc{CARP} (640M) \citep{carp} & \checkmark & --- & --- & 0.391 & 0.213 & 0.172 & 0.159 & 0.162 \\
91 & \textsc{ProteinMPNN} \citep{dauparas2022proteinmpnn} & --- & \checkmark & --- & 0.292 & 0.260 & 0.174 & 0.183 & 0.278 \\
\bottomrule
\end{tabular*}
\vspace{0.2em}

\parbox{\linewidth}{\fontsize{8.5}{10}\selectfont\textit{Source and notes.} Ranked rows use the official ProteinGym substitutions summary; one entry per model series, matching Table~\ref{tab:leaderboard_full}. \textsc{VenusREM2} uses the six-checkpoint \textsc{ProSST} ensemble in Table~\ref{tab:stratified_taxa_msa}.}
\end{center}
\end{table}

\FloatBarrier

\section{Shared model preferences and local MSA preferences}
\label{app:bias_details}

We compare amino-acid preferences across positions, proteins, and model configurations. The profile analyses use cached predictions from 37 configurations, expanded to 49 for the ProteinGym and VenusMutHub background comparisons. Prediction-performance comparisons use all 71 configurations.

\newpage
\subsection{Site profiles and their correlations}
\label{app:profile_separation}
\reviewchange{Each site profile contains scores for the 20 standard amino acids. We compare its MSA scores with two reference vectors: log UniProt amino-acid frequencies and PFM scores averaged across retained positions in the same protein. Pearson correlation is calculated across the available amino-acid entries.} These site-level comparisons use \textsc{ESM-2} 650M with unmasked WT scoring.

For ProteinGym, we reconstruct the PFM and MSA profiles from cached original-model and fixed-$\alpha{=}0.8$ fusion scores. The WT entry is zero, and a position must have at least 15 finite PFM entries. For VenusMutHub and \ViroBenchmarkName{}, we use cached score matrices and require a standard WT residue and 20 finite entries in both profiles. Assays must provide at least 20 such positions, and sites with undefined correlations are removed. We retain the assay with the most eligible positions for each ProteinGym protein ID or each identical WT sequence in the other benchmarks. This gives 27{,}763 positions from 148 ProteinGym proteins, 148{,}983 from 527 VenusMutHub WT sequences, and 29{,}416 from 52 \ViroBenchmarkName{} WT sequences.

\begin{figure}[!htbp]
\centering
\includegraphics[width=0.40\textwidth]{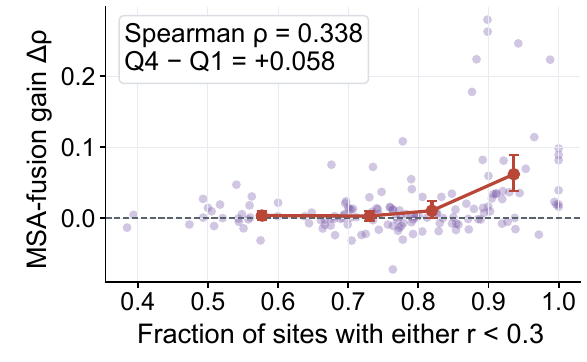}
\caption{\reviewchange{MSA-fusion gain versus the fraction of sites with Pearson $r<0.3$ between their MSA profile and at least one reference (protein-average PFM profile or log UniProt abundance), across} $148$ ProteinGym proteins. Red: quartile means and 95\% bootstrap intervals.}
\label{fig:separation_gain}
\end{figure}

Both correlations are at least $0.3$ at $19.7\%$, $24.8\%$, and $38.1\%$ of retained sites, respectively. \reviewchange{We define the separation index as the fraction of sites for which either correlation is below $0.3$.} Figure~\ref{fig:separation_gain} relates this fraction to the Spearman gain from MSA fusion. Of 217 ProteinGym assays, 177 meet the site-profile requirement and represent 148 proteins. We average assay gains within each protein. Intervals use 10{,}000 protein-level bootstrap draws, and partial rank correlation adjusts for sequence length, $\log_{10}$ MSA depth, MSA occupancy, and original-model Spearman.

The correlation between separation and MSA-fusion gain is $0.338$ (95\% interval $0.179$--$0.481$). The highest-minus-lowest quartile difference in gain is $0.058$ (interval $0.033$--$0.086$), and the adjusted correlation is $0.261$. These results support Section~\ref{sec:results_scales}.

\subsection{Model background preferences}
\label{app:native_background}
\begin{figure}[!ht]
\begin{center}
\includegraphics[width=\textwidth]{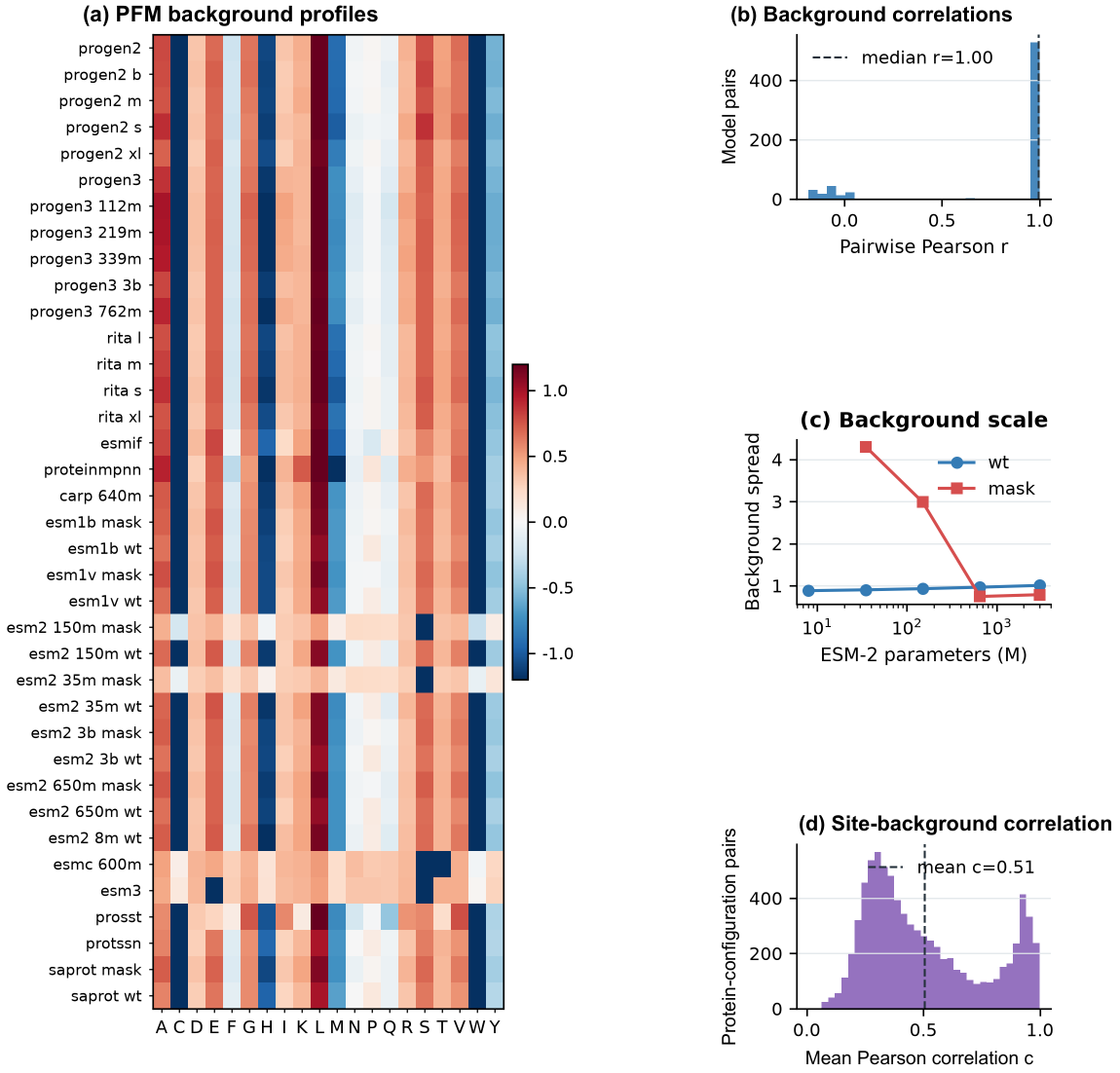}
\end{center}
\caption{Model background preferences in 37 model configurations. \textbf{(a)} Amino-acid background profiles for each configuration. \textbf{(b)} Pairwise Pearson correlations among those profiles. \textbf{(c)} Variation in ESM-2 background scores under masked and WT scoring. \textbf{(d)} Mean Pearson correlation $c$ (Equation~\ref{eq:gate}) between the score vector at each position and the background vector, shown for each protein and configuration.}
\label{fig:bias}
\end{figure}

Figure~\ref{fig:bias} extends the comparison in Figure~\ref{fig:teaser} to 37 configurations with cached background profiles. It compares model amino-acid preferences, their pairwise correlations and score scales, and the correlations between site-specific scores and each protein's model background. The profiles have median pairwise Pearson $r=0.995$; the correlation between the average model preferences and corpus frequencies is $0.993$. The mean correlation $c$ (Equation~\ref{eq:gate}) ranges from $0.01$ to $0.99$ across proteins and configurations, motivating a gate that adapts to the query protein. \textsc{ESM3} differs most from the shared pattern, consistent with its multimodal pretraining \citep{hayes2025esm3}.

\subsection{Checks across benchmarks}
\label{app:replications}
Table~\ref{tab:preference_summary} distinguishes four comparisons. Pairwise is the median Pearson correlation between model background profiles, using 49 configurations for ProteinGym and VenusMutHub and 37 for \ViroBenchmarkName{}. Corpus compares the median standardized model profile with log UniProt frequencies. Model--MSA compares mean MSA composition with the 37-configuration model background, reporting the median correlation within an assay or WT sequence and the correlation after pooling their amino-acid entries. The final columns count retained sites and the fraction with both reference correlations at least $0.3$, as defined in Appendix~\ref{app:profile_separation}.

\begin{table}[H]
\caption{Correlations among model backgrounds across benchmarks and their differences from local MSA preferences. Retained sites meet the dataset-specific conditions in Appendix~\ref{app:profile_separation}.}
\label{tab:preference_summary}
\centering
\small
\setlength{\tabcolsep}{4pt}
\begin{tabular}{lrrrrrr}
\toprule
Dataset & Pairwise & Corpus & \multicolumn{2}{c}{Model--MSA} & Retained sites & Both $r\geq0.3$ \\
\cmidrule(lr){4-5}
& $r$ & $r$ & Median $r$ & Pooled $r$ & $N$ & (\%) \\
\midrule
ProteinGym & 0.995 & 0.991 & 0.965 & 0.758 & 27{,}763 & 19.7 \\
VenusMutHub & 0.996 & 0.989 & 0.964 & 0.863 & 148{,}983 & 24.8 \\
\ViroBenchmarkName{} & 0.990 & 0.885 & 0.976 & 0.621 & 29{,}416 & 38.1 \\
\bottomrule
\end{tabular}
\end{table}

Models have similar background preferences in each benchmark. Correlations between model and MSA profiles are lower when data are pooled across assays, and most sites have a correlation below $0.3$ with at least one reference. Figures~\ref{fig:proteingym_analysis}--\ref{fig:virogym_analysis} retain the dataset-specific distributions and family-level outcomes.

\label{app:dataset_preference}
\subsection{ProteinGym analysis}
\label{app:proteingym_replication}
\begin{figure}[!ht]
\centering
\includegraphics[width=\textwidth]{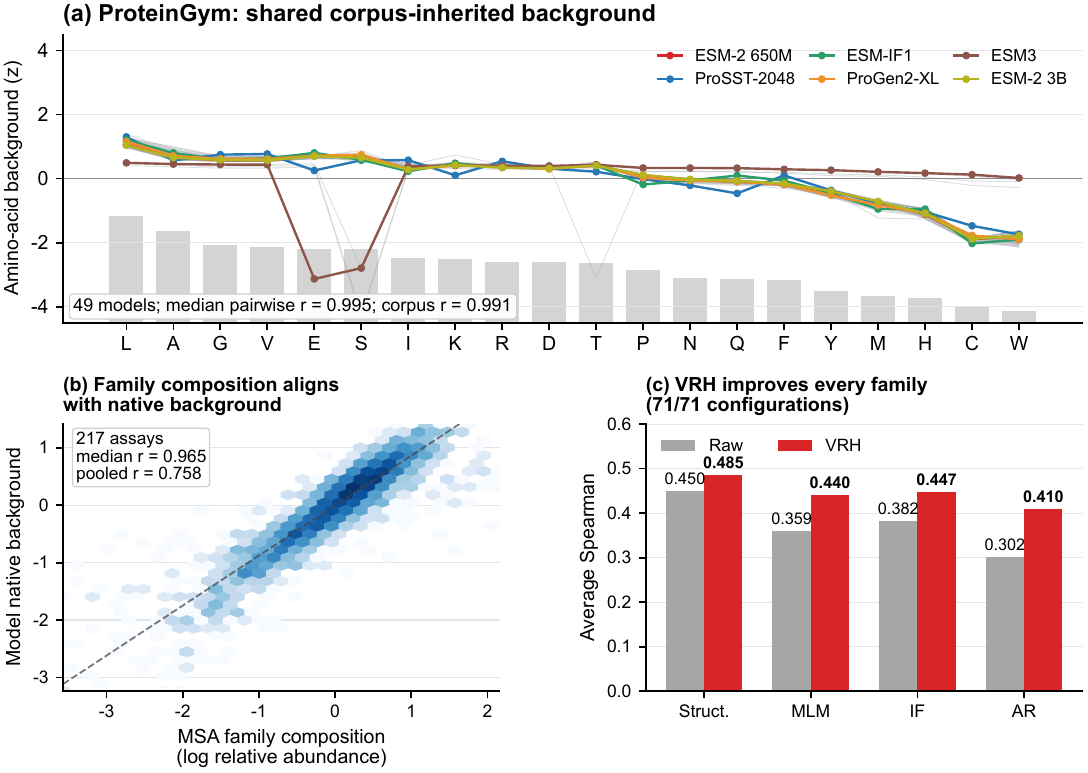}
\caption{ProteinGym analysis. \textbf{(a)} Profiles from 49 configurations and UniProt abundance (gray bars). \textbf{(b)} Average background preferences across the 37-configuration subset versus MSA abundance (central 99\%; correlations use all $217\times20$ pairs). \textbf{(c)} Spearman before and after applying the harness, grouped by model family across 71 configurations.}
\label{fig:proteingym_analysis}
\end{figure}

Figure~\ref{fig:proteingym_analysis} compares protein-average profiles with preferences at individual sites. Only $19.7\%$ of retained sites have correlations of at least $0.3$ with both references, despite strong correlations between protein averages (Table~\ref{tab:preference_summary}). All four model families improve in prediction performance, with family means reported in Appendix~\ref{app:family_results}.

\subsection{VenusMutHub analysis}
\label{app:vmh}
\begin{figure}[!ht]
\centering
\includegraphics[width=\textwidth]{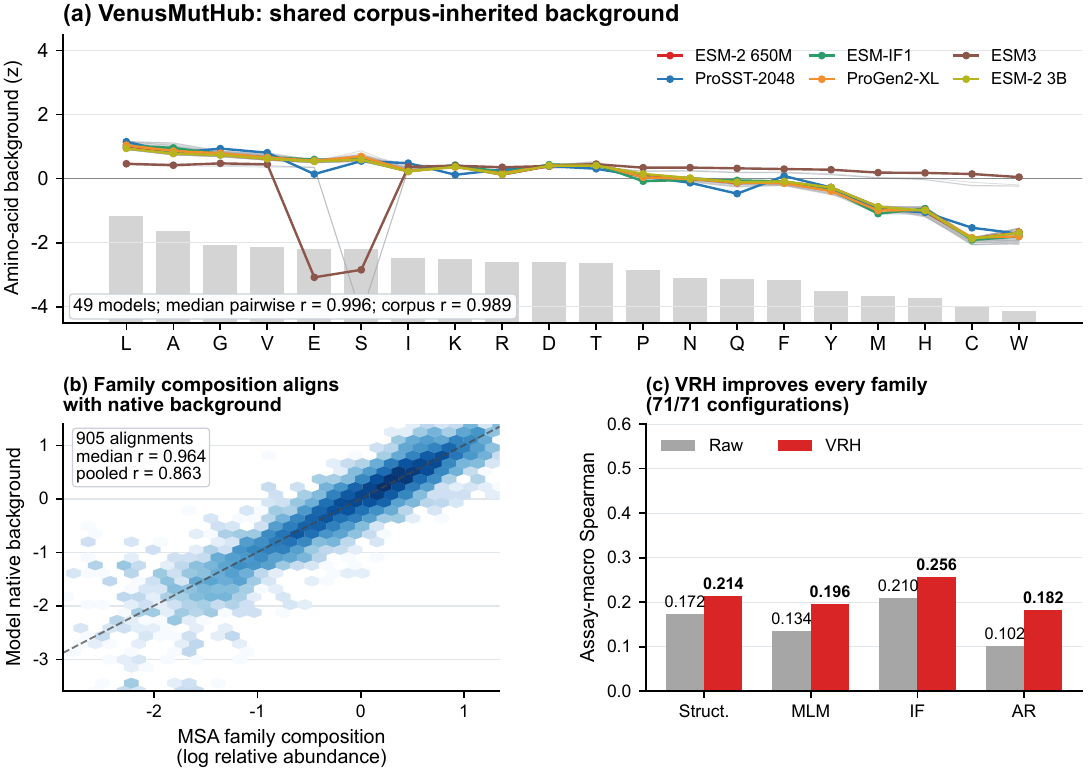}
\caption{VenusMutHub analysis. \textbf{(a)} Profiles from 49 configurations and UniProt abundance (gray bars). \textbf{(b)} Average background preferences across the 37-configuration subset versus MSA abundance (central 99\%; correlations use all $905\times20$ pairs). \textbf{(c)} Spearman before and after applying the harness, grouped by model family across 71 configurations.}
\label{fig:vmh_analysis}
\end{figure}

VenusMutHub reproduces the separation between shared model backgrounds and local family preferences: only $24.8\%$ of retained sites have correlations of at least $0.3$ with both references (Figure~\ref{fig:vmh_analysis}; Table~\ref{tab:preference_summary}). Thus the ProteinGym pattern extends to this broader assay collection. Family-level performance in Appendix~\ref{app:family_results} provides the corresponding prediction comparison.

\FloatBarrier
\subsection{\ViroBenchmarkName{} analysis}
\label{app:virogym_replication}
\begin{figure}[!ht]
\centering
\includegraphics[width=\textwidth]{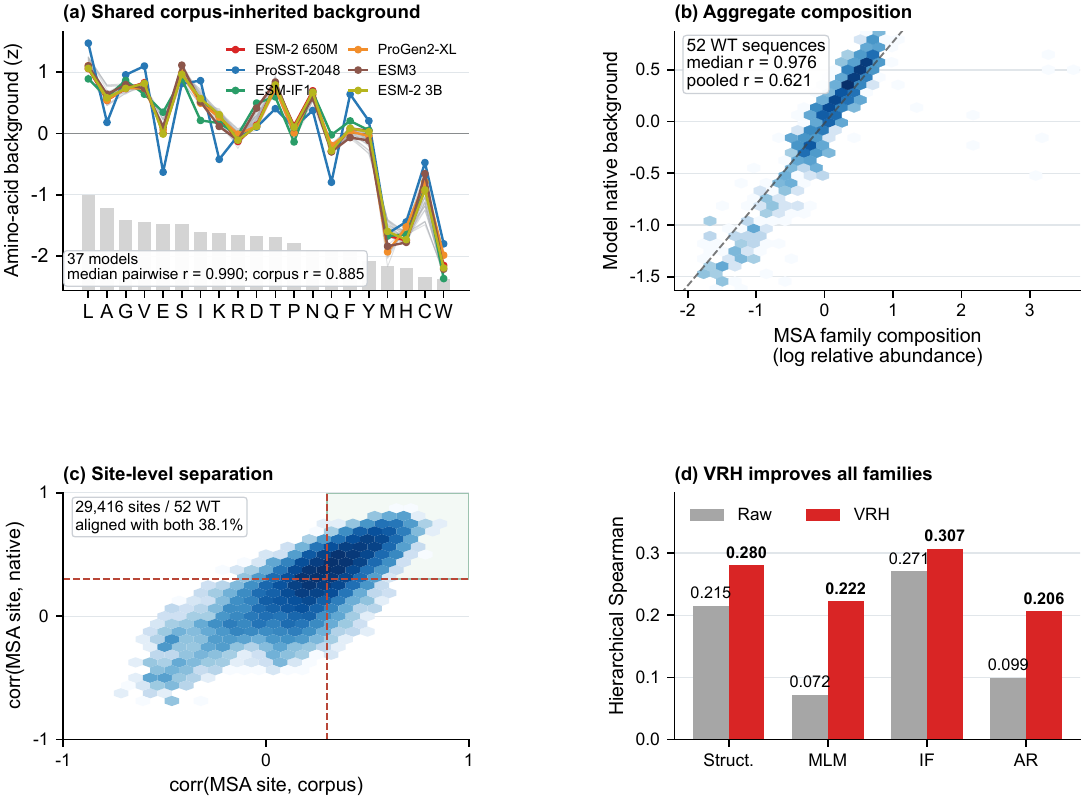}
\caption{\ViroBenchmarkName{} analysis. \textbf{(a)} Profiles from 37 configurations and UniProt abundance. \textbf{(b)} Their average background preferences versus MSA abundance. \textbf{(c)} MSA site preferences versus corpus and model backgrounds. \textbf{(d)} Spearman before and after applying the harness, grouped by model family across 71 configurations.}
\label{fig:virogym_analysis}
\end{figure}

Figure~\ref{fig:virogym_analysis} shows the lowest pooled correlation between model and MSA profiles ($r=0.621$) despite a high assay-level median, with $38.1\%$ of retained sites above the $0.3$ threshold for both references. The median and pooled statistics summarize different levels of variation; high correlations within proteins can therefore coexist with a lower correlation across the collection. Appendix~\ref{app:family_results} reports the gains across model families.

\section{Results across model families}
\label{app:family_results}

All ProteinGym results use one versioned evaluation pipeline. Of 72 configurations, 71 have results before and after applying the harness on all 217 assays. \textsc{VenusREM} was not evaluated with the harness and is excluded from these paired comparisons. Every model family improves, with the largest gains in autoregressive models and further gains in structure-aware encoders (Table~\ref{tab:proteingym}). The \textsc{ProSST} ensemble improves on all five metrics (Table~\ref{tab:prosst_metrics}).

\begin{table}[!ht]
\caption{ProteinGym Average Spearman by model family across 71 configurations and 217 assays. Raw denotes original predictions, and \textsc{\HarnessShortName{}} denotes predictions after applying the harness. $N$ counts configurations; mean and median change summarize their improvements. Wins / $N$ counts configurations with positive change.}
\label{tab:proteingym}
\centering
\small
\renewcommand{\arraystretch}{1.08}
\setlength{\tabcolsep}{4pt}
\begin{tabular*}{\textwidth}{@{\extracolsep{\fill}}lcccccc@{}}
\toprule
Family & $N$ & Raw & \textsc{\HarnessShortName{}} & Mean $\Delta$ & Median $\Delta$ & Wins / $N$ \\
\midrule
Structure-aware & 26 & 0.450 & 0.485 & +0.035 & +0.029 & 26 / 26 \\
\specialrule{0.2pt}{1.2pt}{1.2pt}
Masked sequence LM & 20 & 0.359 & 0.440 & +0.081 & +0.062 & 20 / 20 \\
\specialrule{0.2pt}{1.2pt}{1.2pt}
Inverse folding & 10 & 0.382 & 0.447 & +0.065 & +0.070 & 10 / 10 \\
\specialrule{0.2pt}{1.2pt}{1.2pt}
Autoregressive LM & 15 & 0.302 & 0.410 & +0.108 & +0.098 & 15 / 15 \\
\bottomrule
\end{tabular*}
\end{table}

\begin{table}[!ht]
\caption{\textsc{VenusREM2} ProteinGym metrics, reported as mean $\pm$ standard deviation over 217 assays. These means are not official Average Spearman. Raw and \textsc{\HarnessShortName{}} denote the six-model \textsc{ProSST} ensemble before and after applying the harness; $\Delta$ is the improvement. Top-recall denotes top-variant recall.}
\label{tab:prosst_metrics}
\centering
\small
\setlength{\tabcolsep}{9pt}
\begin{tabular}{lccc}
\toprule
Metric & Raw & \textsc{\HarnessShortName{}} & $\Delta$ \\
\midrule
Spearman & 0.541 $\pm$ 0.180 & 0.575 $\pm$ 0.159 & +0.034 \\
NDCG & 0.794 $\pm$ 0.138 & 0.813 $\pm$ 0.130 & +0.019 \\
AUC & 0.796 $\pm$ 0.100 & 0.815 $\pm$ 0.091 & +0.019 \\
MCC & 0.422 $\pm$ 0.164 & 0.450 $\pm$ 0.151 & +0.028 \\
Top-recall & 0.261 $\pm$ 0.131 & 0.270 $\pm$ 0.131 & +0.008 \\
\bottomrule
\end{tabular}
\end{table}

The aggregate gain is not confined to one of the five ProteinGym biological categories. Across those categories and $71$ configurations, all $355$ configuration--category comparisons improve in Spearman. Table~\ref{tab:function_types} reports the corresponding family means; Stability shows the largest overall increase, gaining $0.095$ from $0.489$ to $0.584$.

\begin{table}[!ht]
\caption{Mean ProteinGym Average Spearman by biological category and model family across the 71 configurations. Counts in the header are the number of configurations in that family. Structure $=$ structure-aware; Masked LM $=$ masked sequence LM; AR LM $=$ autoregressive LM.}
\label{tab:function_types}
\begin{center}
\scriptsize
\setlength{\tabcolsep}{2.0pt}
\resizebox{\textwidth}{!}{%
\begin{tabular}{lcccccccccc}
\toprule
Category & \multicolumn{2}{c}{Structure (26)} & \multicolumn{2}{c}{Masked LM (20)} & \multicolumn{2}{c}{Inverse folding (10)} & \multicolumn{2}{c}{AR LM (15)} & \multicolumn{2}{c}{All (71)} \\
\cmidrule(lr){2-3}\cmidrule(lr){4-5}\cmidrule(lr){6-7}\cmidrule(lr){8-9}\cmidrule(lr){10-11}
& Raw & \textsc{\HarnessShortName{}} & Raw & \textsc{\HarnessShortName{}} & Raw & \textsc{\HarnessShortName{}} & Raw & \textsc{\HarnessShortName{}} & Raw & \textsc{\HarnessShortName{}} \\
\midrule
Activity & 0.442 & 0.486 & 0.371 & 0.453 & 0.318 & 0.414 & 0.321 & 0.434 & 0.379 & 0.456 \\
Binding & 0.381 & 0.407 & 0.288 & 0.353 & 0.327 & 0.384 & 0.251 & 0.313 & 0.320 & 0.369 \\
Expression & 0.459 & 0.487 & 0.369 & 0.433 & 0.382 & 0.429 & 0.334 & 0.400 & 0.397 & 0.445 \\
Organismal Fitness & 0.378 & 0.423 & 0.295 & 0.394 & 0.294 & 0.373 & 0.334 & 0.397 & 0.334 & 0.402 \\
Stability & 0.589 & 0.623 & 0.474 & 0.567 & 0.588 & 0.635 & 0.272 & 0.507 & 0.489 & 0.584 \\
\bottomrule
\end{tabular}}
\end{center}
\end{table}

VenusMutHub repeats the same family order (Table~\ref{tab:venusmuthub}): every family improves, and autoregressive models again gain the most. Table~\ref{tab:bootstrap} summarizes the paired benchmark units: UniProt IDs for ProteinGym, wild-type sequence clusters for VenusMutHub, and groups of assays with the same phenotype and WT protein for \ViroBenchmarkName{}.

\begin{table}[!ht]
\caption{VenusMutHub mean assay Spearman by model family across 71 paired configurations and 905 assays. Column definitions match Table~\ref{tab:proteingym}.}
\label{tab:venusmuthub}
\begin{center}
\small
\renewcommand{\arraystretch}{1.08}
\setlength{\tabcolsep}{4pt}
\begin{tabular*}{\textwidth}{@{\extracolsep{\fill}}lcccccc@{}}
\toprule
Family & $N$ & Raw & \textsc{\HarnessShortName{}} & \shortstack{Mean\\change} & \shortstack{Median\\change} & Wins / $N$ \\
\midrule
Structure-aware & 26 & 0.172 & 0.214 & +0.042 & +0.032 & 26 / 26 \\
Masked sequence LM & 20 & 0.134 & 0.196 & +0.062 & +0.050 & 20 / 20 \\
Inverse folding & 10 & 0.210 & 0.256 & +0.046 & +0.049 & 10 / 10 \\
Autoregressive LM & 15 & 0.102 & 0.182 & +0.080 & +0.070 & 15 / 15 \\
\bottomrule
\end{tabular*}
\end{center}
\end{table}

\ViroBenchmarkName{} extends the comparison to a virus-focused distribution with no ProteinGym assay overlap (Table~\ref{tab:virogym_family}). The largest mean gain occurs for masked sequence models, while every configuration improves in every family.

\begin{table}[!ht]
\caption{\ViroBenchmarkName{} hierarchically aggregated Spearman by model family across 71 paired configurations and 89 assays, excluding wild-type controls. Column definitions match Table~\ref{tab:proteingym}.}
\label{tab:virogym_family}
\begin{center}
\small
\renewcommand{\arraystretch}{1.08}
\setlength{\tabcolsep}{4pt}
\begin{tabular*}{\textwidth}{@{\extracolsep{\fill}}lcccccc@{}}
\toprule
Family & $N$ & Raw & \textsc{\HarnessShortName{}} & \shortstack{Mean\\change} & \shortstack{Median\\change} & Wins / $N$ \\
\midrule
Structure-aware & 26 & 0.215 & 0.280 & +0.065 & +0.063 & 26 / 26 \\
Masked sequence LM & 20 & 0.072 & 0.222 & +0.151 & +0.147 & 20 / 20 \\
Inverse folding & 10 & 0.271 & 0.307 & +0.036 & +0.036 & 10 / 10 \\
Autoregressive LM & 15 & 0.099 & 0.206 & +0.107 & +0.110 & 15 / 15 \\
\bottomrule
\end{tabular*}
\end{center}
\end{table}

\begin{table}[!ht]
\caption{Mean Spearman improvement after applying \textsc{\HarnessName{}} across 71 configurations. $\Delta$ follows each benchmark's primary aggregation: Average Spearman for ProteinGym, mean assay Spearman for VenusMutHub, and hierarchical Spearman for \ViroBenchmarkName{}. Units follow the paired bootstrap in Appendix~\ref{app:metric_defs}; Wins / $N$ counts positive changes.}
\label{tab:bootstrap}
\begin{center}
\small
\setlength{\tabcolsep}{4pt}
\begin{tabular}{llccc}
\toprule
Benchmark & Resampling unit & Units & Mean $\Delta$ & Wins / $N$ \\
\midrule
ProteinGym & UniProt ID & 186 & +0.068 & 71 / 71 \\
VenusMutHub & wild-type sequence & 527 & +0.056 & 71 / 71 \\
\ViroBenchmarkName{} & phenotype--WT group & 52 & +0.094 & 71 / 71 \\
\bottomrule
\end{tabular}
\end{center}
\end{table}

\newpage
\section{Component ablations and validation}
\label{app:staged_audits}

\subsection{Ablations at each stage}
\label{app:stage_ablations}

\paragraph{Stages across model families.}
For configurations with complete intermediate scores, we evaluate the original model, adaptive MSA fusion, gated background correction, RSA weighting, and pLDDT weighting in sequence. We also test background correction without the gate. Component choices are selected on 21 validation assays and then fixed. These assays form fold~8 of a ten-fold partition obtained by shuffling sorted assay identifiers with seed~42. Appendix~\ref{app:staged_tables} reports all configurations. Gated correction raises the aggregate mean on all three benchmarks, and RSA weighting improves all $71$ configurations on each benchmark.

\paragraph{\textsc{ProSST}-2048 component ablation.}
Table~\ref{tab:ablation} starts each correction from the same adaptively fused PFM and MSA scores and uses $1-\alpha$ as the background-correction coefficient. On the random 21-assay validation split, retrieval contributes $0.0201$ of the total $0.0276$ improvement; gating adds $0.0014$, and RSA and pLDDT weighting add a further $0.0061$, yielding $0.5849$ from an original-model value of $0.5573$. The held-out 196-assay ProteinGym test set follows the same order under official aggregation: retrieval contributes $0.0169$, gating $0.0040$, RSA weighting $0.0045$, and pLDDT weighting $0.0011$, producing a final score of $0.5337$ from $0.5072$. The ungated row tests correction without the gate; it is an alternative to the gated row, not an additional stage.

We also compare the gate across all model configurations. Relative to adaptive MSA fusion alone, gated correction improves $71/71$, $62/71$, and $71/71$ configurations on ProteinGym, VenusMutHub, and \ViroBenchmarkName{}, respectively; relative to ungated correction, the corresponding counts are $71/71$, $37/71$, and $68/71$.

\subsection{\reviewchange{Statistical analysis of the \textsc{ProSST}-2048 ablation}}
\label{app:prosst_component_statistics}

\reviewchange{Table~\ref{tab:prosst_component_statistics} provides nine paired statistical comparisons for the configurations in Table~\ref{tab:ablation}, following Appendix~\ref{app:component_statistics}.}

\begin{table}[!htbp]
\begingroup
\ifreviewchanges\color{red}\fi
\caption{\reviewchange{Paired component comparisons for \textsc{ProSST}-2048. Columns report $\Delta$, its 95\% confidence interval (CI), and the two-sided $p$-value (Appendix~\ref{app:component_statistics}); bold $p$-values denote $p<0.05$. $M$: adaptive MSA fusion; $C$: gated background correction; $R$: RSA; $P$: pLDDT; $\mathrm{Full}=M{+}C{+}R{+}P$. $M_{0.8}$ fixes $\alpha=0.8$; $C_{\mathrm{ungated}}$ fixes $\kappa=1$.}}
\label{tab:prosst_component_statistics}
\centering
\footnotesize
\setlength{\tabcolsep}{4pt}
\renewcommand{\arraystretch}{1.06}
\begin{tabular*}{\textwidth}{@{\extracolsep{\fill}}llrrrr@{}}
\toprule
\multirow{2}{*}{Comparison} & \multirow{2}{*}{Score difference} & \multirow{2}{*}{$\Delta$} & \multicolumn{2}{c}{95\% CI} & \multirow{2}{*}{$p$} \\
\cmidrule(lr){4-5}
 & & & Lower & Upper & \\
\midrule
\multicolumn{6}{@{}l}{\textbf{ProteinGym validation (21 assays)}} \\
\specialrule{0.3pt}{1.5pt}{1.5pt}
Adaptive MSA fusion & $M - (\mathrm{Raw})$ & $+0.02016$ & $0.00778$ & $0.03414$ & $\boldsymbol{0.00645}$ \\
\specialrule{0.2pt}{1.0pt}{1.0pt}
Adaptive vs. fixed fusion & $M - (M_{0.8})$ & $+0.00327$ & $-0.00080$ & $0.00742$ & $0.1466$ \\
\specialrule{0.2pt}{1.0pt}{1.0pt}
Background correction & $M{+}C - (M)$ & $+0.00138$ & $-0.00216$ & $0.00457$ & $0.4575$ \\
\specialrule{0.2pt}{1.0pt}{1.0pt}
Gated vs. ungated & $M{+}C - (M{+}C_{\mathrm{ungated}})$ & $+0.01478$ & $0.00587$ & $0.02593$ & $\boldsymbol{0.0025}$ \\
\specialrule{0.2pt}{1.0pt}{1.0pt}
RSA weighting & $M{+}C{+}R - (M{+}C)$ & $+0.00382$ & $-0.00176$ & $0.00919$ & $0.1996$ \\
\specialrule{0.2pt}{1.0pt}{1.0pt}
pLDDT weighting & $M{+}C{+}P - (M{+}C)$ & $+0.00443$ & $-0.00013$ & $0.00988$ & $0.1094$ \\
\specialrule{0.2pt}{1.0pt}{1.0pt}
pLDDT after RSA & $\mathrm{Full} - (M{+}C{+}R)$ & $+0.00221$ & $-0.00155$ & $0.00668$ & $0.3234$ \\
\specialrule{0.2pt}{1.0pt}{1.0pt}
RSA after pLDDT & $\mathrm{Full} - (M{+}C{+}P)$ & $+0.00160$ & $-0.00343$ & $0.00666$ & $0.5489$ \\
\specialrule{0.2pt}{1.0pt}{1.0pt}
Full vs. MSA fusion & $\mathrm{Full} - (M)$ & $+0.00741$ & $-0.00214$ & $0.01579$ & $0.1301$ \\
\midrule
\multicolumn{6}{@{}l}{\textbf{ProteinGym full (217 assays)}} \\
\specialrule{0.3pt}{1.5pt}{1.5pt}
Adaptive MSA fusion & $M - (\mathrm{Raw})$ & $+0.01764$ & $0.01385$ & $0.02145$ & $\boldsymbol{<0.0001}$ \\
\specialrule{0.2pt}{1.0pt}{1.0pt}
Adaptive vs. fixed fusion & $M - (M_{0.8})$ & $+0.00407$ & $0.00230$ & $0.00585$ & $\boldsymbol{<0.0001}$ \\
\specialrule{0.2pt}{1.0pt}{1.0pt}
Background correction & $M{+}C - (M)$ & $+0.00384$ & $0.00196$ & $0.00565$ & $\boldsymbol{<0.0001}$ \\
\specialrule{0.2pt}{1.0pt}{1.0pt}
Gated vs. ungated & $M{+}C - (M{+}C_{\mathrm{ungated}})$ & $+0.00651$ & $0.00312$ & $0.01055$ & $\boldsymbol{0.0003}$ \\
\specialrule{0.2pt}{1.0pt}{1.0pt}
RSA weighting & $M{+}C{+}R - (M{+}C)$ & $+0.00447$ & $0.00170$ & $0.00723$ & $\boldsymbol{0.0008}$ \\
\specialrule{0.2pt}{1.0pt}{1.0pt}
pLDDT weighting & $M{+}C{+}P - (M{+}C)$ & $+0.00448$ & $0.00246$ & $0.00677$ & $\boldsymbol{<0.0001}$ \\
\specialrule{0.2pt}{1.0pt}{1.0pt}
pLDDT after RSA & $\mathrm{Full} - (M{+}C{+}R)$ & $+0.00137$ & $0.00002$ & $0.00285$ & $0.0647$ \\
\specialrule{0.2pt}{1.0pt}{1.0pt}
RSA after pLDDT & $\mathrm{Full} - (M{+}C{+}P)$ & $+0.00136$ & $-0.00101$ & $0.00368$ & $0.2605$ \\
\specialrule{0.2pt}{1.0pt}{1.0pt}
Full vs. MSA fusion & $\mathrm{Full} - (M)$ & $+0.00968$ & $0.00516$ & $0.01416$ & $\boldsymbol{<0.0001}$ \\
\midrule
\multicolumn{6}{@{}l}{\textbf{VenusMutHub (905 assays)}} \\
\specialrule{0.3pt}{1.5pt}{1.5pt}
Adaptive MSA fusion & $M - (\mathrm{Raw})$ & $+0.02106$ & $0.01123$ & $0.03118$ & $\boldsymbol{<0.0001}$ \\
\specialrule{0.2pt}{1.0pt}{1.0pt}
Adaptive vs. fixed fusion & $M - (M_{0.8})$ & $-0.00051$ & $-0.00378$ & $0.00291$ & $0.7676$ \\
\specialrule{0.2pt}{1.0pt}{1.0pt}
Background correction & $M{+}C - (M)$ & $+0.00331$ & $-0.00095$ & $0.00762$ & $0.1291$ \\
\specialrule{0.2pt}{1.0pt}{1.0pt}
Gated vs. ungated & $M{+}C - (M{+}C_{\mathrm{ungated}})$ & $+0.00433$ & $-0.00500$ & $0.01359$ & $0.363$ \\
\specialrule{0.2pt}{1.0pt}{1.0pt}
RSA weighting & $M{+}C{+}R - (M{+}C)$ & $+0.01297$ & $0.00576$ & $0.02046$ & $\boldsymbol{0.00025}$ \\
\specialrule{0.2pt}{1.0pt}{1.0pt}
pLDDT weighting & $M{+}C{+}P - (M{+}C)$ & $-0.00176$ & $-0.00481$ & $0.00093$ & $0.2359$ \\
\specialrule{0.2pt}{1.0pt}{1.0pt}
pLDDT after RSA & $\mathrm{Full} - (M{+}C{+}R)$ & $-0.00083$ & $-0.00335$ & $0.00156$ & $0.5131$ \\
\specialrule{0.2pt}{1.0pt}{1.0pt}
RSA after pLDDT & $\mathrm{Full} - (M{+}C{+}P)$ & $+0.01390$ & $0.00684$ & $0.02138$ & $\boldsymbol{0.00015}$ \\
\specialrule{0.2pt}{1.0pt}{1.0pt}
Full vs. MSA fusion & $\mathrm{Full} - (M)$ & $+0.01544$ & $0.00771$ & $0.02343$ & $\boldsymbol{<0.0001}$ \\
\midrule
\multicolumn{6}{@{}l}{\textbf{\ViroBenchmarkName{} (89 assays)}} \\
\specialrule{0.3pt}{1.5pt}{1.5pt}
Adaptive MSA fusion & $M - (\mathrm{Raw})$ & $+0.01919$ & $0.01196$ & $0.02649$ & $\boldsymbol{0.00015}$ \\
\specialrule{0.2pt}{1.0pt}{1.0pt}
Adaptive vs. fixed fusion & $M - (M_{0.8})$ & $+0.00042$ & $-0.00358$ & $0.00431$ & $0.8932$ \\
\specialrule{0.2pt}{1.0pt}{1.0pt}
Background correction & $M{+}C - (M)$ & $+0.00407$ & $0.00269$ & $0.00545$ & $\boldsymbol{<0.0001}$ \\
\specialrule{0.2pt}{1.0pt}{1.0pt}
Gated vs. ungated & $M{+}C - (M{+}C_{\mathrm{ungated}})$ & $+0.01009$ & $0.00570$ & $0.01436$ & $\boldsymbol{0.0019}$ \\
\specialrule{0.2pt}{1.0pt}{1.0pt}
RSA weighting & $M{+}C{+}R - (M{+}C)$ & $+0.01438$ & $0.00948$ & $0.01934$ & $\boldsymbol{0.00155}$ \\
\specialrule{0.2pt}{1.0pt}{1.0pt}
pLDDT weighting & $M{+}C{+}P - (M{+}C)$ & $+0.00523$ & $-0.00034$ & $0.01070$ & $0.1368$ \\
\specialrule{0.2pt}{1.0pt}{1.0pt}
pLDDT after RSA & $\mathrm{Full} - (M{+}C{+}R)$ & $+0.00170$ & $-0.00204$ & $0.00538$ & $0.476$ \\
\specialrule{0.2pt}{1.0pt}{1.0pt}
RSA after pLDDT & $\mathrm{Full} - (M{+}C{+}P)$ & $+0.01085$ & $0.00709$ & $0.01466$ & $\boldsymbol{0.00725}$ \\
\specialrule{0.2pt}{1.0pt}{1.0pt}
Full vs. MSA fusion & $\mathrm{Full} - (M)$ & $+0.02015$ & $0.01266$ & $0.02779$ & $\boldsymbol{0.00175}$ \\
\bottomrule
\end{tabular*}
\par
\endgroup
\end{table}

\subsection{\reviewchange{Transfer of component ablations to \textsc{SaProt}-650M-AF2}}
\label{app:saprot_transfer}

\reviewchange{To assess transfer to another PFM, we repeat the component ablation with frozen \textsc{SaProt}-650M-AF2 and masked scoring on the same 21 validation assays and all 217 ProteinGym assays (Table~\ref{tab:saprot_component_ablation}). Table~\ref{tab:saprot_component_statistics} reports the nine paired statistical comparisons.}

\begin{table}[!hb]
\begingroup
\ifreviewchanges\color{red}\fi
\caption{\reviewchange{\textsc{SaProt}-650M-AF2 ablation with masked scoring and the settings in Table~\ref{tab:ablation}. PG validation reports mean assay Spearman; PG full reports official Average Spearman. $\Delta_{\mathrm{Raw}}$ is the change from the original PFM score. Both structural-weighting branches follow adaptive MSA fusion and gated background correction.}}
\label{tab:saprot_component_ablation}
\centering
\small
\setlength{\tabcolsep}{4pt}
\begin{tabular}{@{}llcccc@{}}
\toprule
\multirow{2}{*}{Component} & \multirow{2}{*}{Setting} & \multicolumn{2}{c}{PG validation (21)} & \multicolumn{2}{c}{PG full (217)} \\
\cmidrule(lr){3-4}\cmidrule(l){5-6}
& & Score & $\Delta_{\mathrm{Raw}}$ & Score & $\Delta_{\mathrm{Raw}}$ \\
\midrule
Raw & PFM scores & 0.4953 & --- & 0.4575 & --- \\
\midrule
MSA fusion & Fixed weight ($\alpha=0.8$) & 0.5150 & +0.020 & 0.4710 & +0.014 \\
MSA fusion & Adaptive weight & 0.5185 & +0.023 & 0.4742 & +0.017 \\
\midrule
Model correction & Without gate & 0.5203 & +0.025 & 0.4741 & +0.017 \\
Model correction & With gate & 0.5240 & +0.029 & 0.4778 & +0.020 \\
\midrule
Structural weighting & RSA only & 0.5347 & +0.039 & 0.4862 & +0.029 \\
Structural weighting & pLDDT only & 0.5289 & +0.034 & 0.4798 & +0.022 \\
\midrule
\textsc{\HarnessName{}} & Full method & \textbf{0.5369} & \textbf{+0.042} & \textbf{0.4868} & \textbf{+0.029} \\
\bottomrule
\end{tabular}
\par
\endgroup
\end{table}

\begin{table}[!htbp]
\begingroup
\ifreviewchanges\color{red}\fi
\caption{\reviewchange{Paired component comparisons for \textsc{SaProt}-650M-AF2 with masked scoring. Columns report $\Delta$, its 95\% confidence interval (CI), and the two-sided $p$-value (Appendix~\ref{app:component_statistics}); bold $p$-values denote $p<0.05$. Notation follows Table~\ref{tab:prosst_component_statistics}.}}
\label{tab:saprot_component_statistics}
\centering
\footnotesize
\setlength{\tabcolsep}{4pt}
\renewcommand{\arraystretch}{1.06}
\begin{tabular*}{\textwidth}{@{\extracolsep{\fill}}llrrrr@{}}
\toprule
\multirow{2}{*}{Comparison} & \multirow{2}{*}{Score difference} & \multirow{2}{*}{$\Delta$} & \multicolumn{2}{c}{95\% CI} & \multirow{2}{*}{$p$} \\
\cmidrule(lr){4-5}
 & & & Lower & Upper & \\
\midrule
\multicolumn{6}{@{}l}{\textbf{ProteinGym validation (21 assays)}} \\
\specialrule{0.3pt}{1.5pt}{1.5pt}
Adaptive MSA fusion & $M - (\mathrm{Raw})$ & $+0.02324$ & $0.00315$ & $0.04702$ & $0.0569$ \\
\specialrule{0.2pt}{1.0pt}{1.0pt}
Adaptive vs. fixed fusion & $M - (M_{0.8})$ & $+0.00354$ & $-0.00344$ & $0.01281$ & $0.477$ \\
\specialrule{0.2pt}{1.0pt}{1.0pt}
Background correction & $M{+}C - (M)$ & $+0.00551$ & $0.00113$ & $0.00948$ & $\boldsymbol{0.02055}$ \\
\specialrule{0.2pt}{1.0pt}{1.0pt}
Gated vs. ungated & $M{+}C - (M{+}C_{\mathrm{ungated}})$ & $+0.00370$ & $-0.00425$ & $0.01406$ & $0.541$ \\
\specialrule{0.2pt}{1.0pt}{1.0pt}
RSA weighting & $M{+}C{+}R - (M{+}C)$ & $+0.01068$ & $0.00597$ & $0.01549$ & $\boldsymbol{0.00025}$ \\
\specialrule{0.2pt}{1.0pt}{1.0pt}
pLDDT weighting & $M{+}C{+}P - (M{+}C)$ & $+0.00493$ & $-0.00040$ & $0.01101$ & $0.1109$ \\
\specialrule{0.2pt}{1.0pt}{1.0pt}
pLDDT after RSA & $\mathrm{Full} - (M{+}C{+}R)$ & $+0.00217$ & $-0.00196$ & $0.00688$ & $0.3731$ \\
\specialrule{0.2pt}{1.0pt}{1.0pt}
RSA after pLDDT & $\mathrm{Full} - (M{+}C{+}P)$ & $+0.00791$ & $0.00285$ & $0.01271$ & $\boldsymbol{0.00605}$ \\
\specialrule{0.2pt}{1.0pt}{1.0pt}
Full vs. MSA fusion & $\mathrm{Full} - (M)$ & $+0.01836$ & $0.00938$ & $0.02748$ & $\boldsymbol{0.0008}$ \\
\midrule
\multicolumn{6}{@{}l}{\textbf{ProteinGym full (217 assays)}} \\
\specialrule{0.3pt}{1.5pt}{1.5pt}
Adaptive MSA fusion & $M - (\mathrm{Raw})$ & $+0.01673$ & $0.00910$ & $0.02483$ & $\boldsymbol{<0.0001}$ \\
\specialrule{0.2pt}{1.0pt}{1.0pt}
Adaptive vs. fixed fusion & $M - (M_{0.8})$ & $+0.00318$ & $-0.00087$ & $0.00746$ & $0.1622$ \\
\specialrule{0.2pt}{1.0pt}{1.0pt}
Background correction & $M{+}C - (M)$ & $+0.00358$ & $-0.00108$ & $0.00786$ & $0.1109$ \\
\specialrule{0.2pt}{1.0pt}{1.0pt}
Gated vs. ungated & $M{+}C - (M{+}C_{\mathrm{ungated}})$ & $+0.00367$ & $-0.00142$ & $0.00891$ & $0.2112$ \\
\specialrule{0.2pt}{1.0pt}{1.0pt}
RSA weighting & $M{+}C{+}R - (M{+}C)$ & $+0.00843$ & $0.00482$ & $0.01205$ & $\boldsymbol{<0.0001}$ \\
\specialrule{0.2pt}{1.0pt}{1.0pt}
pLDDT weighting & $M{+}C{+}P - (M{+}C)$ & $+0.00196$ & $0.00049$ & $0.00341$ & $\boldsymbol{0.0106}$ \\
\specialrule{0.2pt}{1.0pt}{1.0pt}
pLDDT after RSA & $\mathrm{Full} - (M{+}C{+}R)$ & $+0.00056$ & $-0.00068$ & $0.00170$ & $0.3744$ \\
\specialrule{0.2pt}{1.0pt}{1.0pt}
RSA after pLDDT & $\mathrm{Full} - (M{+}C{+}P)$ & $+0.00703$ & $0.00348$ & $0.01062$ & $\boldsymbol{<0.0001}$ \\
\specialrule{0.2pt}{1.0pt}{1.0pt}
Full vs. MSA fusion & $\mathrm{Full} - (M)$ & $+0.01257$ & $0.00540$ & $0.01987$ & $\boldsymbol{<0.0001}$ \\
\bottomrule
\end{tabular*}
\par
\endgroup
\end{table}

\subsection{PFM and MSA score construction}
\label{app:score_construction}

\reviewchange{The frequencies in Equation~\ref{eq:msa_frequency} count every MSA sequence once, with no sequence weighting or pseudocounts. Gaps are counted as padding symbols, and unobserved symbols have frequency zero. Let $\mathcal T$ contain all tokenizer symbols. Then}
\begin{equation}
\reviewchange{E_{i,a}=f_{i,a}-\log\sum_{\nu\in\mathcal T}e^{f_{i,\nu}}.}
\label{eq:family_field}
\end{equation}
\reviewchange{We retain the 20 amino-acid entries for scoring. Log-softmax subtracts one constant from each row, so $E_{i,a}-E_{i,x_i}=f_{i,a}-f_{i,x_i}$. The MSA contribution therefore depends on frequency differences. Positions outside the alignment use $F_{i,\cdot}=R_{i,\cdot}$ before background correction and structural weighting.}

\reviewchange{Equation~\ref{eq:readouts} gives the log-probability form of the PFM scores. \textsc{ProtSSN} instead supplies raw logits, and long \textsc{ProSST} sequences use log probabilities averaged across overlapping windows. These values are stored as $R$ and enter fusion and background correction directly. Softmax over the 20 amino-acid entries is used when calculating the adaptive weight.}

\begin{samepage}
\reviewchange{ProSST encodes the local 3D neighborhood of each residue with a geometric vector perceptron and assigns it to one of $k$ clusters learned from CATH structures. These indices form structure tokens. A Transformer combines them with amino-acid tokens and relative positions through disentangled attention \citep{li2024prosst}. For checkpoint $k$, let $t^{(k)}$ be the WT structure-token sequence and $\theta_k$ its frozen parameters. The WT score matrix is}
\begin{equation}
\reviewchange{R_{k,i,a}=\log P_{\theta_k,i}(a;x,t^{(k)}).}
\label{eq:prosst_inference}
\end{equation}
\end{samepage}

\newpage
\subsection{Derivation of the MSA mixing coefficient}
\label{app:scale_derivation}

\paragraph{Probabilities and covered positions.}
\reviewchange{Section~\ref{sec:fusion} assumes an MSA covering the full WT sequence. For partial coverage, let $\mathcal A$ be the 20 standard amino acids and $\mathcal I$ the covered positions with $x_i\in\mathcal A$, with $n=|\mathcal I|$. Equations~\ref{eq:wt_weight}--\ref{eq:pref_map} then average over these $n$ positions. The probabilities used in these equations are}
\begin{equation}
p_i(a)=\frac{\exp R_{i,a}}{\sum_{a'\in\mathcal A}\exp R_{i,a'}},
\qquad i\in\mathcal I.
\label{eq:cue_marginal}
\end{equation}
\reviewchange{For $R=\log P$ in Equation~\ref{eq:readouts}, this divides each amino-acid probability by the sum over the 20 amino acids. It also converts raw logits or averaged log probabilities into a distribution on the same 20 residues. Base-20 logarithms give $\rho_{\mathrm{WT}}=1$ for uniform probabilities and a maximum entropy of one. The clipping function is $\operatorname{clip}(z,0,1)=\min\{1,\max\{0,z\}\}$.}

\paragraph{Definitions and scale constraint.}
\reviewchange{The PFM and MSA score changes relative to the WT residue are}
\begin{equation}
\reviewchange{\Delta^R_{i,a}=R_{i,a}-R_{i,x_i},\qquad
\Delta^E_{i,a}=E_{i,a}-E_{i,x_i}.}
\label{eq:contrasts}
\end{equation}
\reviewchange{The weighted average in Equation~\ref{eq:fusion} therefore gives $\Delta^F_{i,a}=(1-\alpha)\Delta^R_{i,a}+\alpha\Delta^E_{i,a}$.}
For $n=|\mathcal I|>0$, let $\mathcal D=\{(i,a):i\in\mathcal I,\ a\in\mathcal A\setminus\{x_i\}\}$, so $|\mathcal D|=19n$. For $X\in\{R,E\}$, the scales in Equation~\ref{eq:native_alpha} are
\begin{equation}
\overline{\Delta}^{X}=\frac{1}{19n}\sum_{(i,a)\in\mathcal D}\Delta^X_{i,a},
\qquad
s_X^2=\frac{1}{19n-1}\sum_{(i,a)\in\mathcal D}
\bigl(\Delta^X_{i,a}-\overline{\Delta}^{X}\bigr)^2.
\label{eq:contrast_scales}
\end{equation}
The set $\mathcal D$ includes every non-WT amino acid at each covered position, so the standard deviations do not depend on the variants measured in an assay. Subtracting a constant from a row preserves both these differences and its Pearson correlation with another row. Thus using only the 20 amino acids in the MSA log-softmax gives the same adaptive coefficient and substitution scores as using the full tokenizer vocabulary.

\reviewchange{The weights $1-\rho_{\mathrm{MSA}}$ and $\rho_{\mathrm{MSA}}$ apply after dividing each set of differences by its standard deviation. Equation~\ref{eq:fusion} uses the original scores, whose weighted standard deviations are $(1-\alpha)s_R$ and $\alpha s_E$. Matching their relative weights requires}
\begin{equation}
\reviewchange{\rho_{\mathrm{MSA}}=\frac{\alpha s_E}{(1-\alpha)s_R+\alpha s_E}.}
\label{eq:scale_constraint}
\end{equation}

\paragraph{Derivation and scope.}
\reviewchange{Equal standard deviations give $\alpha=\rho_{\mathrm{MSA}}$. When MSA differences are smaller, their coefficient must be larger to retain the same relative weight.}
Since $\bar H$, $\rho_{\mathrm{WT}}$, and $\rho_{\mathrm{corr}}$ lie in $[0,1]$, Equation~\ref{eq:pref_map} gives $\rho_{\mathrm{MSA}}\in[0,1]$. For $s_R,s_E>0$, multiplying the scale constraint by its positive denominator yields
\begin{equation}
\begin{aligned}
\rho_{\mathrm{MSA}}(1-\alpha)s_R&=(1-\rho_{\mathrm{MSA}})\alpha s_E,\\
\rho_{\mathrm{MSA}}s_R&=\alpha\bigl[\rho_{\mathrm{MSA}}s_R+(1-\rho_{\mathrm{MSA}})s_E\bigr],\\
\alpha&=\frac{\rho_{\mathrm{MSA}}s_R}{\rho_{\mathrm{MSA}}s_R+(1-\rho_{\mathrm{MSA}})s_E}.
\end{aligned}
\label{eq:alpha_derivation}
\end{equation}
The bracket is positive, giving the unique solution in Equation~\ref{eq:native_alpha}. Its denominator is at least its nonnegative numerator, hence $0\le\alpha\le1$; $\alpha=0$ when $\rho_{\mathrm{MSA}}=0$ and $\alpha=1$ when $\rho_{\mathrm{MSA}}=1$. The constraint concerns the sum of weighted standard deviations, whereas the variance of the combined score also includes covariance. Entropy weighting and the gate are empirical rules evaluated in the component ablations.

\paragraph{Numerical rules.}
WT probabilities are bounded below by $10^{-12}$ before taking logarithms. Entropy is clipped to $[0,1]$ to remove roundoff. For the PFM--MSA correlation, the product of the centered-vector norms is bounded below by $10^{-12}$, and constant vectors give zero correlation. Adaptive fusion requires finite $s_R,s_E$, $s_E>0$, and a positive denominator in the final expression of Equation~\ref{eq:alpha_derivation}. When $s_R=0<s_E$ and $\rho_{\mathrm{MSA}}<1$, the formula gives $\alpha=0$. With no usable MSA, the wrapper uses $\alpha=0$. If an available MSA fails the adaptive calculation, it uses $\alpha=0.8$ and background-correction weight $0.2$.

\begin{figure}[!ht]
\centering
\includegraphics[width=0.62\textwidth]{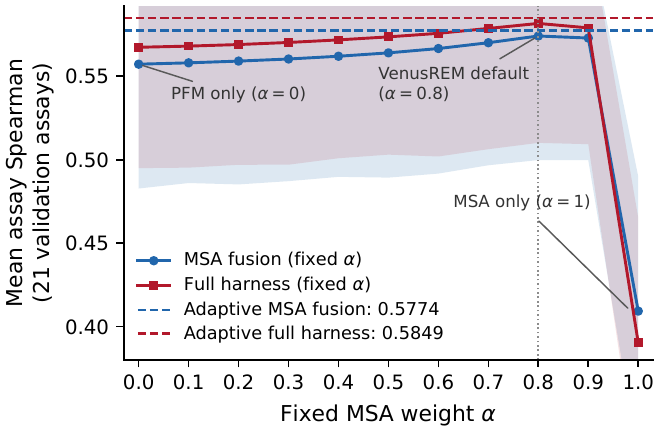}
\caption{Mean assay Spearman on the 21-assay validation split (\textsc{ProSST}-2048) under a fixed mixing coefficient $\alpha$, for MSA fusion alone and the full harness; bands are 95\% assay-bootstrap intervals. Dashed lines: the adaptive $\alpha$ of Equation~\ref{eq:native_alpha}. Both endpoints perform worse, and the adaptive coefficient matches or exceeds the best fixed coefficient.}
\label{fig:alpha_sweep}
\end{figure}

\paragraph{Fixed MSA mixing coefficients.}
\label{app:alpha_sweep}
Figure~\ref{fig:alpha_sweep} tests fixed values $\alpha\in\{0.0,0.1,\dots,1.0\}$ using the same \textsc{ProSST}-2048 score matrices and 21 validation assays as Table~\ref{tab:ablation}. We evaluate both MSA fusion alone and the full harness. At $\alpha{=}0$, fusion returns the original PFM scores; at $\alpha{=}1$, it uses only the MSA and reaches $0.4093$. Both endpoints perform worse than combining the two sources. Performance is stable over $\alpha\in[0.6,0.9]$ ($0.567$--$0.574$ for MSA fusion alone). The adaptive coefficient reaches $0.5774$ with fusion alone and $0.5849$ with the full harness, matching or exceeding every fixed coefficient without tuning on each protein. Averaging fold means over all ten folds preserves this pattern ($0.531$--$0.537$ for fusion alone over $\alpha\in[0.6,0.9]$; $0.398$ at $\alpha{=}1$). Shaded intervals are 95\% assay-level bootstrap intervals over the 21 validation assays.

Table~\ref{tab:gate_ablation} isolates the ProteinGym gate comparison: the fixed ungated correction improves 43 of 71 configurations, whereas gated correction improves all 71 and yields the larger mean gain over retrieval.

\begin{table}[H]
\caption{Effect of gating background correction across 71 ProteinGym configurations (official Average Spearman).}
\label{tab:gate_ablation}
\begin{center}
\small
\setlength{\tabcolsep}{4pt}
\begin{tabular}{lcccc}
\toprule
Comparison & Mean $\Delta$ & Median $\Delta$ & Wins / $N$ & Ties \\
\midrule
Gated $-$ no correction & +0.012 & +0.010 & 71 / 71 & 0 \\
Ungated $\kappa=1$ $-$ no correction & +0.003 & +0.003 & 43 / 71 & 0 \\
Gated $-$ ungated $\kappa=1$ & +0.010 & +0.009 & 71 / 71 & 0 \\
\bottomrule
\end{tabular}
\end{center}
\end{table}

\FloatBarrier
\subsection{Background correction and leave-one-site-out checks}
\label{app:calibration_identities}
\paragraph{Correction of the PFM scores.}
For Equation~\ref{eq:background}, $\bar b=20^{-1}\sum_{a\in\mathcal A}b_a$ and $s_b^2=19^{-1}\sum_{a\in\mathcal A}(b_a-\bar b)^2$. The background is calculated as $\operatorname{logsumexp}_i R_{i,a}-\log L$ to avoid overflow. We set $\widehat b=0$ when $s_b\le10^{-8}$. For the gate, a centered background norm at most $10^{-12}$ gives zero correlation, and each centered site-vector norm is bounded below by $10^{-12}$.

Equation~\ref{eq:gate} uses the positive part of the mean correlation. Constant background or site-score vectors contribute zero correlation.
\reviewchange{For the corrected matrix $G$ in Equation~\ref{eq:corrected_field}, the substitution score before structural weighting is}
\begin{equation}
\begin{aligned}
\reviewchange{S^{\mathrm{cal}}_{i,a}}&\reviewchange{=G_{i,a}-G_{i,x_i}}\\
&\reviewchange{=F_{i,a}-F_{i,x_i}-(1-\alpha)\kappa(R)(\widehat b_a-\widehat b_{x_i}).}
\end{aligned}
\label{eq:calibrated}
\end{equation}
\reviewchange{Expanding $F$ using Equation~\ref{eq:fusion} shows that the MSA term $\alpha\Delta^E_{i,a}$ stays unchanged, and the model background correction is multiplied by $1-\alpha$. Since Pearson correlations lie in $[-1,1]$, $0\le\kappa\le1$; a non-positive mean correlation gives $\kappa=0$. The subtraction therefore acts on the PFM contribution, with its strength controlled by the repeated preference for the same amino acids across positions.}

\begin{figure}[!ht]
\centering
\includegraphics[width=0.95\textwidth]{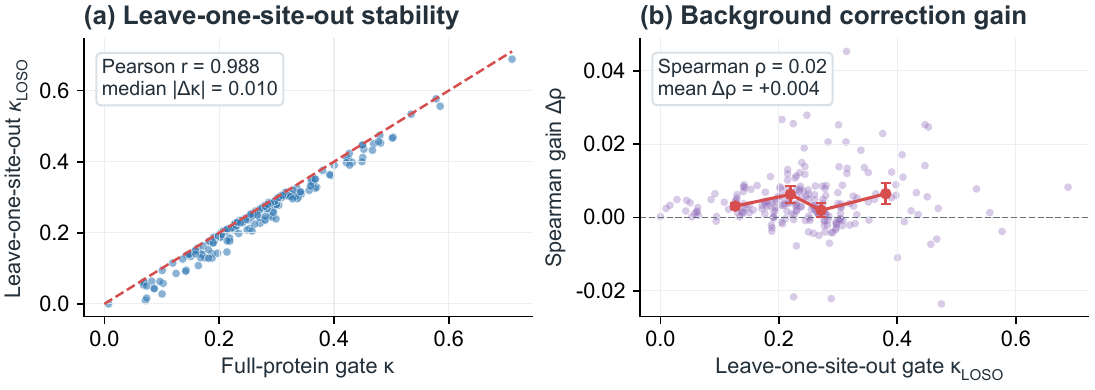}
\caption{Gate stability across 217 \textsc{ProSST-2048} assays. \textbf{(a)} $\kappa_{\mathrm{LOSO}}$ is close to the full-protein gate. \textbf{(b)} Gain from background correction after MSA fusion versus the leave-one-site-out gate $\kappa_{\mathrm{LOSO}}$; red points show quartile means and 95\% bootstrap intervals.}
\label{fig:coherence_mechanism}
\end{figure}

\paragraph{Excluding the scored site.}
For $L>1$, removing site $i$ from background construction gives the background vector $\mathbf b^{(-i)}$ and gate estimate
\begin{equation}
\begin{gathered}
b_a^{(-i)}=\log\!\left(\frac{1}{L-1}\sum_{\substack{1\le j\le L\\j\ne i}}e^{R_{j,a}}\right),\\
c_{\mathrm{LOSO}}=\frac{1}{L}\sum_{i=1}^{L}\operatorname{corr}\!\left(R_{i,\cdot},\mathbf b^{(-i)}\right),
\qquad
\kappa_{\mathrm{LOSO}}=\max(0,c_{\mathrm{LOSO}}).
\end{gathered}
\label{eq:loso_gate}
\end{equation}
LOSO denotes leave-one-site-out: the reference for position $i$ is constructed from the other $L-1$ positions.
Across 217 \textsc{ProSST}-2048 assays and 187 unique WT sequences, leave-one-site-out estimation preserves the full-protein gate (Spearman $\rho=0.985$, Pearson $r=0.988$, median $|\Delta\kappa|=0.010$). Figure~\ref{fig:coherence_mechanism}b compares the per-assay Spearman gain from gated background correction after MSA fusion and reports bootstrap intervals for quartiles of $\kappa_{\mathrm{LOSO}}$. This gate has no detectable association with the gain ($\rho=0.016$, $p=0.811$). The gate estimate is stable, but larger values are not associated with larger gains.

\FloatBarrier
\subsection{Weighting by exposure and confidence}
\label{app:structure_extension}
\reviewchange{RSA is solvent-accessible surface area divided by the residue-specific reference maximum, then clipped to $[0,1]$ to give $r_i$. We similarly set $\ell_i=\operatorname{clip}(\mathrm{pLDDT}_i/100,0,1)$ and $q_i=1-\ell_i$.} The reference means in Equation~\ref{eq:lambda} use strictly positive entries. With $\mathcal J_r=\{i:r_i>0\}$ and $\mathcal J_\ell=\{i:\ell_i>0\}$, they are
\begin{equation}
\bar r=\frac{1}{|\mathcal J_r|}\sum_{i\in\mathcal J_r}r_i,
\qquad
\bar\ell=\frac{1}{|\mathcal J_\ell|}\sum_{i\in\mathcal J_\ell}\ell_i,
\qquad \bar q=1-\bar\ell.
\label{eq:structure_means}
\end{equation}
An empty $\mathcal J_r$ or $\mathcal J_\ell$ gives a zero mean, with $\bar q=1$ in the latter case. A zero RSA entry receives $w_i^{\mathrm{RSA}}=1$. A zero pLDDT entry receives $w_i^{\mathrm{pLDDT}}=1-\bar\ell$. If an entire RSA or pLDDT array is unavailable, its corresponding weight is set to one and the other factor is still applied.

Because $r_i,\bar r,q_i,\bar q\in[0,1]$, both maximum terms in Equation~\ref{eq:lambda} lie in $[0,1]$. Thus each weight and their product lie in $[0,1]$, giving $|S_{i,a}|=\lambda_i|S^{\mathrm{cal}}_{i,a}|\le|S^{\mathrm{cal}}_{i,a}|$. Positive weights preserve the sign of an individual substitution score; a zero weight sets it to zero. Because weights vary across sites, this does not imply unchanged cross-site ranks or preservation of the sign of a multi-substitution sum. Table~\ref{tab:ablation} tests the empirical contributions of the two factors.

\newpage
\subsection{Variant aggregation and ensemble standardization}
\label{app:ensemble_definition}
Equation~\ref{eq:aggregate} sums entries from a fixed score matrix, with at most one substitution per site. The WT has no substitutions and a score of zero. For substitutions at distinct sites, additivity gives $S_k(v_{12})-S_k(v_1)-S_k(v_2)+S_k(\mathrm{WT})=0$, where $v_{12}$ carries the substitutions in both $v_1$ and $v_2$. Appendix~\ref{app:epistasis} separately tests scores recomputed on mutant backgrounds.

For the six-checkpoint ensemble in Section~\ref{sec:venusrem2}, let $\mathcal V_A$ be the set of scored records in assay $A$ and $N_A=|\mathcal V_A|>0$. All six checkpoints provide a finite prediction for each record. For \ViroBenchmarkName{}, this set includes XnX controls retained in the score files. Standardization uses all these records, and evaluation then removes the controls as described in Appendix~\ref{app:metric_defs}. The mean, population standard deviation, and standardized score are
\begin{subequations}\label{eq:ensemble_statistics}
\begin{gather}
\mu_{k,A}=\frac{1}{N_A}\sum_{v\in\mathcal V_A}S_k(v),
\qquad
\sigma_{k,A}^2=\frac{1}{N_A}\sum_{v\in\mathcal V_A}\bigl(S_k(v)-\mu_{k,A}\bigr)^2,
\label{eq:ensemble_moments}\\
z_{k,A}(v)=
\begin{cases}
\bigl(S_k(v)-\mu_{k,A}\bigr)/\sigma_{k,A}, & \sigma_{k,A}>0,\\
0, & \sigma_{k,A}=0.
\end{cases}
\label{eq:ensemble_zscore}
\end{gather}
\end{subequations}
Equation~\ref{eq:prosst_ensemble} averages the six standardized predictions, counting a constant member as zero. The normalization statistics use predicted scores only. Each checkpoint's score matrix depends on the WT protein, MSA, structure, and frozen parameters. The ensemble also depends on $\mathcal V_A$, because changing the records can change the standard deviations and hence the relative checkpoint weights.

\FloatBarrier
\section{Parkin abundance and structural context}
\label{app:parkin_details}

Parkin is a 465-residue RING-between-RING E3 ubiquitin ligase. Its VAMP-seq abundance measurements reflect mutation effects on folding stability and protein homeostasis in cells. The original study linked reduced abundance to thermodynamic destabilization and proteasomal degradation, while also identifying local degradation signals that need not destabilize the native fold \citep{clausen2024parkin}. Its ubiquitin-like Ubl, RING0, RING1, IBR, REP/tether, and RING2 architecture couples folded zinc-binding domains to autoinhibitory linkers \citep{kumar2015parkin}. Of $8{,}756$ measured substitutions across $462$ profiled sites, $89.2\%$ meet the model--MSA separation criterion in Appendix~\ref{app:profile_separation}. The original \textsc{ESM-2} 650M model scored on the unmasked WT sequence reaches $0.503$ Spearman, MSA fusion $0.585$, and the full harness $0.645$; abundance-class AUC rises from $0.769$ to $0.815$ and $0.857$ (Figure~\ref{fig:viral_parkin}).

The rank shift follows domain organization (Figure~\ref{fig:viral_parkin}d). The harness improves every folded domain, including RING0 from $0.425$ to $0.627$, IBR from $0.727$ to $0.786$, and RING2 from $0.427$ to $0.529$; REP/tether rises from $-0.083$ to $0.221$. The flexible Ubl--RING0 linker instead falls from $0.363$ to $0.176$. N232A lies at the $0.4$th measured percentile but is ranked at the $88.8$th percentile by the original model; MSA fusion and the full harness move it to $58.0$ and $37.3$. R396H moves in the opposite direction, from $39.7$ to $71.7$ and $78.0$, toward its measured $77.6$th percentile (Figure~\ref{fig:viral_parkin}e).

\paragraph{Region analysis across proteins.}
For each ProteinGym assay, we group substitutions by AlphaFold2 confidence: pLDDT below 50 or at least 50. We recompute Spearman for each group using \textsc{ESM-2} 650M unmasked WT scoring and retain groups with at least 20 variants at five or more sites. For pLDDT $\geq50$, $83\%$ of 214 qualifying assays improve (mean $\Delta=+0.056$), and $5.6\%$ decrease by more than $0.02$. For pLDDT $<50$, $57\%$ of 65 qualifying assays decrease, with mean $\Delta=-0.020$. Gains are therefore concentrated in regions with higher structural confidence.

\FloatBarrier
\section{Scoring methods, runtime, and epistasis}
\label{app:interface_runtime}

\subsection{Reusing cached score matrices}
\label{app:amortized}

\begin{figure}[!htbp]
\centering
\includegraphics[width=0.40\textwidth]{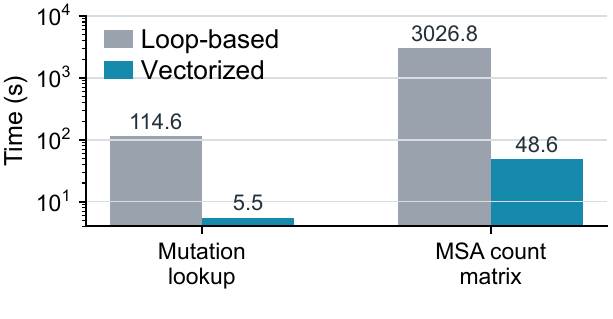}
\caption{Runtime after model inference on ProteinGym ($2.47$ million variants). Cached mutation-score lookup and vectorized MSA counting are compared with loop-based implementations.}
\label{fig:runtime}
\end{figure}

\reviewchange{Figure~\ref{fig:runtime} measures processing time after one \textsc{ProSST} forward pass has produced a score matrix for each protein. Across 2.47 million ProteinGym variants, indexed lookup reduces mutation scoring from $114.56$ to $5.53$ seconds ($20.7\times$), while vectorized counting reduces MSA count-matrix construction from $3{,}026.75$ to $48.61$ seconds ($62.3\times$).} Given PFM and MSA score matrices $R$ and $E$, constructing the final score matrix costs $O(20L)$ per protein. Each $M$-substitution variant then requires $O(M)$ lookup and summation. Keeping the frozen model loaded in memory, tokenizing inputs in batches, and reusing score matrices avoid repeated work for each variant.

\FloatBarrier
\subsection{Scoring methods across architectures}
\label{app:interface_audit}
Table~\ref{tab:score_interfaces} summarizes the inputs and computational cost of the three methods defined in Section~\ref{sec:interface}. Each produces a score matrix $R$, from which substitution scores follow Equation~\ref{eq:contrasts}. The comparisons below test how the choice of input context affects prediction.

\begin{table}[!ht]
\caption{Inputs and cost of three scoring methods. Masked inputs can be batched. In teacher forcing, the target residue is excluded from the input used to predict that position.}
\label{tab:score_interfaces}
\centering
\footnotesize
\setlength{\tabcolsep}{2pt}
\begin{tabular}{@{}>{\raggedright\arraybackslash}p{0.165\textwidth}>{\raggedright\arraybackslash}p{0.19\textwidth}>{\raggedright\arraybackslash}p{0.17\textwidth}>{\raggedright\arraybackslash}p{0.15\textwidth}>{\raggedright\arraybackslash}p{0.225\textwidth}@{}}
\toprule
Scoring & Context at site $i$ & WT residue visible & Score matrix cost & Multi-mutant scoring \\
\midrule
Masked scoring & $x_{\setminus i}$ & No & $L$ masked inputs & Sum of site scores \\
WT scoring & Full WT sequence $x$ & Yes & 1 forward pass & Sum of site scores \\
Teacher forcing & Prefix $x_{<i}$ & \mbox{No (prefix only)} & 1 forward pass & Sum of site scores \\
\bottomrule
\end{tabular}
\end{table}

\paragraph{Bidirectional encoders: masked versus unmasked WT scoring.}
Across eight paired encoder checkpoints, the harness improves results with both masked and unmasked WT scoring on all three benchmarks (Table~\ref{tab:mask_wt}).

\begin{table}[!ht]
\caption{Mean assay Spearman across eight paired checkpoints: \textsc{ESM-1b}, \textsc{ESM-1v}, \textsc{ESM-2} 8M/35M/150M/650M/3B, and \textsc{SaProt} 650M-AF2. The harness uses adaptive MSA fusion. Raw and \textsc{\HarnessShortName{}} denote results before and after applying the harness. ProteinGym uses assay means (not official Average Spearman); both other benchmarks use macro-averages. WT control rows are excluded from \ViroBenchmarkName{} evaluation.}
\label{tab:mask_wt}
\centering
\small
\setlength{\tabcolsep}{7pt}
\begin{tabular}{lrrrrrr}
\toprule
Dataset & \multicolumn{3}{c}{Masked} & \multicolumn{3}{c}{Wild-type} \\
\cmidrule(lr){2-4}\cmidrule(lr){5-7}
& Raw & \textsc{\HarnessShortName{}} & $\Delta$ & Raw & \textsc{\HarnessShortName{}} & $\Delta$ \\
\midrule
ProteinGym & 0.389 & 0.468 & +0.079 & 0.395 & 0.458 & +0.062 \\
VenusMutHub & 0.157 & 0.218 & +0.061 & 0.142 & 0.187 & +0.045 \\
\ViroBenchmarkName{} & 0.082 & 0.239 & +0.156 & 0.076 & 0.263 & +0.188 \\
\bottomrule
\end{tabular}
\end{table}

\paragraph{Autoregressive and inverse-folding models: full-sequence scoring versus cached WT scores.}
Full-sequence scoring evaluates each mutant using teacher forcing and sums log probabilities across the sequence, so a substitution can change predictions at later positions. Scoring from a cached WT matrix runs teacher forcing once on the WT sequence, then subtracts the WT score from the mutant amino-acid score only at substituted positions. Table~\ref{tab:sequence_native_interfaces} compares these approaches using the same frozen checkpoints and ProteinGym assays.

\begin{table}[!ht]
\caption{ProteinGym scoring comparison on 217 assays. Full-sequence values use the official aggregate. WT matrix denotes scores obtained by teacher forcing once on the WT sequence; $\Delta$ is their performance minus full-sequence scoring, and Wins counts assays with positive changes.}
\label{tab:sequence_native_interfaces}
\centering
\small
\setlength{\tabcolsep}{6pt}
\begin{tabular}{llrrrr}
\toprule
Family & Model & Full sequence & WT matrix & $\Delta$ & Wins \\
\midrule
Inverse folding & \textsc{ESM-IF1} & 0.422 & 0.421 & $-0.001$ & 95/217 \\
Autoregressive & \textsc{ProGen2}-S & 0.336 & 0.288 & $-0.048$ & 47/217 \\
Autoregressive & \textsc{ProGen2}-M & 0.380 & 0.337 & $-0.043$ & 54/217 \\
Autoregressive & \textsc{ProGen2}-B & 0.378 & 0.330 & $-0.048$ & 58/217 \\
Autoregressive & \textsc{ProGen2}-L & 0.380 & 0.327 & $-0.053$ & 53/217 \\
Autoregressive & \textsc{ProGen2}-XL & 0.391 & 0.351 & $-0.040$ & 67/217 \\
\bottomrule
\end{tabular}
\end{table}

The two approaches produce almost the same \textsc{ESM-IF1} ranking (assay-level Spearman $0.966$), consistent with structural conditioning concentrating the likelihood change near the substitution. For \textsc{ProGen2}, full-sequence likelihood retains useful downstream context and exceeds scoring from the cached WT matrix by $0.040$--$0.053$ across the five model sizes. Cached scores can be used by \textsc{\HarnessName{}}, but their effect on the underlying model's performance depends on its architecture and scoring method.

\FloatBarrier
\subsection{\textsc{ProteinMPNN} scoring and efficiency}
\label{app:pmpnn_audit}

\reviewchange{In the identifier \texttt{v\_48\_020}, 48 specifies the graph neighborhood size and 020 denotes the standard deviation ($0.20$~\AA{}) of Gaussian noise added to backbone coordinates during training \citep{dauparas2022proteinmpnn}.}

\begin{figure}[!ht]
\begin{center}
\includegraphics[width=0.95\textwidth]{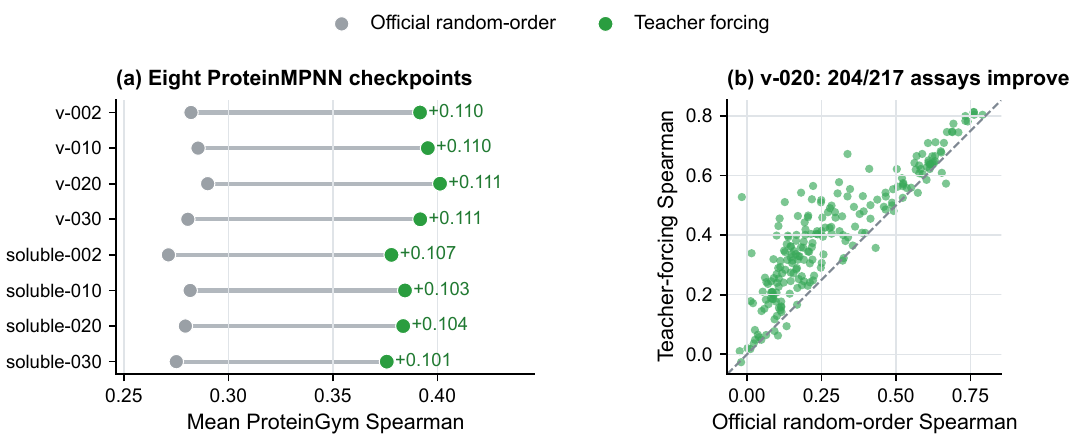}
\end{center}
\caption{\textsc{ProteinMPNN} teacher forcing versus official random-order scoring on AlphaFold2 structures. \textbf{(a)} Mean assay Spearman across eight checkpoints. \textbf{(b)} Per-assay Spearman at v\_48\_020.}
\label{fig:tf}
\end{figure}

\paragraph{Official score versus teacher forcing.}
Official \textsc{ProteinMPNN} scoring rebuilds each mutant sequence and samples a decoding order, making the score order-dependent. We instead use teacher forcing once on the WT sequence and subtract the WT score from the mutant amino-acid score at each substituted position (Table~\ref{tab:pmpnn_interfaces}). On identical AlphaFold2 structures, these cached scores raise mean assay Spearman from $0.271$--$0.290$ to $0.376$--$0.401$ across eight checkpoints while reducing runtime by $219$--$601\times$; at v\_48\_020, $204$ of $217$ assays improve (Figure~\ref{fig:tf}b). This gain contrasts with the nearly unchanged \textsc{ESM-IF1} results and the full-sequence advantage for \textsc{ProGen2}, showing that the effect depends on the model. The epistasis analysis below additionally recomputes scores on mutant backgrounds to test whether they capture interactions between substitutions.

\begin{table}[!ht]
\caption{\textsc{ProteinMPNN} scoring methods. Official scoring samples a decoding order for each mutant. WT teacher forcing (TF) reuses scores from the WT sequence; conditional TF recomputes scores on a mutant background.}
\label{tab:pmpnn_interfaces}
\centering
\small
\renewcommand{\arraystretch}{1.08}
\setlength{\tabcolsep}{2pt}
\begin{tabular}{@{}lllll@{}}
\toprule
Scoring & Context & Evaluations & \shortstack{Random\\order} & Multi-mutant score \\
\midrule
Official random order & Mutant + structure & Per mutant & Yes & May be non-additive \\
WT TF & WT + structure & Per protein & No & Additive \\
Conditional TF & Mutant background + structure & Per background & No & Non-additive \\
\bottomrule
\end{tabular}
\end{table}

\FloatBarrier
\subsection{Pairwise epistasis test}
\label{app:epistasis}

\paragraph{Experimental and predicted interactions.}
We use 535{,}917 complete WT/single/single/double quartets from the GB1 binding landscape of \citet{olson2014epistasis}. Let $A$ and $B$ denote substitutions at different positions. Experimental effects $y$ are measured on the relative $\ln W$ scale, with $y_{\mathrm{WT}}=0$. Pairwise epistasis is
\begin{equation}
\varepsilon_{\mathrm{exp}}=y_{AB}-y_A-y_B+y_{\mathrm{WT}}.
\label{eq:gb1_epistasis}
\end{equation}
We apply the same four-score difference to official \textsc{ProteinMPNN} predictions. Scores from a fixed WT matrix are additive (Appendix~\ref{app:ensemble_definition}), with maximum numerical deviation $4.8\times10^{-7}$. To test context-dependent effects, we therefore rescore substitution $A$ using a sequence carrying $B$, and vice versa. Let $\delta(A\mid B)$ be the replacement amino acid's log probability minus the WT residue's log probability at the position of $A$, after teacher-forced scoring on the $B$ background. Averaging the changes in the two directions gives
\begin{equation}
\varepsilon_{\mathrm{TF}}^{\mathrm{cond}}=\frac{1}{2}\left[
\delta(A\mid B)-\delta(A\mid\mathrm{WT})+
\delta(B\mid A)-\delta(B\mid\mathrm{WT})\right].
\label{eq:gb1_conditional_epistasis}
\end{equation}
The Non-floor subset excludes a quartet if either single or its double mutant has an experimental effect at the assay floor $\ln(0.01)\approx-4.60517$. The code identifies this value with \texttt{numpy.isclose} using relative tolerance $10^{-5}$ and absolute tolerance $10^{-8}$. Table~\ref{tab:pmpnn_epistasis} reports Spearman correlations for fitness and epistasis. On all quartets, epistasis correlations are $0.005$ for official scoring (95\% CI: $-0.002$ to $0.013$) and $0.012$ for conditional teacher forcing (95\% CI: $-0.005$ to $0.029$). Intervals use 1{,}000 bootstrap resamples of residue-position pairs, with seeds 42 and 43, respectively.

\begin{table}[H]
\caption{\textsc{ProteinMPNN} v\_48\_020 on Olson GB1. Entries are Spearman correlations for single-mutant fitness, double-mutant fitness, and pairwise epistasis ($\rho_\varepsilon$). Equations~\ref{eq:gb1_epistasis} and~\ref{eq:gb1_conditional_epistasis} define the interaction scores. Non-floor excludes quartets containing a measurement at the assay floor.}
\label{tab:pmpnn_epistasis}
\begin{center}
\small
\renewcommand{\arraystretch}{1.08}
\setlength{\tabcolsep}{3pt}
\begin{tabular*}{\textwidth}{@{\extracolsep{\fill}}lrrrrrrr@{}}
\toprule
Subset & Quartets & \shortstack{Single\\TF} & \shortstack{Single\\official} & \shortstack{Double\\TF} & \shortstack{Double\\official} & \shortstack{$\rho_\epsilon$\\official} & \shortstack{$\rho_\epsilon$\\conditional TF} \\
\midrule
All & 535{,}917 & 0.306 & 0.105 & 0.284 & 0.165 & 0.005 & 0.012 \\
Non-floor & 400{,}118 & 0.312 & 0.091 & 0.292 & 0.152 & 0.012 & 0.022 \\
\bottomrule
\end{tabular*}
\end{center}
\end{table}

On GB1, official scoring and conditional rescoring both recover little pairwise-interaction signal. Under the same frozen model and structural inputs, the less expensive sum of cached substitution scores predicts single- and double-mutant fitness more accurately.

\FloatBarrier
\section{Additional benchmark tables}
\label{app:additional_tables}

The following tables address three questions: category-specific ranking in Appendix~\ref{app:category_leaderboards}, component contributions in Appendix~\ref{app:staged_tables}, and alternative evaluation criteria in Appendix~\ref{app:per_config_metrics}. They provide the results for each configuration that support the summaries and method checks.

\subsection{Per-category model comparisons}
\label{app:category_leaderboards}
Tables~\ref{tab:leaderboard_pg_category}, \ref{tab:leaderboard_vmh_category}, and~\ref{tab:leaderboard_vvh_category} report each configuration before and after applying the harness for ProteinGym functions, VenusMutHub assay categories, and viral phenotypes, respectively. Aggregation follows each benchmark's primary protocol, and the Overall columns reproduce Tables~\ref{tab:proteingym}, \ref{tab:venusmuthub}, and \ref{tab:virogym_family}. ProSST and \textsc{VenusREM2} use the unmasked WT scoring method described in Appendix~\ref{app:baselines}.

{\fontsize{8.5}{10}\selectfont
\setlength{\LTcapwidth}{\linewidth}
\makeatletter
\let\SavedLTCaption\LT@makecaption
\def\LT@makecaption#1#2#3{\SavedLTCaption{#1}{\normalsize#2}{\normalsize#3}}
\makeatother
\setlength{\tabcolsep}{1.5pt}
\setlength{\LTcapwidth}{\linewidth}
\renewcommand{\arraystretch}{1.08}
\begin{longtable}{@{}cllccccccc@{}}
\caption{ProteinGym results before and after applying the harness, ranked by Overall score across 71 configurations. Scores use official Average Spearman with UniProt-unit aggregation. Each configuration appears without ($\times$) and with ($\checkmark$) the harness. Scoring uses masked inputs (mask), unmasked WT inputs (wt), or teacher forcing (tf). Categories are the five official selection types; the best value in each column is red.}\label{tab:leaderboard_pg_category}\\
\toprule
Rank & Model & Scoring & \textsc{\HarnessShortName{}} & Overall & Activity & Binding & Expression & Organismal & Stability \\
\midrule
\endfirsthead
\multicolumn{10}{c}{\tablename\ \thetable\ continued}\\
\toprule
Rank & Model & Scoring & \textsc{\HarnessShortName{}} & Overall & Activity & Binding & Expression & Organismal & Stability \\
\midrule
\endhead
\midrule
\multicolumn{10}{r}{Continued on next page}\\
\endfoot
\bottomrule
\endlastfoot
1 & \textbf{VenusREM2} & wt & \checkmark & \textcolor{red!75!black}{\textbf{0.556}} & \textcolor{red!75!black}{\textbf{0.541}} & \textcolor{red!75!black}{\textbf{0.495}} & \textcolor{red!75!black}{\textbf{0.557}} & \textcolor{red!75!black}{\textbf{0.494}} & \textcolor{red!75!black}{\textbf{0.691}} \\*
2 & ProSST (K=4096) & wt & \checkmark & 0.541 & 0.516 & 0.494 & 0.543 & 0.475 & 0.674 \\*
3 & ProSST (K=2048) & wt & \checkmark & 0.534 & 0.528 & 0.460 & 0.541 & 0.469 & 0.673 \\
4 & ProSST (K=1024) & wt & \checkmark & 0.532 & 0.513 & 0.467 & 0.533 & 0.476 & 0.670 \\
5 & ProSST (K=128) & wt & \checkmark & 0.525 & 0.508 & 0.472 & 0.517 & 0.463 & 0.663 \\
6 & VenusREM2 & wt & $\times$ & 0.524 & 0.480 & 0.479 & 0.538 & 0.444 & 0.681 \\
7 & ProSST (K=512) & wt & \checkmark & 0.524 & 0.507 & 0.464 & 0.520 & 0.464 & 0.665 \\
8 & ProSST (K=20) & wt & \checkmark & 0.510 & 0.493 & 0.452 & 0.507 & 0.446 & 0.650 \\
9 & ProSST (K=2048) & wt & $\times$ & 0.507 & 0.476 & 0.445 & 0.530 & 0.431 & 0.653 \\
10 & ProSST (K=4096) & wt & $\times$ & 0.498 & 0.444 & 0.472 & 0.507 & 0.416 & 0.652 \\
11 & S3F (wt) & wt & \checkmark & 0.489 & 0.504 & 0.391 & 0.479 & 0.440 & 0.631 \\
12 & S3F (mask) & mask & \checkmark & 0.488 & 0.496 & 0.407 & 0.472 & 0.443 & 0.619 \\
13 & SaProt (650M AF2, mask) & mask & \checkmark & 0.487 & 0.489 & 0.408 & 0.489 & 0.437 & 0.612 \\
14 & SaProt (650M\_PDB, mask) & mask & \checkmark & 0.486 & 0.490 & 0.392 & 0.492 & 0.443 & 0.615 \\
15 & ProSST (K=1024) & wt & $\times$ & 0.485 & 0.433 & 0.436 & 0.499 & 0.414 & 0.642 \\
16 & ESM3 & mask & \checkmark & 0.483 & 0.484 & 0.406 & 0.466 & 0.427 & 0.631 \\
17 & ProtSSN & wt & \checkmark & 0.476 & 0.489 & 0.393 & 0.478 & 0.416 & 0.605 \\
18 & ESM-IF1 & tf & \checkmark & 0.474 & 0.447 & 0.429 & 0.459 & 0.401 & 0.635 \\
19 & ProSST (K=512) & wt & $\times$ & 0.471 & 0.423 & 0.431 & 0.479 & 0.394 & 0.629 \\
20 & ProtSSN (k=20, h=1280) & wt & \checkmark & 0.470 & 0.482 & 0.390 & 0.469 & 0.406 & 0.603 \\
21 & ProtSSN (k=20, h=512) & wt & \checkmark & 0.470 & 0.484 & 0.379 & 0.472 & 0.413 & 0.603 \\
22 & ESM-2 650M (mask) & mask & \checkmark & 0.470 & 0.486 & 0.404 & 0.446 & 0.432 & 0.582 \\
23 & ProtSSN (k=20, h=768) & wt & \checkmark & 0.470 & 0.485 & 0.377 & 0.473 & 0.409 & 0.603 \\
24 & ProSST (K=128) & wt & $\times$ & 0.469 & 0.417 & 0.440 & 0.472 & 0.387 & 0.628 \\
25 & ProtSSN (k=30, h=512) & wt & \checkmark & 0.469 & 0.482 & 0.381 & 0.478 & 0.408 & 0.597 \\
26 & ESM-2 650M (wt) & wt & \checkmark & 0.468 & 0.489 & 0.384 & 0.462 & 0.411 & 0.592 \\
27 & ProtSSN (k=30, h=768) & wt & \checkmark & 0.467 & 0.481 & 0.386 & 0.463 & 0.407 & 0.597 \\
28 & ProtSSN (k=30, h=1280) & wt & \checkmark & 0.466 & 0.479 & 0.385 & 0.460 & 0.406 & 0.597 \\
29 & S3F (mask) & mask & $\times$ & 0.466 & 0.470 & 0.379 & 0.470 & 0.409 & 0.601 \\
30 & ProtSSN (k=10, h=1280) & wt & \checkmark & 0.465 & 0.473 & 0.383 & 0.471 & 0.405 & 0.592 \\
31 & SaProt (35M\_AF2, mask) & mask & \checkmark & 0.465 & 0.450 & 0.384 & 0.464 & 0.404 & 0.622 \\
32 & ESMC-600M & mask & \checkmark & 0.464 & 0.480 & 0.371 & 0.444 & 0.439 & 0.587 \\
33 & SaProt (650M\_PDB, mask) & mask & $\times$ & 0.463 & 0.464 & 0.370 & 0.488 & 0.405 & 0.590 \\
34 & ESM-1v (mask) & mask & \checkmark & 0.462 & 0.482 & 0.364 & 0.453 & 0.440 & 0.572 \\
35 & ESM-2 3B (mask) & mask & \checkmark & 0.462 & 0.483 & 0.377 & 0.441 & 0.437 & 0.571 \\
36 & ProtSSN (k=10, h=512) & wt & \checkmark & 0.462 & 0.479 & 0.379 & 0.457 & 0.402 & 0.592 \\
37 & ESMC-300M & mask & \checkmark & 0.460 & 0.479 & 0.370 & 0.433 & 0.433 & 0.584 \\
38 & S3F (wt) & wt & $\times$ & 0.460 & 0.469 & 0.365 & 0.453 & 0.406 & 0.605 \\
39 & MIF-ST & tf & \checkmark & 0.458 & 0.440 & 0.369 & 0.465 & 0.404 & 0.611 \\
40 & SaProt (650M AF2, mask) & mask & $\times$ & 0.457 & 0.459 & 0.384 & 0.485 & 0.371 & 0.588 \\
41 & ProtSSN (k=10, h=768) & wt & \checkmark & 0.457 & 0.470 & 0.368 & 0.459 & 0.398 & 0.589 \\
42 & ESM-1v (wt) & wt & \checkmark & 0.457 & 0.480 & 0.342 & 0.466 & 0.425 & 0.569 \\
43 & ProteinMPNN (v\_48\_020) & tf & \checkmark & 0.456 & 0.415 & 0.394 & 0.455 & 0.369 & 0.648 \\
44 & ESM-1b (mask) & mask & \checkmark & 0.455 & 0.473 & 0.366 & 0.448 & 0.429 & 0.560 \\
45 & ESM-2 150M (wt) & wt & \checkmark & 0.455 & 0.472 & 0.370 & 0.454 & 0.385 & 0.594 \\
46 & ESM-2 3B (wt) & wt & \checkmark & 0.455 & 0.479 & 0.355 & 0.455 & 0.414 & 0.570 \\
47 & SaProt (650M AF2, wt) & wt & \checkmark & 0.454 & 0.464 & 0.366 & 0.479 & 0.359 & 0.600 \\
48 & ProteinMPNN (v\_48\_030) & tf & \checkmark & 0.453 & 0.416 & 0.384 & 0.453 & 0.367 & 0.645 \\
49 & ESM-2 150M (mask) & mask & \checkmark & 0.452 & 0.466 & 0.367 & 0.432 & 0.409 & 0.583 \\
50 & CARP-640M & wt & \checkmark & 0.451 & 0.465 & 0.364 & 0.438 & 0.428 & 0.563 \\
51 & ProtSSN & wt & $\times$ & 0.451 & 0.467 & 0.371 & 0.452 & 0.398 & 0.570 \\
52 & ProteinMPNN (v\_48\_010) & tf & \checkmark & 0.451 & 0.414 & 0.387 & 0.453 & 0.366 & 0.635 \\
53 & SaProt (650M\_PDB, wt) & wt & \checkmark & 0.450 & 0.451 & 0.346 & 0.474 & 0.377 & 0.604 \\
54 & ESM-1b (wt) & wt & \checkmark & 0.447 & 0.466 & 0.347 & 0.452 & 0.409 & 0.562 \\
55 & ProtSSN (k=20, h=1280) & wt & $\times$ & 0.444 & 0.459 & 0.368 & 0.439 & 0.385 & 0.568 \\
56 & ProtSSN (k=20, h=512) & wt & $\times$ & 0.443 & 0.458 & 0.357 & 0.440 & 0.392 & 0.567 \\
57 & ProteinMPNN (v\_48\_002) & tf & \checkmark & 0.443 & 0.414 & 0.370 & 0.441 & 0.359 & 0.630 \\
58 & ESM3 & mask & $\times$ & 0.442 & 0.441 & 0.381 & 0.449 & 0.353 & 0.586 \\
59 & ProtSSN (k=20, h=768) & wt & $\times$ & 0.442 & 0.461 & 0.353 & 0.443 & 0.387 & 0.567 \\
60 & ProtSSN (k=30, h=768) & wt & $\times$ & 0.441 & 0.459 & 0.363 & 0.435 & 0.386 & 0.562 \\
61 & ProtSSN (k=30, h=512) & wt & $\times$ & 0.441 & 0.455 & 0.355 & 0.449 & 0.385 & 0.560 \\
62 & ProtSSN (k=30, h=1280) & wt & $\times$ & 0.439 & 0.456 & 0.361 & 0.433 & 0.386 & 0.561 \\
63 & ProSST (K=20) & wt & $\times$ & 0.438 & 0.392 & 0.410 & 0.447 & 0.354 & 0.589 \\
64 & CARP-76M & wt & \checkmark & 0.436 & 0.445 & 0.350 & 0.435 & 0.378 & 0.570 \\
65 & ProteinMPNN-Soluble (v\_48\_010) & tf & \checkmark & 0.436 & 0.399 & 0.383 & 0.394 & 0.367 & 0.634 \\
66 & ProteinMPNN-Soluble (v\_48\_020) & tf & \checkmark & 0.435 & 0.398 & 0.371 & 0.399 & 0.369 & 0.640 \\
67 & ProtSSN (k=10, h=1280) & wt & $\times$ & 0.435 & 0.447 & 0.356 & 0.439 & 0.382 & 0.552 \\
68 & ProteinMPNN-Soluble (v\_48\_030) & tf & \checkmark & 0.433 & 0.396 & 0.373 & 0.386 & 0.369 & 0.642 \\
69 & ProtSSN (k=10, h=512) & wt & $\times$ & 0.432 & 0.453 & 0.352 & 0.421 & 0.377 & 0.558 \\
70 & ProteinMPNN-Soluble (v\_48\_002) & tf & \checkmark & 0.430 & 0.397 & 0.379 & 0.388 & 0.361 & 0.626 \\
71 & MIF-ST & tf & $\times$ & 0.430 & 0.410 & 0.341 & 0.441 & 0.381 & 0.576 \\
72 & ESM-2 35M (mask) & mask & \checkmark & 0.430 & 0.437 & 0.344 & 0.410 & 0.379 & 0.578 \\
73 & SaProt (650M\_PDB, wt) & wt & $\times$ & 0.428 & 0.429 & 0.328 & 0.459 & 0.355 & 0.568 \\
74 & ESM-2 35M (wt) & wt & \checkmark & 0.426 & 0.434 & 0.349 & 0.423 & 0.347 & 0.576 \\
75 & SaProt (35M\_AF2, wt) & wt & \checkmark & 0.425 & 0.403 & 0.358 & 0.456 & 0.312 & 0.597 \\
76 & SaProt (650M AF2, wt) & wt & $\times$ & 0.424 & 0.437 & 0.343 & 0.459 & 0.317 & 0.564 \\
77 & ProtSSN (k=10, h=768) & wt & $\times$ & 0.424 & 0.441 & 0.339 & 0.418 & 0.370 & 0.550 \\
78 & ProGen2-M & tf & \checkmark & 0.423 & 0.451 & 0.315 & 0.427 & 0.408 & 0.514 \\
79 & ProGen2-B & tf & \checkmark & 0.422 & 0.454 & 0.303 & 0.431 & 0.407 & 0.514 \\
80 & ESM-IF1 & tf & $\times$ & 0.421 & 0.357 & 0.382 & 0.435 & 0.313 & 0.617 \\
81 & ProGen2-XL & tf & \checkmark & 0.419 & 0.446 & 0.317 & 0.415 & 0.407 & 0.508 \\
82 & ESM-2 650M (wt) & wt & $\times$ & 0.418 & 0.454 & 0.325 & 0.412 & 0.356 & 0.541 \\
83 & ProGen2-L & tf & \checkmark & 0.416 & 0.452 & 0.283 & 0.426 & 0.403 & 0.514 \\
84 & ESM-2 650M (mask) & mask & $\times$ & 0.415 & 0.431 & 0.337 & 0.415 & 0.368 & 0.523 \\
85 & CARP-38M & wt & \checkmark & 0.415 & 0.422 & 0.346 & 0.406 & 0.351 & 0.547 \\
86 & ProGen3-339M & tf & \checkmark & 0.414 & 0.431 & 0.340 & 0.388 & 0.398 & 0.514 \\
87 & ProGen3-762M & tf & \checkmark & 0.413 & 0.441 & 0.310 & 0.391 & 0.409 & 0.516 \\
88 & ProGen3-3B & tf & \checkmark & 0.413 & 0.437 & 0.292 & 0.416 & 0.400 & 0.521 \\
89 & ESM-1v (wt) & wt & $\times$ & 0.410 & 0.445 & 0.293 & 0.430 & 0.381 & 0.500 \\
90 & ESMC-300M & mask & $\times$ & 0.409 & 0.429 & 0.320 & 0.406 & 0.367 & 0.523 \\
91 & ProGen3-219M & tf & \checkmark & 0.409 & 0.422 & 0.347 & 0.377 & 0.391 & 0.508 \\
92 & SaProt (35M\_AF2, mask) & mask & $\times$ & 0.408 & 0.369 & 0.360 & 0.441 & 0.294 & 0.576 \\
93 & RITA-L & tf & \checkmark & 0.407 & 0.432 & 0.308 & 0.409 & 0.400 & 0.486 \\
94 & ESM-2 3B (mask) & mask & $\times$ & 0.407 & 0.423 & 0.321 & 0.403 & 0.378 & 0.509 \\
95 & ESMC-600M & mask & $\times$ & 0.407 & 0.427 & 0.295 & 0.413 & 0.372 & 0.528 \\
96 & ESM-1v (mask) & mask & $\times$ & 0.407 & 0.421 & 0.320 & 0.429 & 0.386 & 0.477 \\
97 & ESM-2 3B (wt) & wt & $\times$ & 0.406 & 0.440 & 0.300 & 0.404 & 0.366 & 0.521 \\
98 & ProGen2-S & tf & \checkmark & 0.406 & 0.425 & 0.328 & 0.381 & 0.385 & 0.512 \\
99 & ProGen3-1B & tf & \checkmark & 0.406 & 0.436 & 0.274 & 0.395 & 0.409 & 0.514 \\
100 & RITA-M & tf & \checkmark & 0.405 & 0.432 & 0.304 & 0.395 & 0.397 & 0.495 \\
101 & RITA-XL & tf & \checkmark & 0.401 & 0.423 & 0.298 & 0.418 & 0.386 & 0.482 \\
102 & ProGen3-112M & tf & \checkmark & 0.401 & 0.413 & 0.340 & 0.365 & 0.379 & 0.509 \\
103 & RITA-S & tf & \checkmark & 0.398 & 0.415 & 0.336 & 0.370 & 0.379 & 0.491 \\
104 & ESM-2 8M (mask) & mask & \checkmark & 0.395 & 0.389 & 0.322 & 0.380 & 0.347 & 0.538 \\
105 & ESM-1b (wt) & wt & $\times$ & 0.394 & 0.428 & 0.287 & 0.406 & 0.350 & 0.500 \\
106 & ESM-2 150M (wt) & wt & $\times$ & 0.393 & 0.422 & 0.316 & 0.400 & 0.297 & 0.528 \\
107 & CARP-640M & wt & $\times$ & 0.390 & 0.403 & 0.289 & 0.401 & 0.372 & 0.486 \\
108 & ProteinMPNN (v\_48\_020) & tf & $\times$ & 0.389 & 0.316 & 0.332 & 0.411 & 0.287 & 0.601 \\
109 & ESM-1b (mask) & mask & $\times$ & 0.389 & 0.406 & 0.297 & 0.409 & 0.354 & 0.480 \\
110 & ESM-2 150M (mask) & mask & $\times$ & 0.388 & 0.400 & 0.326 & 0.402 & 0.304 & 0.510 \\
111 & ESM-2 8M (wt) & wt & \checkmark & 0.387 & 0.376 & 0.323 & 0.401 & 0.307 & 0.531 \\
112 & ProteinMPNN (v\_48\_010) & tf & $\times$ & 0.385 & 0.316 & 0.329 & 0.410 & 0.284 & 0.584 \\
113 & ProteinMPNN (v\_48\_030) & tf & $\times$ & 0.381 & 0.309 & 0.330 & 0.404 & 0.278 & 0.584 \\
114 & ProteinMPNN (v\_48\_002) & tf & $\times$ & 0.378 & 0.318 & 0.312 & 0.396 & 0.276 & 0.587 \\
115 & SaProt (35M\_AF2, wt) & wt & $\times$ & 0.366 & 0.334 & 0.320 & 0.398 & 0.231 & 0.545 \\
116 & ProteinMPNN-Soluble (v\_48\_010) & tf & $\times$ & 0.363 & 0.295 & 0.317 & 0.336 & 0.283 & 0.587 \\
117 & ProteinMPNN-Soluble (v\_48\_020) & tf & $\times$ & 0.361 & 0.288 & 0.309 & 0.339 & 0.286 & 0.585 \\
118 & CARP-76M & wt & $\times$ & 0.360 & 0.378 & 0.290 & 0.372 & 0.273 & 0.487 \\
119 & ProteinMPNN-Soluble (v\_48\_002) & tf & $\times$ & 0.357 & 0.291 & 0.308 & 0.333 & 0.275 & 0.577 \\
120 & CARP-600K & wt & \checkmark & 0.354 & 0.358 & 0.251 & 0.378 & 0.279 & 0.507 \\
121 & ProteinMPNN-Soluble (v\_48\_030) & tf & $\times$ & 0.353 & 0.278 & 0.312 & 0.316 & 0.278 & 0.580 \\
122 & ProGen2-XL & tf & $\times$ & 0.351 & 0.374 & 0.269 & 0.351 & 0.371 & 0.387 \\
123 & ProGen2-M & tf & $\times$ & 0.337 & 0.353 & 0.254 & 0.376 & 0.364 & 0.339 \\
124 & ProGen3-3B & tf & $\times$ & 0.331 & 0.365 & 0.226 & 0.363 & 0.360 & 0.340 \\
125 & ProGen2-B & tf & $\times$ & 0.330 & 0.355 & 0.242 & 0.381 & 0.351 & 0.320 \\
126 & ESM-2 35M (wt) & wt & $\times$ & 0.329 & 0.333 & 0.282 & 0.335 & 0.208 & 0.487 \\
127 & ProGen2-L & tf & $\times$ & 0.327 & 0.368 & 0.216 & 0.373 & 0.355 & 0.323 \\
128 & ESM-2 35M (mask) & mask & $\times$ & 0.321 & 0.316 & 0.291 & 0.343 & 0.217 & 0.439 \\
129 & ProGen3-1B & tf & $\times$ & 0.320 & 0.354 & 0.218 & 0.347 & 0.365 & 0.318 \\
130 & CARP-38M & wt & $\times$ & 0.314 & 0.319 & 0.282 & 0.306 & 0.220 & 0.444 \\
131 & RITA-L & tf & $\times$ & 0.309 & 0.321 & 0.252 & 0.351 & 0.352 & 0.267 \\
132 & RITA-XL & tf & $\times$ & 0.305 & 0.316 & 0.233 & 0.360 & 0.340 & 0.277 \\
133 & ProGen3-762M & tf & $\times$ & 0.304 & 0.328 & 0.244 & 0.330 & 0.346 & 0.274 \\
134 & RITA-M & tf & $\times$ & 0.297 & 0.308 & 0.251 & 0.336 & 0.348 & 0.241 \\
135 & ProGen2-S & tf & $\times$ & 0.288 & 0.300 & 0.272 & 0.309 & 0.291 & 0.269 \\
136 & ProGen3-339M & tf & $\times$ & 0.286 & 0.304 & 0.280 & 0.311 & 0.319 & 0.217 \\
137 & ProGen3-219M & tf & $\times$ & 0.264 & 0.274 & 0.279 & 0.288 & 0.296 & 0.185 \\
138 & RITA-S & tf & $\times$ & 0.254 & 0.248 & 0.267 & 0.277 & 0.289 & 0.190 \\
139 & ESM-2 8M (wt) & wt & $\times$ & 0.251 & 0.217 & 0.244 & 0.264 & 0.131 & 0.399 \\
140 & ProGen3-112M & tf & $\times$ & 0.232 & 0.244 & 0.260 & 0.253 & 0.265 & 0.139 \\
141 & ESM-2 8M (mask) & mask & $\times$ & 0.223 & 0.191 & 0.260 & 0.266 & 0.139 & 0.262 \\
142 & CARP-600K & wt & $\times$ & 0.159 & 0.142 & 0.090 & 0.167 & 0.065 & 0.331 \\
\end{longtable}
}

Table~\ref{tab:leaderboard_vmh_category} retains the VenusMutHub task breakdown. Its within-benchmark ordering shows which configurations support each task beyond the mean over assays.

{\fontsize{8.5}{10}\selectfont
\setlength{\LTcapwidth}{\linewidth}
\makeatletter
\let\SavedLTCaption\LT@makecaption
\def\LT@makecaption#1#2#3{\SavedLTCaption{#1}{\normalsize#2}{\normalsize#3}}
\makeatother
\setlength{\tabcolsep}{1.5pt}
\setlength{\LTcapwidth}{\linewidth}
\renewcommand{\arraystretch}{1.08}
\begin{longtable}{@{}cllccccccc@{}}
\caption{VenusMutHub results before and after applying the harness, ranked by Overall score across 71 configurations. Scores are mean Spearman correlations over assays. Each configuration appears without ($\times$) and with ($\checkmark$) the harness. Scoring uses masked inputs (mask), unmasked WT inputs (wt), or teacher forcing (tf). Categories are the five assay tasks; the best value in each column is red.}\label{tab:leaderboard_vmh_category}\\
\toprule
Rank & Model & Scoring & \textsc{\HarnessShortName{}} & Overall & stability & activity & PPI bind. & selectivity & DTI bind. \\
\midrule
\endfirsthead
\multicolumn{10}{c}{\tablename\ \thetable\ continued}\\
\toprule
Rank & Model & Scoring & \textsc{\HarnessShortName{}} & Overall & stability & activity & PPI bind. & selectivity & DTI bind. \\
\midrule
\endhead
\midrule
\multicolumn{10}{r}{Continued on next page}\\
\endfoot
\bottomrule
\endlastfoot
1 & ProteinMPNN (v\_48\_020) & tf & \checkmark & \textcolor{red!75!black}{\textbf{0.271}} & \textcolor{red!75!black}{\textbf{0.399}} & 0.108 & 0.024 & 0.012 & 0.198 \\*
2 & ESM-IF1 & tf & \checkmark & 0.267 & 0.356 & 0.152 & 0.108 & 0.018 & 0.269 \\*
3 & ProteinMPNN-Soluble (v\_48\_020) & tf & \checkmark & 0.265 & 0.396 & 0.106 & 0.036 & 0.000 & 0.093 \\
4 & ProteinMPNN-Soluble (v\_48\_030) & tf & \checkmark & 0.265 & 0.383 & 0.120 & 0.039 & 0.022 & 0.168 \\
5 & ProteinMPNN (v\_48\_030) & tf & \checkmark & 0.259 & 0.388 & 0.106 & 0.019 & -0.020 & 0.137 \\
6 & \textbf{VenusREM2} & wt & \checkmark & 0.258 & 0.351 & 0.122 & 0.113 & -0.017 & 0.274 \\
7 & ProteinMPNN-Soluble (v\_48\_010) & tf & \checkmark & 0.256 & 0.383 & 0.104 & 0.023 & -0.025 & 0.133 \\
8 & ProteinMPNN (v\_48\_010) & tf & \checkmark & 0.256 & 0.383 & 0.095 & 0.038 & -0.023 & 0.122 \\
9 & ProteinMPNN-Soluble (v\_48\_002) & tf & \checkmark & 0.250 & 0.379 & 0.075 & 0.033 & -0.016 & 0.152 \\
10 & ProSST (K=2048) & wt & \checkmark & 0.249 & 0.348 & 0.118 & 0.103 & -0.048 & 0.204 \\
11 & ProteinMPNN (v\_48\_002) & tf & \checkmark & 0.244 & 0.370 & 0.069 & 0.021 & 0.020 & 0.134 \\
12 & ProSST (K=1024) & wt & \checkmark & 0.242 & 0.334 & 0.120 & 0.083 & -0.014 & 0.236 \\
13 & SaProt (650M\_PDB, mask) & mask & \checkmark & 0.240 & 0.319 & 0.139 & 0.090 & -0.014 & 0.282 \\
14 & SaProt (650M AF2, mask) & mask & \checkmark & 0.240 & 0.316 & 0.161 & 0.091 & 0.004 & 0.200 \\
15 & ProSST (K=20) & wt & \checkmark & 0.238 & 0.327 & 0.083 & 0.122 & -0.036 & 0.312 \\
16 & ProSST (K=4096) & wt & \checkmark & 0.237 & 0.324 & 0.117 & 0.096 & \textcolor{red!75!black}{\textbf{0.031}} & 0.177 \\
17 & SaProt (35M\_AF2, mask) & mask & \checkmark & 0.237 & 0.312 & 0.171 & 0.068 & -0.040 & 0.256 \\
18 & S3F (mask) & mask & \checkmark & 0.234 & 0.316 & 0.140 & 0.095 & -0.030 & 0.195 \\
19 & ProSST (K=128) & wt & \checkmark & 0.230 & 0.305 & 0.117 & 0.121 & -0.015 & 0.261 \\
20 & MIF-ST & tf & \checkmark & 0.230 & 0.352 & 0.044 & 0.079 & -0.106 & 0.169 \\
21 & ESM3 & mask & \checkmark & 0.226 & 0.302 & 0.150 & 0.106 & -0.034 & 0.142 \\
22 & ProSST (K=512) & wt & \checkmark & 0.226 & 0.313 & 0.102 & 0.083 & -0.037 & 0.264 \\
23 & ESM-2 150M (mask) & mask & \checkmark & 0.225 & 0.291 & 0.173 & 0.079 & -0.002 & 0.206 \\
24 & S3F (wt) & wt & \checkmark & 0.224 & 0.304 & 0.111 & 0.123 & -0.022 & 0.175 \\
25 & ESM-2 3B (mask) & mask & \checkmark & 0.223 & 0.277 & \textcolor{red!75!black}{\textbf{0.180}} & 0.113 & -0.081 & 0.295 \\
26 & ProteinMPNN-Soluble (v\_48\_030) & tf & $\times$ & 0.222 & 0.357 & 0.026 & -0.030 & 0.030 & 0.131 \\
27 & ProteinMPNN (v\_48\_020) & tf & $\times$ & 0.221 & 0.366 & 0.006 & -0.033 & -0.001 & 0.116 \\
28 & ProteinMPNN (v\_48\_010) & tf & $\times$ & 0.218 & 0.360 & 0.013 & -0.034 & -0.050 & 0.155 \\
29 & ESM-2 150M (wt) & wt & \checkmark & 0.216 & 0.288 & 0.127 & 0.110 & -0.015 & 0.186 \\
30 & ESM-IF1 & tf & $\times$ & 0.216 & 0.328 & 0.074 & 0.040 & 0.018 & 0.021 \\
31 & ProteinMPNN-Soluble (v\_48\_020) & tf & $\times$ & 0.216 & 0.367 & 0.004 & -0.049 & 0.000 & 0.031 \\
32 & SaProt (650M\_PDB, wt) & wt & \checkmark & 0.215 & 0.304 & 0.059 & 0.078 & -0.034 & 0.319 \\
33 & ESM-2 650M (mask) & mask & \checkmark & 0.213 & 0.280 & 0.149 & 0.100 & -0.063 & 0.197 \\
34 & ProSST (K=2048) & wt & $\times$ & 0.213 & 0.315 & 0.076 & 0.064 & -0.065 & 0.137 \\
35 & ESM-2 35M (mask) & mask & \checkmark & 0.212 & 0.280 & 0.158 & 0.049 & 0.003 & 0.185 \\
36 & ProteinMPNN-Soluble (v\_48\_010) & tf & $\times$ & 0.212 & 0.357 & -0.004 & -0.042 & -0.070 & 0.167 \\
37 & S3F (mask) & mask & $\times$ & 0.211 & 0.295 & 0.109 & 0.043 & -0.007 & 0.203 \\
38 & ESM-1b (mask) & mask & \checkmark & 0.211 & 0.265 & 0.163 & 0.126 & -0.048 & 0.211 \\
39 & ProteinMPNN (v\_48\_030) & tf & $\times$ & 0.210 & 0.357 & 0.009 & -0.060 & -0.022 & 0.067 \\
40 & SaProt (650M AF2, mask) & mask & $\times$ & 0.207 & 0.275 & 0.152 & 0.058 & 0.014 & 0.143 \\
41 & ESM-1v (mask) & mask & \checkmark & 0.206 & 0.268 & 0.165 & 0.081 & -0.082 & 0.201 \\
42 & CARP-640M & wt & \checkmark & 0.206 & 0.271 & 0.151 & 0.076 & -0.094 & 0.236 \\
43 & SaProt (650M\_PDB, mask) & mask & $\times$ & 0.205 & 0.275 & 0.116 & 0.048 & -0.013 & 0.285 \\
44 & SaProt (650M AF2, wt) & wt & \checkmark & 0.202 & 0.286 & 0.093 & 0.057 & -0.035 & 0.175 \\
45 & ProteinMPNN-Soluble (v\_48\_002) & tf & $\times$ & 0.198 & 0.340 & -0.021 & -0.039 & -0.054 & 0.135 \\
46 & ESMC-300M & mask & \checkmark & 0.198 & 0.254 & 0.151 & 0.106 & -0.059 & 0.181 \\
47 & ESM-2 8M (mask) & mask & \checkmark & 0.198 & 0.256 & 0.144 & 0.067 & -0.045 & 0.248 \\
48 & ProtSSN (k=20, h=1280) & wt & \checkmark & 0.198 & 0.282 & 0.072 & 0.083 & -0.034 & 0.162 \\
49 & ProtSSN (k=10, h=512) & wt & \checkmark & 0.197 & 0.286 & 0.064 & 0.079 & -0.050 & 0.164 \\
50 & MIF-ST & tf & $\times$ & 0.196 & 0.317 & 0.017 & 0.029 & -0.110 & 0.135 \\
51 & ProtSSN & wt & \checkmark & 0.195 & 0.281 & 0.079 & 0.084 & -0.050 & 0.110 \\
52 & CARP-76M & wt & \checkmark & 0.194 & 0.259 & 0.119 & 0.083 & -0.034 & 0.196 \\
53 & ProGen2-B & tf & \checkmark & 0.194 & 0.252 & 0.142 & 0.100 & -0.089 & 0.195 \\
54 & ESMC-600M & mask & \checkmark & 0.194 & 0.248 & 0.142 & 0.108 & -0.069 & 0.210 \\
55 & ESM-2 3B (wt) & wt & \checkmark & 0.194 & 0.262 & 0.100 & 0.082 & -0.060 & 0.242 \\
56 & ESM-1b (wt) & wt & \checkmark & 0.193 & 0.255 & 0.098 & \textcolor{red!75!black}{\textbf{0.131}} & -0.069 & 0.246 \\
57 & RITA-L & tf & \checkmark & 0.193 & 0.250 & 0.138 & 0.101 & -0.056 & 0.188 \\
58 & S3F (wt) & wt & $\times$ & 0.193 & 0.271 & 0.082 & 0.068 & -0.024 & 0.182 \\
59 & ProGen2-S & tf & \checkmark & 0.192 & 0.246 & 0.153 & 0.086 & -0.070 & 0.209 \\
60 & ProtSSN (k=20, h=512) & wt & \checkmark & 0.192 & 0.270 & 0.078 & 0.095 & -0.046 & 0.156 \\
61 & ProtSSN (k=10, h=1280) & wt & \checkmark & 0.192 & 0.279 & 0.072 & 0.090 & -0.044 & 0.070 \\
62 & ProtSSN (k=30, h=512) & wt & \checkmark & 0.191 & 0.276 & 0.073 & 0.077 & -0.055 & 0.142 \\
63 & ProteinMPNN (v\_48\_002) & tf & $\times$ & 0.191 & 0.328 & -0.021 & -0.032 & 0.001 & 0.064 \\
64 & ProtSSN (k=20, h=768) & wt & \checkmark & 0.191 & 0.271 & 0.078 & 0.098 & -0.044 & 0.110 \\
65 & ProGen3-219M & tf & \checkmark & 0.190 & 0.242 & 0.155 & 0.101 & -0.059 & 0.157 \\
66 & VenusREM2 & wt & $\times$ & 0.190 & 0.294 & 0.018 & 0.006 & -0.049 & 0.267 \\
67 & ProGen3-339M & tf & \checkmark & 0.189 & 0.242 & 0.162 & 0.081 & -0.071 & 0.169 \\
68 & ProGen2-L & tf & \checkmark & 0.188 & 0.243 & 0.146 & 0.084 & -0.088 & 0.224 \\
69 & ESM-1v (wt) & wt & \checkmark & 0.188 & 0.256 & 0.112 & 0.097 & -0.068 & 0.133 \\
70 & SaProt (35M\_AF2, mask) & mask & $\times$ & 0.188 & 0.258 & 0.153 & -0.017 & -0.047 & 0.180 \\
71 & ESM-2 650M (wt) & wt & \checkmark & 0.187 & 0.259 & 0.089 & 0.081 & -0.056 & 0.188 \\
72 & RITA-M & tf & \checkmark & 0.186 & 0.239 & 0.138 & 0.115 & -0.101 & 0.203 \\
73 & ESM-2 35M (wt) & wt & \checkmark & 0.186 & 0.265 & 0.092 & 0.055 & -0.024 & 0.120 \\
74 & ESM3 & mask & $\times$ & 0.185 & 0.264 & 0.128 & -0.002 & -0.032 & 0.106 \\
75 & ProtSSN (k=30, h=768) & wt & \checkmark & 0.185 & 0.262 & 0.077 & 0.092 & -0.042 & 0.113 \\
76 & ProtSSN (k=30, h=1280) & wt & \checkmark & 0.183 & 0.262 & 0.055 & 0.095 & -0.042 & 0.167 \\
77 & ESM-2 3B (mask) & mask & $\times$ & 0.182 & 0.228 & 0.161 & 0.058 & -0.082 & 0.274 \\
78 & SaProt (650M\_PDB, wt) & wt & $\times$ & 0.182 & 0.256 & 0.050 & 0.033 & -0.025 & 0.363 \\
79 & RITA-S & tf & \checkmark & 0.182 & 0.228 & 0.134 & 0.121 & -0.041 & 0.177 \\
80 & ProGen2-M & tf & \checkmark & 0.182 & 0.240 & 0.130 & 0.094 & -0.117 & 0.187 \\
81 & ProGen3-762M & tf & \checkmark & 0.180 & 0.228 & 0.159 & 0.099 & -0.077 & 0.122 \\
82 & SaProt (35M\_AF2, wt) & wt & \checkmark & 0.179 & 0.250 & 0.098 & 0.033 & -0.023 & 0.171 \\
83 & ProtSSN (k=10, h=768) & wt & \checkmark & 0.178 & 0.258 & 0.060 & 0.077 & -0.041 & 0.118 \\
84 & SaProt (650M AF2, wt) & wt & $\times$ & 0.178 & 0.249 & 0.092 & 0.031 & -0.023 & 0.190 \\
85 & ProGen3-112M & tf & \checkmark & 0.177 & 0.219 & 0.130 & 0.121 & -0.038 & 0.203 \\
86 & ESM-2 8M (wt) & wt & \checkmark & 0.176 & 0.239 & 0.079 & 0.047 & -0.052 & 0.318 \\
87 & ProGen2-XL & tf & \checkmark & 0.176 & 0.232 & 0.136 & 0.095 & -0.111 & 0.131 \\
88 & ProtSSN & wt & $\times$ & 0.174 & 0.251 & 0.079 & 0.056 & -0.048 & 0.111 \\
89 & ProtSSN (k=20, h=1280) & wt & $\times$ & 0.174 & 0.247 & 0.073 & 0.067 & -0.035 & 0.138 \\
90 & ProGen3-1B & tf & \checkmark & 0.173 & 0.220 & 0.127 & 0.121 & -0.096 & 0.175 \\
91 & ProGen3-3B & tf & \checkmark & 0.170 & 0.228 & 0.110 & 0.113 & -0.115 & 0.126 \\
92 & ESM-2 650M (mask) & mask & $\times$ & 0.169 & 0.224 & 0.134 & 0.019 & -0.054 & 0.220 \\
93 & ProtSSN (k=30, h=512) & wt & $\times$ & 0.169 & 0.244 & 0.076 & 0.030 & -0.057 & 0.182 \\
94 & ProtSSN (k=10, h=512) & wt & $\times$ & 0.169 & 0.247 & 0.061 & 0.044 & -0.053 & 0.168 \\
95 & ProtSSN (k=20, h=512) & wt & $\times$ & 0.168 & 0.236 & 0.076 & 0.064 & -0.043 & 0.171 \\
96 & ESM-2 150M (mask) & mask & $\times$ & 0.167 & 0.221 & 0.144 & -0.004 & -0.033 & 0.190 \\
97 & ProtSSN (k=10, h=1280) & wt & $\times$ & 0.166 & 0.245 & 0.073 & 0.051 & -0.047 & 0.062 \\
98 & ESM-1b (mask) & mask & $\times$ & 0.166 & 0.214 & 0.129 & 0.067 & -0.048 & 0.174 \\
99 & ProtSSN (k=20, h=768) & wt & $\times$ & 0.165 & 0.233 & 0.079 & 0.077 & -0.046 & 0.109 \\
100 & ESM-2 150M (wt) & wt & $\times$ & 0.165 & 0.223 & 0.120 & -0.004 & -0.029 & 0.225 \\
101 & ProtSSN (k=30, h=1280) & wt & $\times$ & 0.162 & 0.232 & 0.056 & 0.061 & -0.044 & 0.169 \\
102 & ProSST (K=1024) & wt & $\times$ & 0.162 & 0.265 & -0.016 & 0.004 & -0.043 & 0.184 \\
103 & ESM-1v (mask) & mask & $\times$ & 0.162 & 0.218 & 0.141 & 0.003 & -0.093 & 0.180 \\
104 & ProSST (K=4096) & wt & $\times$ & 0.161 & 0.250 & 0.013 & 0.004 & 0.029 & 0.155 \\
105 & ESM-2 3B (wt) & wt & $\times$ & 0.161 & 0.218 & 0.092 & 0.030 & -0.058 & 0.269 \\
106 & RITA-XL & tf & \checkmark & 0.160 & 0.216 & 0.112 & 0.069 & -0.100 & 0.162 \\
107 & ProtSSN (k=30, h=768) & wt & $\times$ & 0.160 & 0.230 & 0.075 & 0.056 & -0.050 & 0.097 \\
108 & CARP-38M & wt & \checkmark & 0.158 & 0.216 & 0.075 & 0.065 & -0.042 & 0.205 \\
109 & CARP-640M & wt & $\times$ & 0.157 & 0.221 & 0.116 & 0.005 & -0.119 & 0.187 \\
110 & ESMC-600M & mask & $\times$ & 0.155 & 0.205 & 0.114 & 0.060 & -0.074 & 0.164 \\
111 & ESM-1v (wt) & wt & $\times$ & 0.153 & 0.212 & 0.109 & 0.015 & -0.075 & 0.172 \\
112 & ESM-1b (wt) & wt & $\times$ & 0.153 & 0.201 & 0.092 & 0.072 & -0.057 & 0.209 \\
113 & ESM-2 650M (wt) & wt & $\times$ & 0.152 & 0.210 & 0.087 & 0.015 & -0.060 & 0.241 \\
114 & ProtSSN (k=10, h=768) & wt & $\times$ & 0.152 & 0.221 & 0.052 & 0.053 & -0.037 & 0.132 \\
115 & CARP-600K & wt & \checkmark & 0.151 & 0.212 & 0.040 & 0.074 & -0.071 & 0.260 \\
116 & ESMC-300M & mask & $\times$ & 0.144 & 0.195 & 0.123 & 0.011 & -0.073 & 0.132 \\
117 & ProSST (K=20) & wt & $\times$ & 0.143 & 0.230 & -0.031 & 0.013 & -0.087 & 0.300 \\
118 & SaProt (35M\_AF2, wt) & wt & $\times$ & 0.142 & 0.201 & 0.086 & -0.031 & -0.037 & 0.234 \\
119 & ProSST (K=128) & wt & $\times$ & 0.139 & 0.225 & -0.020 & 0.009 & -0.044 & 0.204 \\
120 & CARP-76M & wt & $\times$ & 0.131 & 0.182 & 0.092 & -0.013 & -0.033 & 0.168 \\
121 & ProGen2-B & tf & $\times$ & 0.124 & 0.172 & 0.111 & 0.004 & -0.109 & 0.114 \\
122 & ProSST (K=512) & wt & $\times$ & 0.124 & 0.219 & -0.045 & -0.017 & -0.089 & 0.180 \\
123 & ProGen2-XL & tf & $\times$ & 0.123 & 0.170 & 0.105 & 0.030 & -0.132 & 0.113 \\
124 & ESM-2 35M (mask) & mask & $\times$ & 0.122 & 0.179 & 0.106 & -0.095 & 0.025 & 0.078 \\
125 & ProGen2-L & tf & $\times$ & 0.120 & 0.157 & 0.114 & -0.011 & -0.108 & 0.227 \\
126 & ProGen2-M & tf & $\times$ & 0.115 & 0.168 & 0.086 & -0.004 & -0.129 & 0.104 \\
127 & ProGen3-762M & tf & $\times$ & 0.111 & 0.132 & 0.178 & -0.020 & -0.087 & 0.090 \\
128 & RITA-L & tf & $\times$ & 0.110 & 0.158 & 0.091 & -0.018 & -0.066 & 0.080 \\
129 & ProGen3-339M & tf & $\times$ & 0.110 & 0.133 & 0.173 & -0.051 & -0.077 & 0.146 \\
130 & ProGen3-3B & tf & $\times$ & 0.110 & 0.160 & 0.082 & 0.039 & -0.123 & 0.009 \\
131 & ESM-2 35M (wt) & wt & $\times$ & 0.110 & 0.169 & 0.065 & -0.092 & -0.037 & 0.175 \\
132 & ProGen3-1B & tf & $\times$ & 0.108 & 0.151 & 0.112 & 0.013 & -0.099 & 0.001 \\
133 & RITA-M & tf & $\times$ & 0.100 & 0.137 & 0.081 & -0.021 & -0.068 & 0.184 \\
134 & RITA-XL & tf & $\times$ & 0.095 & 0.132 & 0.081 & -0.011 & -0.100 & 0.142 \\
135 & ProGen3-219M & tf & $\times$ & 0.092 & 0.122 & 0.111 & -0.071 & -0.070 & 0.192 \\
136 & ProGen2-S & tf & $\times$ & 0.091 & 0.137 & 0.067 & -0.053 & -0.066 & 0.127 \\
137 & ESM-2 8M (mask) & mask & $\times$ & 0.079 & 0.110 & 0.055 & -0.085 & -0.057 & 0.317 \\
138 & CARP-38M & wt & $\times$ & 0.073 & 0.108 & 0.020 & -0.042 & -0.054 & 0.245 \\
139 & ProGen3-112M & tf & $\times$ & 0.072 & 0.086 & 0.105 & -0.048 & -0.044 & 0.156 \\
140 & ESM-2 8M (wt) & wt & $\times$ & 0.065 & 0.102 & 0.017 & -0.115 & -0.076 & \textcolor{red!75!black}{\textbf{0.368}} \\
141 & RITA-S & tf & $\times$ & 0.056 & 0.068 & 0.083 & -0.038 & -0.064 & 0.143 \\
142 & CARP-600K & wt & $\times$ & 0.021 & 0.057 & -0.048 & -0.104 & -0.103 & 0.272 \\
\end{longtable}
}

Table~\ref{tab:leaderboard_vvh_category} retains the viral phenotype breakdown under hierarchical aggregation. Low immune-escape correlations remain visible alongside stronger intrinsic-fitness and expression scores.

{\fontsize{8.5}{10}\selectfont
\setlength{\LTcapwidth}{\linewidth}
\makeatletter
\let\SavedLTCaption\LT@makecaption
\def\LT@makecaption#1#2#3{\SavedLTCaption{#1}{\normalsize#2}{\normalsize#3}}
\makeatother
\setlength{\tabcolsep}{0.9pt}
\setlength{\LTcapwidth}{\linewidth}
\renewcommand{\arraystretch}{1.08}
\begin{longtable}{@{}c>{\raggedright\arraybackslash}p{0.25\textwidth}lccccccccc@{}}
\caption{\ViroBenchmarkName{} results before and after applying the harness, ranked by Overall score across 71 configurations. WT controls are excluded. Spearman correlations are averaged within groups sharing a phenotype and WT protein, then across proteins within each phenotype, and finally across the seven phenotypes. Each configuration appears without ($\times$) and with ($\checkmark$) the harness. Scoring uses masked inputs (mask), unmasked WT inputs (wt), or teacher forcing (tf). The best value in each column is red.}\label{tab:leaderboard_vvh_category}\\
\toprule
Rank & Model & Scoring & \textsc{\HarnessShortName{}} & Overall & fitness & activity & expression & \shortstack{cell\\entry} & binding & stability & \shortstack{immune\\esc.} \\
\midrule
\endfirsthead
\multicolumn{12}{c}{\tablename\ \thetable\ continued}\\
\toprule
Rank & Model & Scoring & \textsc{\HarnessShortName{}} & Overall & fitness & activity & expression & \shortstack{cell\\entry} & binding & stability & \shortstack{immune\\esc.} \\
\midrule
\endhead
\midrule
\multicolumn{12}{r}{Continued on next page}\\
\endfoot
\bottomrule
\endlastfoot
1 & SaProt (650M\_PDB, mask) & mask & \checkmark & \textcolor{red!75!black}{\textbf{0.320}} & 0.394 & 0.577 & 0.569 & 0.484 & 0.145 & 0.045 & 0.026 \\*
2 & S3F (wt) & wt & \checkmark & 0.318 & 0.404 & 0.582 & 0.540 & \textcolor{red!75!black}{\textbf{0.493}} & 0.111 & 0.052 & \textcolor{red!75!black}{\textbf{0.046}} \\*
3 & ESM-IF1 & tf & \checkmark & 0.318 & 0.359 & 0.603 & 0.566 & 0.439 & \textcolor{red!75!black}{\textbf{0.173}} & 0.060 & 0.026 \\
4 & MIF-ST & tf & \checkmark & 0.317 & 0.391 & 0.542 & 0.553 & 0.476 & 0.144 & 0.068 & 0.042 \\
5 & ProSST (K=2048) & wt & \checkmark & 0.310 & 0.366 & 0.570 & 0.561 & 0.469 & 0.126 & 0.048 & 0.032 \\
6 & SaProt (650M\_PDB, wt) & wt & \checkmark & 0.307 & 0.343 & \textcolor{red!75!black}{\textbf{0.608}} & 0.564 & 0.432 & 0.131 & 0.052 & 0.022 \\
7 & ProteinMPNN (v\_48\_010) & tf & \checkmark & 0.306 & 0.352 & 0.544 & 0.565 & 0.441 & 0.143 & 0.066 & 0.033 \\
8 & SaProt (650M\_PDB, mask) & mask & $\times$ & 0.306 & 0.360 & 0.606 & 0.545 & 0.445 & 0.130 & 0.045 & 0.015 \\
9 & ProteinMPNN (v\_48\_020) & tf & \checkmark & 0.306 & 0.357 & 0.536 & 0.567 & 0.444 & 0.144 & 0.060 & 0.032 \\
10 & ProteinMPNN (v\_48\_002) & tf & \checkmark & 0.305 & 0.346 & 0.543 & 0.567 & 0.438 & 0.146 & 0.064 & 0.034 \\
11 & ProteinMPNN-Soluble (v\_48\_002) & tf & \checkmark & 0.305 & 0.347 & 0.545 & 0.560 & 0.438 & 0.145 & 0.066 & 0.036 \\
12 & ProteinMPNN-Soluble (v\_48\_020) & tf & \checkmark & 0.304 & 0.360 & 0.528 & 0.559 & 0.440 & 0.145 & 0.061 & 0.035 \\
13 & ProteinMPNN (v\_48\_030) & tf & \checkmark & 0.304 & 0.352 & 0.538 & 0.570 & 0.441 & 0.142 & 0.049 & 0.035 \\
14 & ProteinMPNN-Soluble (v\_48\_010) & tf & \checkmark & 0.304 & 0.345 & 0.541 & 0.557 & 0.438 & 0.146 & 0.064 & 0.035 \\
15 & S3F (mask) & mask & \checkmark & 0.302 & \textcolor{red!75!black}{\textbf{0.414}} & 0.511 & 0.509 & 0.475 & 0.118 & 0.053 & 0.036 \\
16 & ProteinMPNN-Soluble (v\_48\_030) & tf & \checkmark & 0.299 & 0.357 & 0.514 & 0.556 & 0.438 & 0.137 & 0.055 & 0.035 \\
17 & VenusREM2 & wt & \checkmark & 0.297 & 0.353 & 0.480 & \textcolor{red!75!black}{\textbf{0.580}} & 0.462 & 0.126 & 0.046 & 0.033 \\
18 & ProtSSN & wt & \checkmark & 0.294 & 0.357 & 0.565 & 0.502 & 0.435 & 0.117 & 0.044 & 0.036 \\
19 & ProtSSN (k=20, h=512) & wt & \checkmark & 0.291 & 0.360 & 0.545 & 0.502 & 0.435 & 0.119 & 0.042 & 0.033 \\
20 & ProtSSN (k=20, h=1280) & wt & \checkmark & 0.289 & 0.352 & 0.547 & 0.499 & 0.426 & 0.114 & 0.047 & 0.034 \\
21 & MIF-ST & tf & $\times$ & 0.288 & 0.374 & 0.480 & 0.485 & 0.448 & 0.114 & 0.074 & 0.044 \\
22 & ProtSSN (k=30, h=512) & wt & \checkmark & 0.288 & 0.357 & 0.560 & 0.484 & 0.430 & 0.110 & 0.037 & 0.037 \\
23 & ProtSSN (k=20, h=768) & wt & \checkmark & 0.287 & 0.352 & 0.551 & 0.485 & 0.428 & 0.114 & 0.046 & 0.034 \\
24 & ProtSSN (k=30, h=768) & wt & \checkmark & 0.286 & 0.357 & 0.547 & 0.487 & 0.421 & 0.111 & 0.044 & 0.033 \\
25 & ESM-IF1 & tf & $\times$ & 0.285 & 0.248 & 0.606 & 0.542 & 0.343 & 0.169 & 0.065 & 0.024 \\
26 & ProtSSN (k=10, h=1280) & wt & \checkmark & 0.284 & 0.352 & 0.544 & 0.466 & 0.426 & 0.113 & 0.052 & 0.035 \\
27 & SaProt (650M AF2, mask) & mask & \checkmark & 0.281 & 0.360 & 0.501 & 0.495 & 0.426 & 0.118 & 0.035 & 0.032 \\
28 & S3F (wt) & wt & $\times$ & 0.279 & 0.377 & 0.549 & 0.451 & 0.433 & 0.055 & 0.047 & 0.044 \\
29 & ProtSSN (k=10, h=768) & wt & \checkmark & 0.279 & 0.347 & 0.522 & 0.472 & 0.413 & 0.118 & 0.044 & 0.037 \\
30 & ProtSSN (k=30, h=1280) & wt & \checkmark & 0.279 & 0.346 & 0.533 & 0.474 & 0.417 & 0.107 & 0.036 & 0.037 \\
31 & ProSST (K=1024) & wt & \checkmark & 0.275 & 0.329 & 0.436 & 0.538 & 0.425 & 0.116 & 0.056 & 0.028 \\
32 & SaProt (650M\_PDB, wt) & wt & $\times$ & 0.273 & 0.312 & 0.546 & 0.492 & 0.395 & 0.098 & 0.057 & 0.013 \\
33 & ProtSSN (k=10, h=512) & wt & \checkmark & 0.272 & 0.333 & 0.523 & 0.447 & 0.410 & 0.106 & 0.049 & 0.034 \\
34 & ProSST (K=4096) & wt & \checkmark & 0.271 & 0.315 & 0.405 & 0.548 & 0.429 & 0.121 & 0.045 & 0.035 \\
35 & ProSST (K=2048) & wt & $\times$ & 0.271 & 0.307 & 0.550 & 0.497 & 0.407 & 0.080 & 0.026 & 0.030 \\
36 & ProteinMPNN (v\_48\_010) & tf & $\times$ & 0.271 & 0.250 & 0.538 & 0.501 & 0.383 & 0.121 & 0.067 & 0.035 \\
37 & ProteinMPNN (v\_48\_020) & tf & $\times$ & 0.270 & 0.256 & 0.526 & 0.507 & 0.386 & 0.123 & 0.059 & 0.033 \\
38 & ProteinMPNN (v\_48\_002) & tf & $\times$ & 0.269 & 0.241 & 0.533 & 0.508 & 0.377 & 0.126 & 0.065 & 0.036 \\
39 & S3F (mask) & mask & $\times$ & 0.269 & 0.373 & 0.477 & 0.442 & 0.433 & 0.077 & 0.049 & 0.036 \\
40 & ProSST (K=512) & wt & \checkmark & 0.269 & 0.324 & 0.418 & 0.543 & 0.418 & 0.111 & 0.035 & 0.033 \\
41 & ProteinMPNN-Soluble (v\_48\_002) & tf & $\times$ & 0.269 & 0.238 & 0.532 & 0.502 & 0.377 & 0.122 & 0.069 & 0.040 \\
42 & ProteinMPNN-Soluble (v\_48\_010) & tf & $\times$ & 0.268 & 0.241 & 0.533 & 0.499 & 0.377 & 0.122 & 0.066 & 0.037 \\
43 & ProteinMPNN (v\_48\_030) & tf & $\times$ & 0.266 & 0.241 & 0.529 & 0.515 & 0.374 & 0.119 & 0.045 & 0.038 \\
44 & ProteinMPNN-Soluble (v\_48\_020) & tf & $\times$ & 0.265 & 0.249 & 0.515 & 0.494 & 0.376 & 0.120 & 0.060 & 0.038 \\
45 & ProSST (K=128) & wt & \checkmark & 0.264 & 0.326 & 0.395 & 0.519 & 0.415 & 0.120 & 0.048 & 0.028 \\
46 & ESM-2 3B (wt) & wt & \checkmark & 0.260 & 0.362 & 0.521 & 0.391 & 0.396 & 0.073 & 0.050 & 0.027 \\
47 & ProSST (K=20) & wt & \checkmark & 0.259 & 0.312 & 0.413 & 0.528 & 0.396 & 0.119 & 0.020 & 0.028 \\
48 & SaProt (650M AF2, wt) & wt & \checkmark & 0.259 & 0.268 & 0.542 & 0.487 & 0.333 & 0.100 & 0.049 & 0.031 \\
49 & ProteinMPNN-Soluble (v\_48\_030) & tf & $\times$ & 0.256 & 0.237 & 0.497 & 0.489 & 0.366 & 0.109 & 0.054 & 0.037 \\
50 & ESM-2 650M (wt) & wt & \checkmark & 0.252 & 0.364 & 0.457 & 0.392 & 0.369 & 0.095 & 0.053 & 0.036 \\
51 & ESM3 & mask & \checkmark & 0.249 & 0.320 & 0.462 & 0.406 & 0.369 & 0.099 & 0.059 & 0.030 \\
52 & ESM-1b (wt) & wt & \checkmark & 0.248 & 0.320 & 0.445 & 0.404 & 0.348 & 0.098 & 0.086 & 0.036 \\
53 & RITA-XL & tf & \checkmark & 0.247 & 0.359 & 0.420 & 0.406 & 0.371 & 0.107 & 0.058 & 0.010 \\
54 & RITA-L & tf & \checkmark & 0.247 & 0.363 & 0.449 & 0.390 & 0.356 & 0.113 & 0.052 & 0.007 \\
55 & SaProt (35M\_AF2, mask) & mask & \checkmark & 0.245 & 0.325 & 0.403 & 0.424 & 0.382 & 0.097 & 0.049 & 0.032 \\
56 & ESM-2 3B (mask) & mask & \checkmark & 0.243 & 0.393 & 0.456 & 0.314 & 0.403 & 0.050 & 0.053 & 0.036 \\
57 & ESM-1v (wt) & wt & \checkmark & 0.238 & 0.287 & 0.396 & 0.405 & 0.363 & 0.114 & 0.069 & 0.029 \\
58 & RITA-M & tf & \checkmark & 0.237 & 0.370 & 0.420 & 0.374 & 0.352 & 0.089 & 0.058 & -0.007 \\
59 & VenusREM2 & wt & $\times$ & 0.236 & 0.242 & 0.420 & 0.500 & 0.351 & 0.072 & 0.034 & 0.032 \\
60 & RITA-S & tf & \checkmark & 0.236 & 0.378 & 0.420 & 0.354 & 0.345 & 0.092 & 0.063 & -0.004 \\
61 & ProGen2-XL & tf & \checkmark & 0.235 & 0.350 & 0.392 & 0.396 & 0.367 & 0.106 & 0.039 & -0.003 \\
62 & ESM-2 150M (wt) & wt & \checkmark & 0.235 & 0.291 & 0.451 & 0.395 & 0.308 & 0.107 & 0.062 & 0.032 \\
63 & ProtSSN (k=20, h=512) & wt & $\times$ & 0.233 & 0.352 & 0.420 & 0.361 & 0.381 & 0.050 & 0.037 & 0.033 \\
64 & ProtSSN & wt & $\times$ & 0.233 & 0.352 & 0.434 & 0.350 & 0.376 & 0.042 & 0.041 & 0.036 \\
65 & ProtSSN (k=20, h=1280) & wt & $\times$ & 0.230 & 0.347 & 0.419 & 0.357 & 0.366 & 0.044 & 0.044 & 0.034 \\
66 & CARP-76M & wt & \checkmark & 0.230 & 0.265 & 0.405 & 0.407 & 0.316 & 0.109 & 0.076 & 0.029 \\
67 & ESMC-600M & mask & \checkmark & 0.228 & 0.318 & 0.409 & 0.335 & 0.365 & 0.071 & 0.059 & 0.041 \\
68 & ESM-2 650M (mask) & mask & \checkmark & 0.227 & 0.388 & 0.376 & 0.309 & 0.372 & 0.058 & 0.049 & 0.041 \\
69 & ProtSSN (k=20, h=768) & wt & $\times$ & 0.226 & 0.345 & 0.420 & 0.329 & 0.367 & 0.042 & 0.044 & 0.034 \\
70 & ProtSSN (k=30, h=512) & wt & $\times$ & 0.225 & 0.347 & 0.433 & 0.325 & 0.369 & 0.032 & 0.033 & 0.038 \\
71 & ProtSSN (k=30, h=768) & wt & $\times$ & 0.223 & 0.350 & 0.413 & 0.335 & 0.357 & 0.037 & 0.041 & 0.031 \\
72 & ESM-1b (mask) & mask & \checkmark & 0.223 & 0.355 & 0.362 & 0.314 & 0.355 & 0.063 & 0.075 & 0.039 \\
73 & SaProt (650M AF2, mask) & mask & $\times$ & 0.222 & 0.255 & 0.477 & 0.398 & 0.299 & 0.075 & 0.025 & 0.026 \\
74 & CARP-38M & wt & \checkmark & 0.222 & 0.244 & 0.378 & 0.404 & 0.312 & 0.112 & 0.073 & 0.030 \\
75 & ESMC-300M & mask & \checkmark & 0.220 & 0.311 & 0.370 & 0.339 & 0.356 & 0.070 & 0.054 & 0.038 \\
76 & ProtSSN (k=10, h=1280) & wt & $\times$ & 0.220 & 0.344 & 0.407 & 0.303 & 0.363 & 0.039 & 0.047 & 0.036 \\
77 & CARP-640M & wt & \checkmark & 0.215 & 0.374 & 0.286 & 0.299 & 0.361 & 0.065 & 0.087 & 0.034 \\
78 & ProtSSN (k=10, h=768) & wt & $\times$ & 0.214 & 0.335 & 0.384 & 0.309 & 0.345 & 0.045 & 0.040 & 0.038 \\
79 & ESM-2 8M (wt) & wt & \checkmark & 0.214 & 0.231 & 0.371 & 0.392 & 0.298 & 0.114 & 0.059 & 0.029 \\
80 & ESM-2 35M (wt) & wt & \checkmark & 0.213 & 0.245 & 0.359 & 0.390 & 0.292 & 0.109 & 0.069 & 0.026 \\
81 & ProGen2-M & tf & \checkmark & 0.212 & 0.368 & 0.295 & 0.349 & 0.341 & 0.079 & 0.044 & 0.006 \\
82 & ProGen3-3B & tf & \checkmark & 0.211 & 0.366 & 0.270 & 0.367 & 0.359 & 0.078 & 0.034 & 0.004 \\
83 & CARP-600K & wt & \checkmark & 0.210 & 0.221 & 0.354 & 0.395 & 0.293 & 0.116 & 0.069 & 0.025 \\
84 & ProtSSN (k=30, h=1280) & wt & $\times$ & 0.210 & 0.336 & 0.392 & 0.305 & 0.345 & 0.025 & 0.032 & 0.036 \\
85 & SaProt (35M\_AF2, wt) & wt & \checkmark & 0.209 & 0.210 & 0.421 & 0.391 & 0.275 & 0.059 & 0.072 & 0.036 \\
86 & ESM-2 150M (mask) & mask & \checkmark & 0.209 & 0.320 & 0.380 & 0.303 & 0.304 & 0.069 & 0.054 & 0.035 \\
87 & ProGen3-1B & tf & \checkmark & 0.207 & 0.336 & 0.280 & 0.335 & 0.344 & 0.094 & 0.056 & 0.003 \\
88 & ProSST (K=1024) & wt & $\times$ & 0.204 & 0.210 & 0.351 & 0.436 & 0.296 & 0.063 & 0.049 & 0.025 \\
89 & ESM-1v (mask) & mask & \checkmark & 0.202 & 0.326 & 0.271 & 0.289 & 0.356 & 0.069 & 0.074 & 0.032 \\
90 & ProtSSN (k=10, h=512) & wt & $\times$ & 0.202 & 0.320 & 0.376 & 0.277 & 0.338 & 0.028 & 0.043 & 0.033 \\
91 & ProSST (K=4096) & wt & $\times$ & 0.200 & 0.186 & 0.318 & 0.460 & 0.303 & 0.073 & 0.029 & 0.032 \\
92 & ProGen3-762M & tf & \checkmark & 0.192 & 0.316 & 0.249 & 0.317 & 0.341 & 0.063 & 0.038 & 0.023 \\
93 & ProSST (K=512) & wt & $\times$ & 0.191 & 0.194 & 0.314 & 0.441 & 0.281 & 0.055 & 0.021 & 0.032 \\
94 & ProGen3-339M & tf & \checkmark & 0.191 & 0.291 & 0.248 & 0.315 & 0.337 & 0.065 & 0.067 & 0.013 \\
95 & SaProt (650M AF2, wt) & wt & $\times$ & 0.188 & 0.213 & 0.423 & 0.349 & 0.231 & 0.043 & 0.033 & 0.024 \\
96 & ESM-2 35M (mask) & mask & \checkmark & 0.184 & 0.274 & 0.280 & 0.285 & 0.288 & 0.070 & 0.060 & 0.032 \\
97 & ProGen3-112M & tf & \checkmark & 0.182 & 0.247 & 0.243 & 0.303 & 0.324 & 0.065 & 0.064 & 0.027 \\
98 & ProGen3-219M & tf & \checkmark & 0.181 & 0.275 & 0.242 & 0.282 & 0.335 & 0.063 & 0.048 & 0.021 \\
99 & ProSST (K=128) & wt & $\times$ & 0.179 & 0.196 & 0.266 & 0.393 & 0.272 & 0.059 & 0.042 & 0.025 \\
100 & ProGen2-B & tf & \checkmark & 0.175 & 0.371 & 0.055 & 0.333 & 0.343 & 0.071 & 0.048 & 0.006 \\
101 & ESM-2 8M (mask) & mask & \checkmark & 0.173 & 0.258 & 0.240 & 0.275 & 0.280 & 0.074 & 0.054 & 0.033 \\
102 & ProGen2-L & tf & \checkmark & 0.173 & 0.358 & 0.090 & 0.300 & 0.340 & 0.076 & 0.042 & 0.007 \\
103 & RITA-XL & tf & $\times$ & 0.172 & 0.327 & 0.236 & 0.272 & 0.270 & 0.043 & 0.065 & -0.008 \\
104 & ProGen2-S & tf & \checkmark & 0.171 & 0.300 & 0.149 & 0.301 & 0.320 & 0.077 & 0.050 & 0.001 \\
105 & RITA-L & tf & $\times$ & 0.166 & 0.323 & 0.253 & 0.241 & 0.249 & 0.052 & 0.060 & -0.014 \\
106 & ProGen2-XL & tf & $\times$ & 0.162 & 0.308 & 0.240 & 0.265 & 0.259 & 0.052 & 0.037 & -0.029 \\
107 & ProSST (K=20) & wt & $\times$ & 0.161 & 0.179 & 0.276 & 0.373 & 0.232 & 0.052 & -0.006 & 0.023 \\
108 & RITA-M & tf & $\times$ & 0.155 & 0.335 & 0.229 & 0.227 & 0.230 & 0.020 & 0.068 & -0.026 \\
109 & SaProt (35M\_AF2, mask) & mask & $\times$ & 0.149 & 0.189 & 0.303 & 0.246 & 0.205 & 0.030 & 0.042 & 0.025 \\
110 & RITA-S & tf & $\times$ & 0.148 & 0.319 & 0.234 & 0.206 & 0.205 & 0.025 & 0.072 & -0.021 \\
111 & ESM-2 3B (mask) & mask & $\times$ & 0.145 & 0.330 & 0.352 & 0.032 & 0.290 & -0.054 & 0.048 & 0.018 \\
112 & ESM-2 3B (wt) & wt & $\times$ & 0.138 & 0.321 & 0.342 & 0.019 & 0.277 & -0.055 & 0.046 & 0.017 \\
113 & ESM3 & mask & $\times$ & 0.137 & 0.203 & 0.338 & 0.168 & 0.186 & 0.013 & 0.037 & 0.015 \\
114 & ProGen2-M & tf & $\times$ & 0.117 & 0.307 & 0.121 & 0.191 & 0.161 & 0.012 & 0.038 & -0.011 \\
115 & ESM-2 650M (wt) & wt & $\times$ & 0.114 & 0.321 & 0.232 & -0.004 & 0.210 & -0.040 & 0.050 & 0.030 \\
116 & ESM-2 650M (mask) & mask & $\times$ & 0.111 & 0.326 & 0.199 & -0.001 & 0.218 & -0.040 & 0.042 & 0.031 \\
117 & ESMC-600M & mask & $\times$ & 0.108 & 0.190 & 0.263 & 0.077 & 0.180 & -0.015 & 0.036 & 0.024 \\
118 & SaProt (35M\_AF2, wt) & wt & $\times$ & 0.108 & 0.122 & 0.272 & 0.193 & 0.117 & -0.025 & 0.048 & 0.026 \\
119 & ProGen3-3B & tf & $\times$ & 0.101 & 0.274 & 0.011 & 0.191 & 0.222 & -0.006 & 0.030 & -0.013 \\
120 & ESMC-300M & mask & $\times$ & 0.097 & 0.181 & 0.220 & 0.074 & 0.165 & -0.020 & 0.033 & 0.023 \\
121 & ESM-1b (wt) & wt & $\times$ & 0.094 & 0.248 & 0.190 & 0.025 & 0.125 & -0.025 & 0.066 & 0.027 \\
122 & ESM-1b (mask) & mask & $\times$ & 0.093 & 0.253 & 0.190 & 0.020 & 0.129 & -0.031 & 0.063 & 0.029 \\
123 & ProGen3-1B & tf & $\times$ & 0.092 & 0.225 & 0.030 & 0.139 & 0.187 & 0.017 & 0.056 & -0.009 \\
124 & CARP-640M & wt & $\times$ & 0.092 & 0.291 & 0.066 & -0.003 & 0.174 & -0.026 & \textcolor{red!75!black}{\textbf{0.116}} & 0.024 \\
125 & ProGen2-B & tf & $\times$ & 0.076 & 0.311 & -0.143 & 0.169 & 0.162 & -0.003 & 0.049 & -0.013 \\
126 & ESM-2 150M (mask) & mask & $\times$ & 0.071 & 0.193 & 0.214 & -0.007 & 0.054 & -0.024 & 0.050 & 0.020 \\
127 & ESM-1v (wt) & wt & $\times$ & 0.069 & 0.192 & 0.073 & -0.014 & 0.166 & -0.017 & 0.069 & 0.016 \\
128 & ESM-2 150M (wt) & wt & $\times$ & 0.069 & 0.188 & 0.207 & -0.006 & 0.047 & -0.023 & 0.049 & 0.019 \\
129 & ProGen2-L & tf & $\times$ & 0.064 & 0.293 & -0.156 & 0.122 & 0.156 & 0.008 & 0.032 & -0.009 \\
130 & ESM-1v (mask) & mask & $\times$ & 0.064 & 0.193 & 0.031 & -0.023 & 0.173 & -0.020 & 0.073 & 0.017 \\
131 & ProGen3-762M & tf & $\times$ & 0.062 & 0.181 & 0.010 & 0.068 & 0.155 & -0.013 & 0.026 & 0.008 \\
132 & ProGen3-339M & tf & $\times$ & 0.058 & 0.146 & 0.009 & 0.067 & 0.138 & -0.012 & 0.063 & -0.003 \\
133 & CARP-76M & wt & $\times$ & 0.050 & 0.129 & 0.106 & 0.014 & 0.057 & -0.016 & 0.049 & 0.013 \\
134 & ProGen2-S & tf & $\times$ & 0.045 & 0.132 & -0.028 & 0.080 & 0.089 & 0.002 & 0.045 & -0.008 \\
135 & ProGen3-112M & tf & $\times$ & 0.034 & 0.060 & 0.001 & 0.029 & 0.112 & -0.011 & 0.043 & 0.006 \\
136 & ProGen3-219M & tf & $\times$ & 0.033 & 0.102 & -0.018 & -0.005 & 0.122 & -0.018 & 0.040 & 0.005 \\
137 & CARP-38M & wt & $\times$ & 0.027 & 0.088 & 0.028 & -0.013 & 0.047 & -0.019 & 0.043 & 0.013 \\
138 & ESM-2 35M (wt) & wt & $\times$ & 0.026 & 0.098 & 0.052 & -0.018 & 0.015 & -0.024 & 0.048 & 0.012 \\
139 & ESM-2 35M (mask) & mask & $\times$ & 0.026 & 0.106 & 0.038 & -0.025 & 0.024 & -0.022 & 0.046 & 0.014 \\
140 & ESM-2 8M (wt) & wt & $\times$ & 0.017 & 0.065 & 0.021 & -0.026 & 0.021 & -0.017 & 0.041 & 0.012 \\
141 & ESM-2 8M (mask) & mask & $\times$ & 0.015 & 0.066 & 0.002 & -0.029 & 0.020 & -0.016 & 0.048 & 0.012 \\
142 & CARP-600K & wt & $\times$ & 0.006 & 0.045 & -0.032 & -0.015 & 0.010 & -0.014 & 0.043 & 0.006 \\
\end{longtable}
}

\subsection{Per-configuration scores at each stage}
\label{app:staged_tables}
These tables report component results for all 71 configurations on each benchmark, grouped into structure-aware, masked sequence, inverse-folding, and autoregressive models. They support the comparisons in Appendix~\ref{app:stage_ablations}. Starting from the original PFM scores, we apply adaptive MSA fusion, then compare background correction with and without the gate. RSA weighting is added to the gated correction, followed by pLDDT weighting. The final column gives the complete harness result.

\paragraph{ProteinGym: structure-based models.} Tables~\ref{tab:staged_per_config_inverse} and~\ref{tab:staged_per_config} trace the inverse-folding and structure-aware configurations, respectively. The stages test whether family evidence and background correction still contribute when the PFM already uses structure.

\begin{table}[H]
\centering
\caption{Scores at each stage for 10 inverse-folding configurations on ProteinGym (official 217-assay Average Spearman).}
\label{tab:staged_per_config_inverse}
\small
\setlength{\tabcolsep}{0pt}
\renewcommand{\arraystretch}{1.08}
\begin{tabular}{@{}p{0.40\textwidth}*{6}{>{\centering\arraybackslash}p{0.10\textwidth}}@{}}
\toprule
Model & Raw & \shortstack{MSA\\fusion} & \shortstack{+ Ungated\\correction} & \shortstack{+ Gated\\correction} & + RSA & \shortstack{+ pLDDT\\(Full)} \\
\midrule
ESM-IF1 & 0.421 & 0.466 & 0.457 & 0.467 & 0.473 & 0.474 \\
MIF-ST & 0.430 & 0.440 & 0.441 & 0.444 & 0.456 & 0.458 \\
ProteinMPNN (v\_48\_002) & 0.378 & 0.425 & 0.417 & 0.431 & 0.441 & 0.443 \\
ProteinMPNN (v\_48\_010) & 0.385 & 0.431 & 0.424 & 0.439 & 0.449 & 0.451 \\
ProteinMPNN (v\_48\_020) & 0.389 & 0.434 & 0.430 & 0.444 & 0.454 & 0.456 \\
ProteinMPNN (v\_48\_030) & 0.381 & 0.428 & 0.431 & 0.440 & 0.451 & 0.453 \\
ProteinMPNN-Soluble (v\_48\_002) & 0.357 & 0.409 & 0.408 & 0.418 & 0.428 & 0.430 \\
ProteinMPNN-Soluble (v\_48\_010) & 0.363 & 0.412 & 0.415 & 0.423 & 0.433 & 0.436 \\
ProteinMPNN-Soluble (v\_48\_020) & 0.361 & 0.413 & 0.414 & 0.422 & 0.433 & 0.435 \\
ProteinMPNN-Soluble (v\_48\_030) & 0.353 & 0.408 & 0.413 & 0.419 & 0.430 & 0.433 \\
\bottomrule
\end{tabular}
\end{table}

\begin{table}[H]
\centering
\caption{Scores at each stage for 26 structure-aware configurations on ProteinGym (official 217-assay Average Spearman).}
\label{tab:staged_per_config}
\small
\setlength{\tabcolsep}{0pt}
\renewcommand{\arraystretch}{1.08}
\begin{tabular}{@{}p{0.40\textwidth}*{6}{>{\centering\arraybackslash}p{0.10\textwidth}}@{}}
\toprule
Model & Raw & \shortstack{MSA\\fusion} & \shortstack{+ Ungated\\correction} & \shortstack{+ Gated\\correction} & + RSA & \shortstack{+ pLDDT\\(Full)} \\
\midrule
ESM3 & 0.442 & 0.461 & 0.460 & 0.469 & 0.481 & 0.483 \\
ProSST (K=1024) & 0.485 & 0.517 & 0.513 & 0.522 & 0.530 & 0.532 \\
ProSST (K=128) & 0.469 & 0.506 & 0.503 & 0.512 & 0.522 & 0.525 \\
ProSST (K=20) & 0.438 & 0.486 & 0.481 & 0.495 & 0.507 & 0.510 \\
ProSST (K=2048) & 0.507 & 0.525 & 0.522 & 0.528 & 0.533 & 0.534 \\
ProSST (K=4096) & 0.498 & 0.525 & 0.522 & 0.531 & 0.538 & 0.541 \\
ProSST (K=512) & 0.471 & 0.507 & 0.503 & 0.513 & 0.522 & 0.524 \\
ProtSSN & 0.451 & 0.453 & 0.433 & 0.454 & 0.474 & 0.476 \\
ProtSSN (k=10, h=1280) & 0.435 & 0.438 & 0.420 & 0.438 & 0.462 & 0.465 \\
ProtSSN (k=10, h=512) & 0.432 & 0.435 & 0.418 & 0.435 & 0.458 & 0.462 \\
ProtSSN (k=10, h=768) & 0.424 & 0.426 & 0.408 & 0.427 & 0.454 & 0.457 \\
ProtSSN (k=20, h=1280) & 0.444 & 0.446 & 0.427 & 0.446 & 0.467 & 0.470 \\
ProtSSN (k=20, h=512) & 0.443 & 0.445 & 0.428 & 0.445 & 0.467 & 0.470 \\
ProtSSN (k=20, h=768) & 0.442 & 0.444 & 0.424 & 0.445 & 0.466 & 0.470 \\
ProtSSN (k=30, h=1280) & 0.439 & 0.441 & 0.422 & 0.442 & 0.463 & 0.466 \\
ProtSSN (k=30, h=512) & 0.441 & 0.443 & 0.426 & 0.444 & 0.466 & 0.469 \\
ProtSSN (k=30, h=768) & 0.441 & 0.443 & 0.428 & 0.444 & 0.464 & 0.467 \\
S3F (mask) & 0.466 & 0.475 & 0.477 & 0.479 & 0.486 & 0.488 \\
S3F (wt) & 0.460 & 0.471 & 0.466 & 0.474 & 0.485 & 0.489 \\
SaProt (35M\_AF2, mask) & 0.408 & 0.444 & 0.448 & 0.452 & 0.463 & 0.465 \\
SaProt (35M\_AF2, wt) & 0.366 & 0.373 & 0.376 & 0.381 & 0.419 & 0.425 \\
SaProt (650M\_PDB, mask) & 0.463 & 0.478 & 0.476 & 0.480 & 0.487 & 0.486 \\
SaProt (650M\_PDB, wt) & 0.428 & 0.432 & 0.430 & 0.436 & 0.451 & 0.450 \\
SaProt (650M AF2, mask) & 0.457 & 0.474 & 0.474 & 0.478 & 0.486 & 0.487 \\
SaProt (650M AF2, wt) & 0.424 & 0.427 & 0.427 & 0.433 & 0.452 & 0.454 \\
VenusREM2 & 0.524 & 0.542 & 0.538 & 0.550 & 0.554 & 0.556 \\
\bottomrule
\end{tabular}
\end{table}

\paragraph{ProteinGym: sequence models.} Tables~\ref{tab:staged_per_config_autoregressive} and~\ref{tab:staged_per_config_masked} trace the autoregressive and masked sequence configurations, respectively. The full harness improves on the original model for every configuration. Intermediate scores show the contributions of MSA fusion, background correction, and structural weighting.

\begin{table}[H]
\centering
\caption{Scores at each stage for 15 autoregressive language model configurations on ProteinGym (official 217-assay Average Spearman).}
\label{tab:staged_per_config_autoregressive}
\small
\setlength{\tabcolsep}{0pt}
\renewcommand{\arraystretch}{1.08}
\begin{tabular}{@{}p{0.40\textwidth}*{6}{>{\centering\arraybackslash}p{0.10\textwidth}}@{}}
\toprule
Model & Raw & \shortstack{MSA\\fusion} & \shortstack{+ Ungated\\correction} & \shortstack{+ Gated\\correction} & + RSA & \shortstack{+ pLDDT\\(Full)} \\
\midrule
ProGen2-B & 0.330 & 0.377 & 0.387 & 0.396 & 0.419 & 0.422 \\
ProGen2-L & 0.327 & 0.371 & 0.382 & 0.390 & 0.413 & 0.416 \\
ProGen2-M & 0.337 & 0.380 & 0.391 & 0.398 & 0.420 & 0.423 \\
ProGen2-S & 0.288 & 0.362 & 0.378 & 0.386 & 0.404 & 0.406 \\
ProGen2-XL & 0.351 & 0.373 & 0.375 & 0.385 & 0.415 & 0.419 \\
ProGen3-112M & 0.232 & 0.341 & 0.367 & 0.381 & 0.398 & 0.401 \\
ProGen3-1B & 0.320 & 0.368 & 0.373 & 0.380 & 0.403 & 0.406 \\
ProGen3-219M & 0.264 & 0.354 & 0.378 & 0.389 & 0.406 & 0.409 \\
ProGen3-339M & 0.286 & 0.364 & 0.385 & 0.395 & 0.412 & 0.414 \\
ProGen3-3B & 0.331 & 0.370 & 0.375 & 0.385 & 0.411 & 0.413 \\
ProGen3-762M & 0.304 & 0.369 & 0.382 & 0.392 & 0.411 & 0.413 \\
RITA-L & 0.309 & 0.358 & 0.368 & 0.377 & 0.404 & 0.407 \\
RITA-M & 0.297 & 0.358 & 0.370 & 0.379 & 0.402 & 0.405 \\
RITA-S & 0.254 & 0.344 & 0.366 & 0.376 & 0.395 & 0.398 \\
RITA-XL & 0.305 & 0.352 & 0.358 & 0.367 & 0.398 & 0.401 \\
\bottomrule
\end{tabular}
\end{table}

\begin{table}[H]
\centering
\caption{Scores at each stage for 20 masked sequence language model configurations on ProteinGym (official 217-assay Average Spearman).}
\label{tab:staged_per_config_masked}
\small
\setlength{\tabcolsep}{0pt}
\renewcommand{\arraystretch}{1.08}
\begin{tabular}{@{}p{0.40\textwidth}*{6}{>{\centering\arraybackslash}p{0.10\textwidth}}@{}}
\toprule
Model & Raw & \shortstack{MSA\\fusion} & \shortstack{+ Ungated\\correction} & \shortstack{+ Gated\\correction} & + RSA & \shortstack{+ pLDDT\\(Full)} \\
\midrule
CARP-38M & 0.314 & 0.342 & 0.356 & 0.367 & 0.409 & 0.415 \\
CARP-600K & 0.159 & 0.221 & 0.249 & 0.271 & 0.345 & 0.354 \\
CARP-640M & 0.390 & 0.427 & 0.433 & 0.437 & 0.450 & 0.451 \\
CARP-76M & 0.360 & 0.380 & 0.391 & 0.399 & 0.432 & 0.436 \\
ESM-1b (mask) & 0.389 & 0.430 & 0.436 & 0.440 & 0.453 & 0.455 \\
ESM-1b (wt) & 0.394 & 0.407 & 0.414 & 0.418 & 0.445 & 0.447 \\
ESM-1v (mask) & 0.407 & 0.436 & 0.444 & 0.450 & 0.461 & 0.462 \\
ESM-1v (wt) & 0.410 & 0.419 & 0.428 & 0.433 & 0.455 & 0.457 \\
ESM-2 150M (mask) & 0.388 & 0.424 & 0.434 & 0.438 & 0.450 & 0.452 \\
ESM-2 150M (wt) & 0.393 & 0.406 & 0.416 & 0.421 & 0.451 & 0.455 \\
ESM-2 35M (mask) & 0.321 & 0.395 & 0.407 & 0.414 & 0.427 & 0.430 \\
ESM-2 35M (wt) & 0.329 & 0.355 & 0.367 & 0.377 & 0.419 & 0.426 \\
ESM-2 3B (mask) & 0.407 & 0.436 & 0.444 & 0.445 & 0.459 & 0.462 \\
ESM-2 3B (wt) & 0.406 & 0.417 & 0.425 & 0.426 & 0.451 & 0.455 \\
ESM-2 650M (mask) & 0.415 & 0.445 & 0.454 & 0.457 & 0.468 & 0.470 \\
ESM-2 650M (wt) & 0.418 & 0.429 & 0.438 & 0.440 & 0.465 & 0.468 \\
ESM-2 8M (mask) & 0.223 & 0.345 & 0.367 & 0.378 & 0.392 & 0.395 \\
ESM-2 8M (wt) & 0.251 & 0.291 & 0.307 & 0.323 & 0.380 & 0.387 \\
ESMC-300M & 0.409 & 0.438 & 0.440 & 0.446 & 0.458 & 0.460 \\
ESMC-600M & 0.407 & 0.438 & 0.445 & 0.449 & 0.462 & 0.464 \\
\bottomrule
\end{tabular}
\end{table}

\paragraph{VenusMutHub: structure-based models.} Tables~\ref{tab:staged_per_config_vmh_inverse} and~\ref{tab:staged_per_config_vmh} trace the inverse-folding and structure-aware configurations, respectively. The stages test whether family evidence and background correction still contribute when the PFM already uses structure.

\begin{table}[H]
\centering
\caption{Scores at each stage for 10 inverse-folding configurations on VenusMutHub (mean Spearman over 905 assays).}
\label{tab:staged_per_config_vmh_inverse}
\small
\setlength{\tabcolsep}{0pt}
\renewcommand{\arraystretch}{1.08}
\begin{tabular}{@{}p{0.40\textwidth}*{6}{>{\centering\arraybackslash}p{0.10\textwidth}}@{}}
\toprule
Model & Raw & \shortstack{MSA\\fusion} & \shortstack{+ Ungated\\correction} & \shortstack{+ Gated\\correction} & + RSA & \shortstack{+ pLDDT\\(Full)} \\
\midrule
ESM-IF1 & 0.216 & 0.260 & 0.243 & 0.260 & 0.268 & 0.267 \\
MIF-ST & 0.196 & 0.209 & 0.203 & 0.210 & 0.228 & 0.230 \\
ProteinMPNN (v\_48\_002) & 0.191 & 0.233 & 0.232 & 0.241 & 0.245 & 0.244 \\
ProteinMPNN (v\_48\_010) & 0.218 & 0.240 & 0.246 & 0.247 & 0.254 & 0.256 \\
ProteinMPNN (v\_48\_020) & 0.221 & 0.257 & 0.254 & 0.262 & 0.269 & 0.271 \\
ProteinMPNN (v\_48\_030) & 0.210 & 0.244 & 0.251 & 0.251 & 0.260 & 0.259 \\
ProteinMPNN-Soluble (v\_48\_002) & 0.198 & 0.236 & 0.231 & 0.241 & 0.249 & 0.250 \\
ProteinMPNN-Soluble (v\_48\_010) & 0.212 & 0.246 & 0.241 & 0.248 & 0.256 & 0.256 \\
ProteinMPNN-Soluble (v\_48\_020) & 0.216 & 0.249 & 0.253 & 0.257 & 0.266 & 0.265 \\
ProteinMPNN-Soluble (v\_48\_030) & 0.222 & 0.250 & 0.253 & 0.255 & 0.264 & 0.265 \\
\bottomrule
\end{tabular}
\end{table}

\begin{table}[H]
\centering
\caption{Scores at each stage for 26 structure-aware configurations on VenusMutHub (mean Spearman over 905 assays).}
\label{tab:staged_per_config_vmh}
\small
\setlength{\tabcolsep}{0pt}
\renewcommand{\arraystretch}{1.08}
\begin{tabular}{@{}p{0.40\textwidth}*{6}{>{\centering\arraybackslash}p{0.10\textwidth}}@{}}
\toprule
Model & Raw & \shortstack{MSA\\fusion} & \shortstack{+ Ungated\\correction} & \shortstack{+ Gated\\correction} & + RSA & \shortstack{+ pLDDT\\(Full)} \\
\midrule
ESM3 & 0.185 & 0.210 & 0.214 & 0.213 & 0.226 & 0.226 \\
ProSST (K=1024) & 0.162 & 0.223 & 0.236 & 0.231 & 0.241 & 0.242 \\
ProSST (K=128) & 0.139 & 0.213 & 0.225 & 0.221 & 0.231 & 0.230 \\
ProSST (K=20) & 0.143 & 0.213 & 0.216 & 0.224 & 0.237 & 0.238 \\
ProSST (K=2048) & 0.213 & 0.234 & 0.233 & 0.237 & 0.250 & 0.249 \\
ProSST (K=4096) & 0.161 & 0.225 & 0.228 & 0.230 & 0.238 & 0.237 \\
ProSST (K=512) & 0.124 & 0.210 & 0.220 & 0.217 & 0.226 & 0.226 \\
ProtSSN & 0.174 & 0.175 & 0.158 & 0.174 & 0.197 & 0.195 \\
ProtSSN (k=10, h=1280) & 0.166 & 0.168 & 0.155 & 0.167 & 0.191 & 0.192 \\
ProtSSN (k=10, h=512) & 0.169 & 0.171 & 0.157 & 0.171 & 0.196 & 0.197 \\
ProtSSN (k=10, h=768) & 0.152 & 0.156 & 0.143 & 0.155 & 0.178 & 0.178 \\
ProtSSN (k=20, h=1280) & 0.174 & 0.176 & 0.162 & 0.175 & 0.198 & 0.198 \\
ProtSSN (k=20, h=512) & 0.168 & 0.171 & 0.158 & 0.171 & 0.192 & 0.192 \\
ProtSSN (k=20, h=768) & 0.165 & 0.169 & 0.155 & 0.168 & 0.189 & 0.191 \\
ProtSSN (k=30, h=1280) & 0.162 & 0.164 & 0.144 & 0.162 & 0.184 & 0.183 \\
ProtSSN (k=30, h=512) & 0.169 & 0.172 & 0.158 & 0.172 & 0.193 & 0.191 \\
ProtSSN (k=30, h=768) & 0.160 & 0.165 & 0.158 & 0.164 & 0.185 & 0.185 \\
S3F (mask) & 0.211 & 0.221 & 0.228 & 0.224 & 0.233 & 0.234 \\
S3F (wt) & 0.193 & 0.208 & 0.212 & 0.209 & 0.225 & 0.224 \\
SaProt (35M\_AF2, mask) & 0.188 & 0.224 & 0.228 & 0.228 & 0.236 & 0.237 \\
SaProt (35M\_AF2, wt) & 0.142 & 0.148 & 0.153 & 0.156 & 0.181 & 0.179 \\
SaProt (650M\_PDB, mask) & 0.205 & 0.219 & 0.220 & 0.220 & 0.237 & 0.240 \\
SaProt (650M\_PDB, wt) & 0.182 & 0.185 & 0.187 & 0.188 & 0.217 & 0.215 \\
SaProt (650M AF2, mask) & 0.207 & 0.225 & 0.230 & 0.225 & 0.239 & 0.240 \\
SaProt (650M AF2, wt) & 0.178 & 0.180 & 0.189 & 0.183 & 0.203 & 0.202 \\
VenusREM2 & 0.190 & 0.243 & 0.255 & 0.248 & 0.257 & 0.258 \\
\bottomrule
\end{tabular}
\end{table}

\paragraph{VenusMutHub: sequence models.} Tables~\ref{tab:staged_per_config_vmh_autoregressive} and~\ref{tab:staged_per_config_vmh_masked} trace the autoregressive and masked sequence configurations, respectively. The full harness improves on the original model for every configuration. Intermediate scores show the contributions of MSA fusion, background correction, and structural weighting.

\begin{table}[H]
\centering
\caption{Scores at each stage for 15 autoregressive language model configurations on VenusMutHub (mean Spearman over 905 assays).}
\label{tab:staged_per_config_vmh_autoregressive}
\small
\setlength{\tabcolsep}{0pt}
\renewcommand{\arraystretch}{1.08}
\begin{tabular}{@{}p{0.40\textwidth}*{6}{>{\centering\arraybackslash}p{0.10\textwidth}}@{}}
\toprule
Model & Raw & \shortstack{MSA\\fusion} & \shortstack{+ Ungated\\correction} & \shortstack{+ Gated\\correction} & + RSA & \shortstack{+ pLDDT\\(Full)} \\
\midrule
ProGen2-B & 0.124 & 0.168 & 0.183 & 0.180 & 0.193 & 0.194 \\
ProGen2-L & 0.120 & 0.159 & 0.175 & 0.175 & 0.189 & 0.188 \\
ProGen2-M & 0.115 & 0.154 & 0.173 & 0.167 & 0.182 & 0.182 \\
ProGen2-S & 0.091 & 0.149 & 0.179 & 0.179 & 0.191 & 0.192 \\
ProGen2-XL & 0.123 & 0.148 & 0.149 & 0.153 & 0.178 & 0.176 \\
ProGen3-112M & 0.072 & 0.141 & 0.160 & 0.162 & 0.176 & 0.177 \\
ProGen3-1B & 0.108 & 0.152 & 0.160 & 0.158 & 0.174 & 0.173 \\
ProGen3-219M & 0.092 & 0.150 & 0.184 & 0.181 & 0.191 & 0.190 \\
ProGen3-339M & 0.110 & 0.154 & 0.177 & 0.174 & 0.190 & 0.189 \\
ProGen3-3B & 0.110 & 0.149 & 0.147 & 0.153 & 0.173 & 0.170 \\
ProGen3-762M & 0.111 & 0.160 & 0.174 & 0.172 & 0.184 & 0.180 \\
RITA-L & 0.110 & 0.160 & 0.182 & 0.179 & 0.193 & 0.193 \\
RITA-M & 0.100 & 0.155 & 0.178 & 0.175 & 0.188 & 0.186 \\
RITA-S & 0.056 & 0.148 & 0.173 & 0.171 & 0.181 & 0.182 \\
RITA-XL & 0.095 & 0.132 & 0.142 & 0.139 & 0.157 & 0.160 \\
\bottomrule
\end{tabular}
\end{table}

\begin{table}[H]
\centering
\caption{Scores at each stage for 20 masked sequence language model configurations on VenusMutHub (mean Spearman over 905 assays).}
\label{tab:staged_per_config_vmh_masked}
\small
\setlength{\tabcolsep}{0pt}
\renewcommand{\arraystretch}{1.08}
\begin{tabular}{@{}p{0.40\textwidth}*{6}{>{\centering\arraybackslash}p{0.10\textwidth}}@{}}
\toprule
Model & Raw & \shortstack{MSA\\fusion} & \shortstack{+ Ungated\\correction} & \shortstack{+ Gated\\correction} & + RSA & \shortstack{+ pLDDT\\(Full)} \\
\midrule
CARP-38M & 0.073 & 0.102 & 0.136 & 0.132 & 0.159 & 0.158 \\
CARP-600K & 0.021 & 0.063 & 0.120 & 0.115 & 0.150 & 0.151 \\
CARP-640M & 0.157 & 0.187 & 0.191 & 0.194 & 0.206 & 0.206 \\
CARP-76M & 0.131 & 0.149 & 0.172 & 0.174 & 0.192 & 0.194 \\
ESM-1b (mask) & 0.166 & 0.198 & 0.200 & 0.200 & 0.214 & 0.211 \\
ESM-1b (wt) & 0.153 & 0.166 & 0.174 & 0.171 & 0.194 & 0.193 \\
ESM-1v (mask) & 0.162 & 0.188 & 0.193 & 0.193 & 0.206 & 0.206 \\
ESM-1v (wt) & 0.153 & 0.162 & 0.165 & 0.168 & 0.187 & 0.188 \\
ESM-2 150M (mask) & 0.167 & 0.203 & 0.221 & 0.216 & 0.226 & 0.225 \\
ESM-2 150M (wt) & 0.165 & 0.177 & 0.192 & 0.191 & 0.216 & 0.216 \\
ESM-2 35M (mask) & 0.122 & 0.177 & 0.207 & 0.199 & 0.212 & 0.212 \\
ESM-2 35M (wt) & 0.110 & 0.134 & 0.169 & 0.159 & 0.187 & 0.186 \\
ESM-2 3B (mask) & 0.182 & 0.205 & 0.209 & 0.208 & 0.221 & 0.223 \\
ESM-2 3B (wt) & 0.161 & 0.170 & 0.170 & 0.174 & 0.194 & 0.194 \\
ESM-2 650M (mask) & 0.169 & 0.197 & 0.199 & 0.200 & 0.214 & 0.213 \\
ESM-2 650M (wt) & 0.152 & 0.163 & 0.167 & 0.165 & 0.187 & 0.187 \\
ESM-2 8M (mask) & 0.079 & 0.158 & 0.189 & 0.187 & 0.197 & 0.198 \\
ESM-2 8M (wt) & 0.065 & 0.103 & 0.149 & 0.146 & 0.176 & 0.176 \\
ESMC-300M & 0.144 & 0.177 & 0.181 & 0.182 & 0.196 & 0.198 \\
ESMC-600M & 0.155 & 0.172 & 0.181 & 0.176 & 0.193 & 0.194 \\
\bottomrule
\end{tabular}
\end{table}

\paragraph{\ViroBenchmarkName{}: structure-based models.} Tables~\ref{tab:staged_per_config_viro_inverse} and~\ref{tab:staged_per_config_viro} trace the inverse-folding and structure-aware configurations, respectively. The stages test whether family evidence and background correction still contribute when the PFM already uses structure.

\begin{table}[H]
\centering
\caption{Scores at each stage for 10 inverse-folding configurations on \ViroBenchmarkName{}: mutant-only hierarchical Spearman over 89 assays, 52 groups, and 7 phenotypes; background correction uses coefficient $\beta=1-\alpha$ before structural weighting.}
\label{tab:staged_per_config_viro_inverse}
\small
\setlength{\tabcolsep}{0pt}
\renewcommand{\arraystretch}{1.08}
\begin{tabular}{@{}p{0.40\textwidth}*{6}{>{\centering\arraybackslash}p{0.10\textwidth}}@{}}
\toprule
Model & Raw & \shortstack{MSA\\fusion} & \shortstack{+ Ungated\\correction} & \shortstack{+ Gated\\correction} & + RSA & \shortstack{+ pLDDT\\(Full)} \\
\midrule
ESM-IF1 & 0.285 & 0.305 & 0.300 & 0.307 & 0.316 & 0.318 \\
MIF-ST & 0.288 & 0.299 & 0.304 & 0.303 & 0.316 & 0.317 \\
ProteinMPNN (v\_48\_002) & 0.269 & 0.285 & 0.283 & 0.291 & 0.304 & 0.305 \\
ProteinMPNN (v\_48\_010) & 0.271 & 0.285 & 0.285 & 0.292 & 0.305 & 0.306 \\
ProteinMPNN (v\_48\_020) & 0.270 & 0.284 & 0.285 & 0.291 & 0.304 & 0.306 \\
ProteinMPNN (v\_48\_030) & 0.266 & 0.280 & 0.284 & 0.288 & 0.302 & 0.304 \\
ProteinMPNN-Soluble (v\_48\_002) & 0.269 & 0.284 & 0.285 & 0.292 & 0.304 & 0.305 \\
ProteinMPNN-Soluble (v\_48\_010) & 0.268 & 0.281 & 0.285 & 0.289 & 0.302 & 0.304 \\
ProteinMPNN-Soluble (v\_48\_020) & 0.265 & 0.280 & 0.282 & 0.288 & 0.302 & 0.304 \\
ProteinMPNN-Soluble (v\_48\_030) & 0.256 & 0.272 & 0.277 & 0.281 & 0.297 & 0.299 \\
\bottomrule
\end{tabular}
\end{table}

\begin{table}[H]
\centering
\caption{Scores at each stage for 26 structure-aware configurations on \ViroBenchmarkName{}: mutant-only hierarchical Spearman over 89 assays, 52 groups, and 7 phenotypes; background correction uses coefficient $\beta=1-\alpha$ before structural weighting.}
\label{tab:staged_per_config_viro}
\small
\setlength{\tabcolsep}{0pt}
\renewcommand{\arraystretch}{1.08}
\begin{tabular}{@{}p{0.40\textwidth}*{6}{>{\centering\arraybackslash}p{0.10\textwidth}}@{}}
\toprule
Model & Raw & \shortstack{MSA\\fusion} & \shortstack{+ Ungated\\correction} & \shortstack{+ Gated\\correction} & + RSA & \shortstack{+ pLDDT\\(Full)} \\
\midrule
ESM3 & 0.137 & 0.183 & 0.196 & 0.211 & 0.244 & 0.249 \\
ProSST (K=1024) & 0.204 & 0.243 & 0.247 & 0.250 & 0.272 & 0.275 \\
ProSST (K=128) & 0.179 & 0.223 & 0.232 & 0.234 & 0.261 & 0.264 \\
ProSST (K=20) & 0.161 & 0.211 & 0.213 & 0.224 & 0.254 & 0.259 \\
ProSST (K=2048) & 0.271 & 0.290 & 0.284 & 0.294 & 0.309 & 0.310 \\
ProSST (K=4096) & 0.200 & 0.237 & 0.242 & 0.244 & 0.268 & 0.271 \\
ProSST (K=512) & 0.191 & 0.231 & 0.239 & 0.241 & 0.265 & 0.269 \\
ProtSSN & 0.233 & 0.236 & 0.208 & 0.237 & 0.288 & 0.294 \\
ProtSSN (k=10, h=1280) & 0.220 & 0.223 & 0.204 & 0.224 & 0.278 & 0.284 \\
ProtSSN (k=10, h=512) & 0.202 & 0.206 & 0.180 & 0.207 & 0.265 & 0.272 \\
ProtSSN (k=10, h=768) & 0.214 & 0.217 & 0.192 & 0.218 & 0.273 & 0.279 \\
ProtSSN (k=20, h=1280) & 0.230 & 0.233 & 0.208 & 0.234 & 0.283 & 0.289 \\
ProtSSN (k=20, h=512) & 0.233 & 0.237 & 0.211 & 0.237 & 0.285 & 0.291 \\
ProtSSN (k=20, h=768) & 0.226 & 0.229 & 0.205 & 0.230 & 0.281 & 0.287 \\
ProtSSN (k=30, h=1280) & 0.210 & 0.214 & 0.181 & 0.215 & 0.273 & 0.279 \\
ProtSSN (k=30, h=512) & 0.225 & 0.229 & 0.205 & 0.229 & 0.282 & 0.288 \\
ProtSSN (k=30, h=768) & 0.223 & 0.227 & 0.199 & 0.228 & 0.280 & 0.286 \\
S3F (mask) & 0.269 & 0.280 & 0.284 & 0.285 & 0.301 & 0.302 \\
S3F (wt) & 0.279 & 0.291 & 0.291 & 0.296 & 0.314 & 0.318 \\
SaProt (35M\_AF2, mask) & 0.149 & 0.207 & 0.216 & 0.217 & 0.242 & 0.245 \\
SaProt (35M\_AF2, wt) & 0.108 & 0.121 & 0.126 & 0.127 & 0.199 & 0.209 \\
SaProt (650M\_PDB, mask) & 0.306 & 0.307 & 0.309 & 0.309 & 0.320 & 0.320 \\
SaProt (650M\_PDB, wt) & 0.273 & 0.280 & 0.280 & 0.283 & 0.307 & 0.307 \\
SaProt (650M AF2, mask) & 0.222 & 0.253 & 0.259 & 0.259 & 0.279 & 0.281 \\
SaProt (650M AF2, wt) & 0.188 & 0.195 & 0.200 & 0.201 & 0.254 & 0.259 \\
VenusREM2 & 0.236 & 0.263 & 0.267 & 0.274 & 0.294 & 0.297 \\
\bottomrule
\end{tabular}
\end{table}

\paragraph{\ViroBenchmarkName{}: sequence models.} Tables~\ref{tab:staged_per_config_viro_autoregressive} and~\ref{tab:staged_per_config_viro_masked} trace the autoregressive and masked sequence configurations, respectively. The full harness improves on the original model for every configuration. Intermediate scores show the contributions of MSA fusion, background correction, and structural weighting.

\begin{table}[H]
\centering
\caption{Scores at each stage for 15 autoregressive language model configurations on \ViroBenchmarkName{}: mutant-only hierarchical Spearman over 89 assays, 52 groups, and 7 phenotypes; background correction uses coefficient $\beta=1-\alpha$ before structural weighting.}
\label{tab:staged_per_config_viro_autoregressive}
\small
\setlength{\tabcolsep}{0pt}
\renewcommand{\arraystretch}{1.08}
\begin{tabular}{@{}p{0.40\textwidth}*{6}{>{\centering\arraybackslash}p{0.10\textwidth}}@{}}
\toprule
Model & Raw & \shortstack{MSA\\fusion} & \shortstack{+ Ungated\\correction} & \shortstack{+ Gated\\correction} & + RSA & \shortstack{+ pLDDT\\(Full)} \\
\midrule
ProGen2-B & 0.076 & 0.123 & 0.132 & 0.136 & 0.172 & 0.175 \\
ProGen2-L & 0.064 & 0.121 & 0.136 & 0.137 & 0.169 & 0.173 \\
ProGen2-M & 0.117 & 0.152 & 0.159 & 0.165 & 0.208 & 0.212 \\
ProGen2-S & 0.045 & 0.114 & 0.128 & 0.133 & 0.167 & 0.171 \\
ProGen2-XL & 0.162 & 0.179 & 0.179 & 0.186 & 0.232 & 0.235 \\
ProGen3-112M & 0.034 & 0.117 & 0.152 & 0.159 & 0.178 & 0.182 \\
ProGen3-1B & 0.092 & 0.149 & 0.171 & 0.177 & 0.204 & 0.207 \\
ProGen3-219M & 0.033 & 0.115 & 0.152 & 0.158 & 0.177 & 0.181 \\
ProGen3-339M & 0.058 & 0.133 & 0.162 & 0.167 & 0.187 & 0.191 \\
ProGen3-3B & 0.101 & 0.154 & 0.174 & 0.180 & 0.208 & 0.211 \\
ProGen3-762M & 0.062 & 0.132 & 0.159 & 0.165 & 0.189 & 0.192 \\
RITA-L & 0.166 & 0.181 & 0.184 & 0.189 & 0.245 & 0.247 \\
RITA-M & 0.155 & 0.175 & 0.176 & 0.182 & 0.234 & 0.237 \\
RITA-S & 0.148 & 0.176 & 0.179 & 0.184 & 0.233 & 0.236 \\
RITA-XL & 0.172 & 0.184 & 0.187 & 0.191 & 0.244 & 0.247 \\
\bottomrule
\end{tabular}
\end{table}

\begin{table}[H]
\centering
\caption{Scores at each stage for 20 masked sequence language model configurations on \ViroBenchmarkName{}: mutant-only hierarchical Spearman over 89 assays, 52 groups, and 7 phenotypes; background correction uses coefficient $\beta=1-\alpha$ before structural weighting.}
\label{tab:staged_per_config_viro_masked}
\small
\setlength{\tabcolsep}{0pt}
\renewcommand{\arraystretch}{1.08}
\begin{tabular}{@{}p{0.40\textwidth}*{6}{>{\centering\arraybackslash}p{0.10\textwidth}}@{}}
\toprule
Model & Raw & \shortstack{MSA\\fusion} & \shortstack{+ Ungated\\correction} & \shortstack{+ Gated\\correction} & + RSA & \shortstack{+ pLDDT\\(Full)} \\
\midrule
CARP-38M & 0.027 & 0.067 & 0.096 & 0.112 & 0.213 & 0.222 \\
CARP-600K & 0.006 & 0.047 & 0.087 & 0.098 & 0.201 & 0.210 \\
CARP-640M & 0.092 & 0.162 & 0.185 & 0.190 & 0.212 & 0.215 \\
CARP-76M & 0.050 & 0.090 & 0.113 & 0.130 & 0.221 & 0.230 \\
ESM-1b (mask) & 0.093 & 0.169 & 0.192 & 0.197 & 0.220 & 0.223 \\
ESM-1b (wt) & 0.094 & 0.127 & 0.150 & 0.159 & 0.240 & 0.248 \\
ESM-1v (mask) & 0.064 & 0.140 & 0.173 & 0.180 & 0.200 & 0.202 \\
ESM-1v (wt) & 0.069 & 0.099 & 0.124 & 0.137 & 0.229 & 0.238 \\
ESM-2 150M (mask) & 0.071 & 0.153 & 0.175 & 0.182 & 0.205 & 0.209 \\
ESM-2 150M (wt) & 0.069 & 0.103 & 0.122 & 0.136 & 0.226 & 0.235 \\
ESM-2 35M (mask) & 0.026 & 0.119 & 0.152 & 0.158 & 0.180 & 0.184 \\
ESM-2 35M (wt) & 0.026 & 0.063 & 0.092 & 0.099 & 0.203 & 0.213 \\
ESM-2 3B (mask) & 0.145 & 0.199 & 0.216 & 0.218 & 0.240 & 0.243 \\
ESM-2 3B (wt) & 0.138 & 0.164 & 0.183 & 0.186 & 0.251 & 0.260 \\
ESM-2 650M (mask) & 0.111 & 0.177 & 0.198 & 0.201 & 0.224 & 0.227 \\
ESM-2 650M (wt) & 0.114 & 0.142 & 0.163 & 0.168 & 0.243 & 0.252 \\
ESM-2 8M (mask) & 0.015 & 0.108 & 0.140 & 0.149 & 0.170 & 0.173 \\
ESM-2 8M (wt) & 0.017 & 0.053 & 0.082 & 0.097 & 0.203 & 0.214 \\
ESMC-300M & 0.097 & 0.164 & 0.184 & 0.193 & 0.216 & 0.220 \\
ESMC-600M & 0.108 & 0.173 & 0.191 & 0.200 & 0.224 & 0.228 \\
\bottomrule
\end{tabular}
\end{table}

\FloatBarrier
\subsection{Per-configuration five-metric results}
\label{app:per_config_metrics}
Tables~\ref{tab:pg_metrics_family_1}--\ref{tab:viro_metrics_family_4} are the configuration-level source of Figure~\ref{fig:scatter}: all $355$ model--metric pairs improve on ProteinGym and \ViroBenchmarkName{}, and $353$ of $355$ improve or remain unchanged at the reported precision on VenusMutHub.

\begingroup
\renewcommand{\arraystretch}{1.00}
\paragraph{ProteinGym: structural models.} Tables~\ref{tab:pg_metrics_family_1} and~\ref{tab:pg_metrics_family_3} compare the five metrics for structure-aware and inverse-folding models, respectively. The paired columns distinguish changes in ranking, binary discrimination, and recovery of top variants under the ProteinGym definitions.

\begin{table}[H]
\centering
\caption{Five-metric results for ProteinGym structure-aware models before (Raw) and after (\textsc{\HarnessShortName{}}) applying the harness.}
\label{tab:pg_metrics_family_1}
\fontsize{9}{10.8}\selectfont
\setlength{\tabcolsep}{0pt}
\begin{tabular}{@{}p{0.40\textwidth}*{10}{>{\centering\arraybackslash}p{0.06\textwidth}}@{}}
\toprule
Model & \multicolumn{2}{c}{Spearman} & \multicolumn{2}{c}{NDCG} & \multicolumn{2}{c}{AUC} & \multicolumn{2}{c}{MCC} & \multicolumn{2}{c}{Top-recall} \\
\cmidrule(lr){2-3}\cmidrule(lr){4-5}\cmidrule(lr){6-7}\cmidrule(lr){8-9}\cmidrule(lr){10-11}
& Raw & \textsc{\HarnessShortName{}} & Raw & \textsc{\HarnessShortName{}} & Raw & \textsc{\HarnessShortName{}} & Raw & \textsc{\HarnessShortName{}} & Raw & \textsc{\HarnessShortName{}} \\
\midrule
ESM3 & 0.442 & 0.483 & 0.767 & 0.780 & 0.744 & 0.765 & 0.346 & 0.377 & 0.228 & 0.236 \\
ProSST (K=1024) & 0.485 & 0.532 & 0.761 & 0.790 & 0.764 & 0.788 & 0.372 & 0.410 & 0.230 & 0.251 \\
ProSST (K=128) & 0.469 & 0.525 & 0.754 & 0.789 & 0.757 & 0.786 & 0.363 & 0.407 & 0.227 & 0.253 \\
ProSST (K=20) & 0.438 & 0.510 & 0.745 & 0.786 & 0.739 & 0.777 & 0.336 & 0.393 & 0.210 & 0.242 \\
ProSST (K=2048) & 0.507 & 0.534 & 0.757 & 0.778 & 0.777 & 0.791 & 0.398 & 0.416 & 0.236 & 0.249 \\
ProSST (K=4096) & 0.498 & 0.541 & 0.774 & 0.798 & 0.773 & 0.794 & 0.385 & 0.419 & 0.232 & 0.255 \\
ProSST (K=512) & 0.471 & 0.524 & 0.759 & 0.791 & 0.757 & 0.785 & 0.360 & 0.404 & 0.222 & 0.248 \\
ProtSSN & 0.451 & 0.476 & 0.766 & 0.777 & 0.748 & 0.762 & 0.353 & 0.373 & 0.228 & 0.231 \\
ProtSSN (k=10, h=1280) & 0.435 & 0.465 & 0.762 & 0.776 & 0.740 & 0.756 & 0.343 & 0.365 & 0.226 & 0.231 \\
ProtSSN (k=10, h=512) & 0.432 & 0.462 & 0.759 & 0.772 & 0.738 & 0.754 & 0.341 & 0.362 & 0.221 & 0.224 \\
ProtSSN (k=10, h=768) & 0.424 & 0.457 & 0.760 & 0.773 & 0.732 & 0.750 & 0.332 & 0.357 & 0.226 & 0.228 \\
ProtSSN (k=20, h=1280) & 0.444 & 0.470 & 0.762 & 0.775 & 0.744 & 0.759 & 0.350 & 0.369 & 0.228 & 0.230 \\
ProtSSN (k=20, h=512) & 0.443 & 0.470 & 0.764 & 0.777 & 0.743 & 0.758 & 0.346 & 0.367 & 0.227 & 0.231 \\
ProtSSN (k=20, h=768) & 0.442 & 0.470 & 0.763 & 0.775 & 0.742 & 0.757 & 0.344 & 0.366 & 0.225 & 0.229 \\
ProtSSN (k=30, h=1280) & 0.439 & 0.466 & 0.759 & 0.773 & 0.742 & 0.756 & 0.344 & 0.364 & 0.223 & 0.224 \\
ProtSSN (k=30, h=512) & 0.441 & 0.469 & 0.765 & 0.776 & 0.741 & 0.757 & 0.341 & 0.365 & 0.223 & 0.228 \\
ProtSSN (k=30, h=768) & 0.441 & 0.467 & 0.766 & 0.776 & 0.742 & 0.757 & 0.345 & 0.364 & 0.226 & 0.228 \\
S3F (mask) & 0.466 & 0.488 & 0.770 & 0.785 & 0.755 & 0.766 & 0.365 & 0.383 & 0.237 & 0.242 \\
S3F (wt) & 0.460 & 0.489 & 0.773 & 0.784 & 0.752 & 0.767 & 0.361 & 0.384 & 0.233 & 0.240 \\
SaProt (35M\_AF2, mask) & 0.408 & 0.465 & 0.736 & 0.763 & 0.724 & 0.754 & 0.318 & 0.362 & 0.215 & 0.236 \\
SaProt (35M\_AF2, wt) & 0.366 & 0.425 & 0.726 & 0.753 & 0.702 & 0.733 & 0.286 & 0.328 & 0.200 & 0.218 \\
SaProt (650M\_PDB, mask) & 0.463 & 0.486 & 0.775 & 0.783 & 0.756 & 0.767 & 0.365 & 0.381 & 0.231 & 0.237 \\
SaProt (650M\_PDB, wt) & 0.428 & 0.450 & 0.766 & 0.771 & 0.736 & 0.749 & 0.336 & 0.355 & 0.221 & 0.223 \\
SaProt (650M AF2, mask) & 0.457 & 0.487 & 0.769 & 0.780 & 0.752 & 0.767 & 0.359 & 0.383 & 0.234 & 0.243 \\
SaProt (650M AF2, wt) & 0.424 & 0.454 & 0.757 & 0.770 & 0.735 & 0.750 & 0.335 & 0.355 & 0.223 & 0.226 \\
VenusREM2 & 0.524 & 0.556 & 0.791 & 0.808 & 0.786 & 0.802 & 0.405 & 0.430 & 0.256 & 0.266 \\
\bottomrule
\end{tabular}
\end{table}

\begin{table}[H]
\centering
\caption{Five-metric results for ProteinGym inverse-folding models before (Raw) and after (\textsc{\HarnessShortName{}}) applying the harness.}
\label{tab:pg_metrics_family_3}
\fontsize{9}{10.8}\selectfont
\setlength{\tabcolsep}{0pt}
\begin{tabular}{@{}p{0.40\textwidth}*{10}{>{\centering\arraybackslash}p{0.06\textwidth}}@{}}
\toprule
Model & \multicolumn{2}{c}{Spearman} & \multicolumn{2}{c}{NDCG} & \multicolumn{2}{c}{AUC} & \multicolumn{2}{c}{MCC} & \multicolumn{2}{c}{Top-recall} \\
\cmidrule(lr){2-3}\cmidrule(lr){4-5}\cmidrule(lr){6-7}\cmidrule(lr){8-9}\cmidrule(lr){10-11}
& Raw & \textsc{\HarnessShortName{}} & Raw & \textsc{\HarnessShortName{}} & Raw & \textsc{\HarnessShortName{}} & Raw & \textsc{\HarnessShortName{}} & Raw & \textsc{\HarnessShortName{}} \\
\midrule
ESM-IF1 & 0.421 & 0.474 & 0.747 & 0.777 & 0.729 & 0.759 & 0.328 & 0.372 & 0.223 & 0.239 \\
MIF-ST & 0.430 & 0.458 & 0.772 & 0.778 & 0.736 & 0.752 & 0.339 & 0.362 & 0.233 & 0.235 \\
ProteinMPNN (v\_48\_002) & 0.378 & 0.443 & 0.741 & 0.770 & 0.706 & 0.741 & 0.302 & 0.345 & 0.218 & 0.239 \\
ProteinMPNN (v\_48\_010) & 0.385 & 0.451 & 0.747 & 0.774 & 0.709 & 0.745 & 0.303 & 0.350 & 0.225 & 0.244 \\
ProteinMPNN (v\_48\_020) & 0.389 & 0.456 & 0.746 & 0.774 & 0.710 & 0.747 & 0.304 & 0.353 & 0.226 & 0.245 \\
ProteinMPNN (v\_48\_030) & 0.381 & 0.453 & 0.736 & 0.769 & 0.707 & 0.746 & 0.296 & 0.352 & 0.218 & 0.240 \\
ProteinMPNN-Soluble (v\_48\_002) & 0.357 & 0.430 & 0.731 & 0.765 & 0.695 & 0.734 & 0.282 & 0.333 & 0.213 & 0.232 \\
ProteinMPNN-Soluble (v\_48\_010) & 0.363 & 0.436 & 0.735 & 0.767 & 0.697 & 0.737 & 0.286 & 0.337 & 0.215 & 0.236 \\
ProteinMPNN-Soluble (v\_48\_020) & 0.361 & 0.435 & 0.731 & 0.764 & 0.696 & 0.737 & 0.283 & 0.335 & 0.215 & 0.233 \\
ProteinMPNN-Soluble (v\_48\_030) & 0.353 & 0.433 & 0.726 & 0.762 & 0.691 & 0.735 & 0.273 & 0.336 & 0.211 & 0.231 \\
\bottomrule
\end{tabular}
\end{table}

\paragraph{ProteinGym: sequence models.} Tables~\ref{tab:pg_metrics_family_2} and~\ref{tab:pg_metrics_family_4} compare the five metrics for masked and autoregressive models, respectively. The paired columns distinguish changes in ranking, binary discrimination, and recovery of top variants under the ProteinGym definitions.

\begin{table}[H]
\centering
\caption{Five-metric results for ProteinGym masked protein LMs before (Raw) and after (\textsc{\HarnessShortName{}}) applying the harness.}
\label{tab:pg_metrics_family_2}
\fontsize{9}{10.8}\selectfont
\setlength{\tabcolsep}{0pt}
\begin{tabular}{@{}p{0.40\textwidth}*{10}{>{\centering\arraybackslash}p{0.06\textwidth}}@{}}
\toprule
Model & \multicolumn{2}{c}{Spearman} & \multicolumn{2}{c}{NDCG} & \multicolumn{2}{c}{AUC} & \multicolumn{2}{c}{MCC} & \multicolumn{2}{c}{Top-recall} \\
\cmidrule(lr){2-3}\cmidrule(lr){4-5}\cmidrule(lr){6-7}\cmidrule(lr){8-9}\cmidrule(lr){10-11}
& Raw & \textsc{\HarnessShortName{}} & Raw & \textsc{\HarnessShortName{}} & Raw & \textsc{\HarnessShortName{}} & Raw & \textsc{\HarnessShortName{}} & Raw & \textsc{\HarnessShortName{}} \\
\midrule
CARP-38M & 0.314 & 0.415 & 0.709 & 0.755 & 0.675 & 0.729 & 0.248 & 0.322 & 0.186 & 0.212 \\
CARP-600K & 0.159 & 0.354 & 0.649 & 0.740 & 0.590 & 0.694 & 0.127 & 0.271 & 0.144 & 0.194 \\
CARP-640M & 0.390 & 0.451 & 0.747 & 0.767 & 0.716 & 0.748 & 0.307 & 0.355 & 0.208 & 0.223 \\
CARP-76M & 0.360 & 0.436 & 0.728 & 0.759 & 0.699 & 0.740 & 0.282 & 0.339 & 0.196 & 0.214 \\
ESM-1b (mask) & 0.389 & 0.455 & 0.742 & 0.765 & 0.715 & 0.750 & 0.306 & 0.360 & 0.202 & 0.218 \\
ESM-1b (wt) & 0.394 & 0.447 & 0.747 & 0.768 & 0.719 & 0.747 & 0.311 & 0.353 & 0.202 & 0.214 \\
ESM-1v (mask) & 0.407 & 0.462 & 0.750 & 0.763 & 0.724 & 0.753 & 0.320 & 0.361 & 0.211 & 0.221 \\
ESM-1v (wt) & 0.410 & 0.457 & 0.752 & 0.771 & 0.727 & 0.752 & 0.324 & 0.357 & 0.210 & 0.219 \\
ESM-2 150M (mask) & 0.388 & 0.452 & 0.729 & 0.754 & 0.715 & 0.748 & 0.304 & 0.355 & 0.205 & 0.222 \\
ESM-2 150M (wt) & 0.393 & 0.455 & 0.736 & 0.763 & 0.718 & 0.751 & 0.310 & 0.355 & 0.210 & 0.225 \\
ESM-2 35M (mask) & 0.321 & 0.430 & 0.705 & 0.746 & 0.676 & 0.735 & 0.250 & 0.335 & 0.186 & 0.215 \\
ESM-2 35M (wt) & 0.329 & 0.426 & 0.712 & 0.756 & 0.682 & 0.734 & 0.257 & 0.329 & 0.192 & 0.217 \\
ESM-2 3B (mask) & 0.407 & 0.462 & 0.756 & 0.777 & 0.724 & 0.754 & 0.321 & 0.364 & 0.213 & 0.228 \\
ESM-2 3B (wt) & 0.406 & 0.455 & 0.759 & 0.776 & 0.725 & 0.751 & 0.324 & 0.358 & 0.213 & 0.221 \\
ESM-2 650M (mask) & 0.415 & 0.470 & 0.748 & 0.774 & 0.729 & 0.758 & 0.328 & 0.369 & 0.217 & 0.232 \\
ESM-2 650M (wt) & 0.418 & 0.468 & 0.751 & 0.775 & 0.732 & 0.758 & 0.334 & 0.366 & 0.214 & 0.226 \\
ESM-2 8M (mask) & 0.223 & 0.395 & 0.667 & 0.731 & 0.623 & 0.716 & 0.170 & 0.310 & 0.153 & 0.200 \\
ESM-2 8M (wt) & 0.251 & 0.387 & 0.685 & 0.748 & 0.641 & 0.713 & 0.196 & 0.297 & 0.167 & 0.204 \\
ESMC-300M & 0.409 & 0.460 & 0.751 & 0.766 & 0.728 & 0.754 & 0.322 & 0.363 & 0.213 & 0.222 \\
ESMC-600M & 0.407 & 0.464 & 0.750 & 0.770 & 0.726 & 0.755 & 0.321 & 0.364 & 0.204 & 0.221 \\
\bottomrule
\end{tabular}
\end{table}

\begin{table}[H]
\centering
\caption{Five-metric results for ProteinGym autoregressive language models before (Raw) and after (\textsc{\HarnessShortName{}}) applying the harness.}
\label{tab:pg_metrics_family_4}
\fontsize{9}{10.8}\selectfont
\setlength{\tabcolsep}{0pt}
\begin{tabular}{@{}p{0.40\textwidth}*{10}{>{\centering\arraybackslash}p{0.06\textwidth}}@{}}
\toprule
Model & \multicolumn{2}{c}{Spearman} & \multicolumn{2}{c}{NDCG} & \multicolumn{2}{c}{AUC} & \multicolumn{2}{c}{MCC} & \multicolumn{2}{c}{Top-recall} \\
\cmidrule(lr){2-3}\cmidrule(lr){4-5}\cmidrule(lr){6-7}\cmidrule(lr){8-9}\cmidrule(lr){10-11}
& Raw & \textsc{\HarnessShortName{}} & Raw & \textsc{\HarnessShortName{}} & Raw & \textsc{\HarnessShortName{}} & Raw & \textsc{\HarnessShortName{}} & Raw & \textsc{\HarnessShortName{}} \\
\midrule
ProGen2-B & 0.330 & 0.422 & 0.736 & 0.765 & 0.682 & 0.731 & 0.259 & 0.332 & 0.193 & 0.212 \\
ProGen2-L & 0.327 & 0.416 & 0.737 & 0.764 & 0.681 & 0.728 & 0.258 & 0.328 & 0.192 & 0.214 \\
ProGen2-M & 0.337 & 0.423 & 0.736 & 0.764 & 0.687 & 0.732 & 0.266 & 0.333 & 0.191 & 0.213 \\
ProGen2-S & 0.288 & 0.406 & 0.708 & 0.747 & 0.658 & 0.721 & 0.223 & 0.314 & 0.175 & 0.207 \\
ProGen2-XL & 0.351 & 0.419 & 0.755 & 0.772 & 0.694 & 0.730 & 0.275 & 0.329 & 0.198 & 0.207 \\
ProGen3-112M & 0.232 & 0.401 & 0.680 & 0.735 & 0.628 & 0.719 & 0.179 & 0.315 & 0.160 & 0.199 \\
ProGen3-1B & 0.320 & 0.406 & 0.732 & 0.757 & 0.677 & 0.722 & 0.251 & 0.316 & 0.191 & 0.210 \\
ProGen3-219M & 0.264 & 0.409 & 0.698 & 0.745 & 0.643 & 0.722 & 0.203 & 0.316 & 0.175 & 0.210 \\
ProGen3-339M & 0.286 & 0.414 & 0.708 & 0.748 & 0.656 & 0.726 & 0.222 & 0.322 & 0.176 & 0.208 \\
ProGen3-3B & 0.331 & 0.413 & 0.744 & 0.765 & 0.683 & 0.727 & 0.257 & 0.323 & 0.195 & 0.210 \\
ProGen3-762M & 0.304 & 0.413 & 0.719 & 0.753 & 0.667 & 0.726 & 0.237 & 0.321 & 0.181 & 0.209 \\
RITA-L & 0.309 & 0.407 & 0.727 & 0.757 & 0.672 & 0.725 & 0.241 & 0.321 & 0.180 & 0.204 \\
RITA-M & 0.297 & 0.405 & 0.714 & 0.751 & 0.664 & 0.722 & 0.230 & 0.315 & 0.177 & 0.200 \\
RITA-S & 0.254 & 0.398 & 0.696 & 0.743 & 0.641 & 0.718 & 0.203 & 0.314 & 0.170 & 0.199 \\
RITA-XL & 0.305 & 0.401 & 0.731 & 0.762 & 0.670 & 0.722 & 0.241 & 0.318 & 0.179 & 0.203 \\
\bottomrule
\end{tabular}
\end{table}

\paragraph{VenusMutHub: structural models.} Tables~\ref{tab:vmh_metrics_family_1} and~\ref{tab:vmh_metrics_family_3} compare the five metrics for structure-aware and inverse-folding models, respectively. These assay-macro results retain the VenusMutHub metric definitions; each comparison measures the change after applying the harness to the same configuration.

\begin{table}[H]
\centering
\caption{Five-metric results for VenusMutHub structure-aware models before (Raw) and after (\textsc{\HarnessShortName{}}) applying the harness.}
\label{tab:vmh_metrics_family_1}
\fontsize{9}{10.8}\selectfont
\setlength{\tabcolsep}{0pt}
\begin{tabular}{@{}p{0.40\textwidth}*{10}{>{\centering\arraybackslash}p{0.06\textwidth}}@{}}
\toprule
Model & \multicolumn{2}{c}{Spearman} & \multicolumn{2}{c}{NDCG} & \multicolumn{2}{c}{AUC} & \multicolumn{2}{c}{MCC} & \multicolumn{2}{c}{Top-recall} \\
\cmidrule(lr){2-3}\cmidrule(lr){4-5}\cmidrule(lr){6-7}\cmidrule(lr){8-9}\cmidrule(lr){10-11}
& Raw & \textsc{\HarnessShortName{}} & Raw & \textsc{\HarnessShortName{}} & Raw & \textsc{\HarnessShortName{}} & Raw & \textsc{\HarnessShortName{}} & Raw & \textsc{\HarnessShortName{}} \\
\midrule
ESM3 & 0.185 & 0.226 & 0.837 & 0.842 & 0.598 & 0.619 & 0.147 & 0.187 & 0.219 & 0.219 \\
ProSST (K=1024) & 0.162 & 0.242 & 0.822 & 0.845 & 0.585 & 0.627 & 0.131 & 0.187 & 0.201 & 0.254 \\
ProSST (K=128) & 0.139 & 0.230 & 0.817 & 0.840 & 0.575 & 0.620 & 0.103 & 0.176 & 0.181 & 0.234 \\
ProSST (K=20) & 0.143 & 0.238 & 0.820 & 0.841 & 0.576 & 0.627 & 0.114 & 0.189 & 0.191 & 0.225 \\
ProSST (K=2048) & 0.213 & 0.249 & 0.840 & 0.847 & 0.614 & 0.630 & 0.170 & 0.199 & 0.228 & 0.246 \\
ProSST (K=4096) & 0.161 & 0.237 & 0.823 & 0.842 & 0.585 & 0.623 & 0.120 & 0.188 & 0.193 & 0.234 \\
ProSST (K=512) & 0.124 & 0.226 & 0.813 & 0.840 & 0.564 & 0.617 & 0.089 & 0.164 & 0.170 & 0.224 \\
ProtSSN & 0.174 & 0.195 & 0.836 & 0.839 & 0.588 & 0.601 & 0.134 & 0.148 & 0.220 & 0.220 \\
ProtSSN (k=10, h=1280) & 0.166 & 0.192 & 0.834 & 0.838 & 0.584 & 0.598 & 0.126 & 0.141 & 0.221 & 0.224 \\
ProtSSN (k=10, h=512) & 0.169 & 0.197 & 0.835 & 0.839 & 0.585 & 0.600 & 0.129 & 0.145 & 0.217 & 0.222 \\
ProtSSN (k=10, h=768) & 0.152 & 0.178 & 0.830 & 0.835 & 0.579 & 0.593 & 0.115 & 0.133 & 0.207 & 0.212 \\
ProtSSN (k=20, h=1280) & 0.174 & 0.198 & 0.836 & 0.838 & 0.590 & 0.605 & 0.127 & 0.154 & 0.208 & 0.209 \\
ProtSSN (k=20, h=512) & 0.168 & 0.192 & 0.835 & 0.838 & 0.586 & 0.600 & 0.122 & 0.143 & 0.212 & 0.216 \\
ProtSSN (k=20, h=768) & 0.165 & 0.191 & 0.835 & 0.840 & 0.583 & 0.598 & 0.125 & 0.146 & 0.222 & 0.224 \\
ProtSSN (k=30, h=1280) & 0.162 & 0.183 & 0.832 & 0.836 & 0.582 & 0.594 & 0.126 & 0.142 & 0.215 & 0.216 \\
ProtSSN (k=30, h=512) & 0.169 & 0.191 & 0.835 & 0.838 & 0.585 & 0.599 & 0.131 & 0.152 & 0.210 & 0.216 \\
ProtSSN (k=30, h=768) & 0.160 & 0.185 & 0.833 & 0.838 & 0.581 & 0.595 & 0.113 & 0.134 & 0.199 & 0.210 \\
S3F (mask) & 0.211 & 0.234 & 0.839 & 0.843 & 0.606 & 0.621 & 0.156 & 0.178 & 0.230 & 0.221 \\
S3F (wt) & 0.193 & 0.224 & 0.837 & 0.841 & 0.596 & 0.614 & 0.129 & 0.163 & 0.226 & 0.235 \\
SaProt (35M\_AF2, mask) & 0.188 & 0.237 & 0.832 & 0.841 & 0.597 & 0.624 & 0.162 & 0.186 & 0.208 & 0.227 \\
SaProt (35M\_AF2, wt) & 0.142 & 0.179 & 0.827 & 0.834 & 0.576 & 0.596 & 0.116 & 0.140 & 0.209 & 0.215 \\
SaProt (650M\_PDB, mask) & 0.205 & 0.240 & 0.836 & 0.844 & 0.604 & 0.623 & 0.151 & 0.179 & 0.220 & 0.230 \\
SaProt (650M\_PDB, wt) & 0.182 & 0.215 & 0.836 & 0.840 & 0.594 & 0.610 & 0.127 & 0.156 & 0.224 & 0.218 \\
SaProt (650M AF2, mask) & 0.207 & 0.240 & 0.837 & 0.842 & 0.606 & 0.623 & 0.161 & 0.184 & 0.223 & 0.230 \\
SaProt (650M AF2, wt) & 0.178 & 0.202 & 0.832 & 0.838 & 0.591 & 0.604 & 0.138 & 0.153 & 0.222 & 0.228 \\
VenusREM2 & 0.190 & 0.258 & 0.830 & 0.846 & 0.601 & 0.636 & 0.143 & 0.198 & 0.211 & 0.250 \\
\bottomrule
\end{tabular}
\end{table}

\begin{table}[H]
\centering
\caption{Five-metric results for VenusMutHub inverse-folding models before (Raw) and after (\textsc{\HarnessShortName{}}) applying the harness.}
\label{tab:vmh_metrics_family_3}
\fontsize{9}{10.8}\selectfont
\setlength{\tabcolsep}{0pt}
\begin{tabular}{@{}p{0.40\textwidth}*{10}{>{\centering\arraybackslash}p{0.06\textwidth}}@{}}
\toprule
Model & \multicolumn{2}{c}{Spearman} & \multicolumn{2}{c}{NDCG} & \multicolumn{2}{c}{AUC} & \multicolumn{2}{c}{MCC} & \multicolumn{2}{c}{Top-recall} \\
\cmidrule(lr){2-3}\cmidrule(lr){4-5}\cmidrule(lr){6-7}\cmidrule(lr){8-9}\cmidrule(lr){10-11}
& Raw & \textsc{\HarnessShortName{}} & Raw & \textsc{\HarnessShortName{}} & Raw & \textsc{\HarnessShortName{}} & Raw & \textsc{\HarnessShortName{}} & Raw & \textsc{\HarnessShortName{}} \\
\midrule
ESM-IF1 & 0.216 & 0.267 & 0.833 & 0.846 & 0.613 & 0.640 & 0.164 & 0.206 & 0.203 & 0.235 \\
MIF-ST & 0.196 & 0.230 & 0.837 & 0.842 & 0.599 & 0.618 & 0.141 & 0.172 & 0.219 & 0.229 \\
ProteinMPNN (v\_48\_002) & 0.191 & 0.244 & 0.830 & 0.844 & 0.598 & 0.626 & 0.149 & 0.183 & 0.199 & 0.230 \\
ProteinMPNN (v\_48\_010) & 0.218 & 0.256 & 0.836 & 0.848 & 0.611 & 0.631 & 0.163 & 0.199 & 0.229 & 0.253 \\
ProteinMPNN (v\_48\_020) & 0.221 & 0.271 & 0.835 & 0.851 & 0.611 & 0.638 & 0.164 & 0.194 & 0.225 & 0.266 \\
ProteinMPNN (v\_48\_030) & 0.210 & 0.259 & 0.835 & 0.848 & 0.603 & 0.628 & 0.152 & 0.183 & 0.227 & 0.249 \\
ProteinMPNN-Soluble (v\_48\_002) & 0.198 & 0.250 & 0.829 & 0.845 & 0.599 & 0.626 & 0.141 & 0.186 & 0.202 & 0.238 \\
ProteinMPNN-Soluble (v\_48\_010) & 0.212 & 0.256 & 0.837 & 0.846 & 0.604 & 0.630 & 0.148 & 0.193 & 0.236 & 0.242 \\
ProteinMPNN-Soluble (v\_48\_020) & 0.216 & 0.265 & 0.837 & 0.849 & 0.609 & 0.634 & 0.162 & 0.199 & 0.235 & 0.253 \\
ProteinMPNN-Soluble (v\_48\_030) & 0.222 & 0.265 & 0.839 & 0.849 & 0.610 & 0.637 & 0.163 & 0.212 & 0.236 & 0.255 \\
\bottomrule
\end{tabular}
\end{table}

\paragraph{VenusMutHub: sequence models.} Tables~\ref{tab:vmh_metrics_family_2} and~\ref{tab:vmh_metrics_family_4} compare the five metrics for masked and autoregressive models, respectively. These assay-macro results retain the VenusMutHub metric definitions; each comparison measures the change after applying the harness to the same configuration.

\begin{table}[H]
\centering
\caption{Five-metric results for VenusMutHub masked protein LMs before (Raw) and after (\textsc{\HarnessShortName{}}) applying the harness.}
\label{tab:vmh_metrics_family_2}
\fontsize{9}{10.8}\selectfont
\setlength{\tabcolsep}{0pt}
\begin{tabular}{@{}p{0.40\textwidth}*{10}{>{\centering\arraybackslash}p{0.06\textwidth}}@{}}
\toprule
Model & \multicolumn{2}{c}{Spearman} & \multicolumn{2}{c}{NDCG} & \multicolumn{2}{c}{AUC} & \multicolumn{2}{c}{MCC} & \multicolumn{2}{c}{Top-recall} \\
\cmidrule(lr){2-3}\cmidrule(lr){4-5}\cmidrule(lr){6-7}\cmidrule(lr){8-9}\cmidrule(lr){10-11}
& Raw & \textsc{\HarnessShortName{}} & Raw & \textsc{\HarnessShortName{}} & Raw & \textsc{\HarnessShortName{}} & Raw & \textsc{\HarnessShortName{}} & Raw & \textsc{\HarnessShortName{}} \\
\midrule
CARP-38M & 0.073 & 0.158 & 0.816 & 0.831 & 0.543 & 0.587 & 0.057 & 0.120 & 0.195 & 0.206 \\
CARP-600K & 0.021 & 0.151 & 0.804 & 0.830 & 0.513 & 0.581 & 0.013 & 0.109 & 0.160 & 0.198 \\
CARP-640M & 0.157 & 0.206 & 0.830 & 0.838 & 0.584 & 0.609 & 0.128 & 0.152 & 0.192 & 0.212 \\
CARP-76M & 0.131 & 0.194 & 0.826 & 0.837 & 0.567 & 0.600 & 0.099 & 0.152 & 0.200 & 0.224 \\
ESM-1b (mask) & 0.166 & 0.211 & 0.831 & 0.838 & 0.589 & 0.612 & 0.137 & 0.162 & 0.208 & 0.215 \\
ESM-1b (wt) & 0.153 & 0.193 & 0.831 & 0.836 & 0.582 & 0.604 & 0.123 & 0.156 & 0.204 & 0.221 \\
ESM-1v (mask) & 0.162 & 0.206 & 0.830 & 0.838 & 0.581 & 0.606 & 0.124 & 0.165 & 0.198 & 0.215 \\
ESM-1v (wt) & 0.153 & 0.188 & 0.831 & 0.837 & 0.577 & 0.595 & 0.122 & 0.147 & 0.199 & 0.210 \\
ESM-2 150M (mask) & 0.167 & 0.225 & 0.831 & 0.840 & 0.588 & 0.617 & 0.136 & 0.179 & 0.197 & 0.221 \\
ESM-2 150M (wt) & 0.165 & 0.216 & 0.833 & 0.840 & 0.588 & 0.615 & 0.129 & 0.177 & 0.206 & 0.216 \\
ESM-2 35M (mask) & 0.122 & 0.212 & 0.821 & 0.840 & 0.565 & 0.612 & 0.093 & 0.165 & 0.174 & 0.230 \\
ESM-2 35M (wt) & 0.110 & 0.186 & 0.823 & 0.835 & 0.559 & 0.600 & 0.080 & 0.151 & 0.194 & 0.215 \\
ESM-2 3B (mask) & 0.182 & 0.223 & 0.832 & 0.838 & 0.590 & 0.613 & 0.147 & 0.174 & 0.208 & 0.218 \\
ESM-2 3B (wt) & 0.161 & 0.194 & 0.832 & 0.836 & 0.581 & 0.600 & 0.111 & 0.142 & 0.215 & 0.217 \\
ESM-2 650M (mask) & 0.169 & 0.213 & 0.832 & 0.839 & 0.588 & 0.610 & 0.130 & 0.166 & 0.213 & 0.221 \\
ESM-2 650M (wt) & 0.152 & 0.187 & 0.830 & 0.836 & 0.579 & 0.598 & 0.118 & 0.140 & 0.204 & 0.216 \\
ESM-2 8M (mask) & 0.079 & 0.198 & 0.807 & 0.834 & 0.539 & 0.602 & 0.067 & 0.162 & 0.141 & 0.207 \\
ESM-2 8M (wt) & 0.065 & 0.176 & 0.814 & 0.834 & 0.535 & 0.592 & 0.058 & 0.131 & 0.169 & 0.204 \\
ESMC-300M & 0.144 & 0.198 & 0.828 & 0.836 & 0.575 & 0.602 & 0.119 & 0.156 & 0.198 & 0.212 \\
ESMC-600M & 0.155 & 0.194 & 0.830 & 0.837 & 0.580 & 0.601 & 0.134 & 0.146 & 0.197 & 0.210 \\
\bottomrule
\end{tabular}
\end{table}

\begin{table}[H]
\centering
\caption{Five-metric results for VenusMutHub autoregressive language models before (Raw) and after (\textsc{\HarnessShortName{}}) applying the harness.}
\label{tab:vmh_metrics_family_4}
\fontsize{9}{10.8}\selectfont
\setlength{\tabcolsep}{0pt}
\begin{tabular}{@{}p{0.40\textwidth}*{10}{>{\centering\arraybackslash}p{0.06\textwidth}}@{}}
\toprule
Model & \multicolumn{2}{c}{Spearman} & \multicolumn{2}{c}{NDCG} & \multicolumn{2}{c}{AUC} & \multicolumn{2}{c}{MCC} & \multicolumn{2}{c}{Top-recall} \\
\cmidrule(lr){2-3}\cmidrule(lr){4-5}\cmidrule(lr){6-7}\cmidrule(lr){8-9}\cmidrule(lr){10-11}
& Raw & \textsc{\HarnessShortName{}} & Raw & \textsc{\HarnessShortName{}} & Raw & \textsc{\HarnessShortName{}} & Raw & \textsc{\HarnessShortName{}} & Raw & \textsc{\HarnessShortName{}} \\
\midrule
ProGen2-B & 0.124 & 0.194 & 0.823 & 0.837 & 0.564 & 0.598 & 0.101 & 0.159 & 0.184 & 0.210 \\
ProGen2-L & 0.120 & 0.188 & 0.823 & 0.835 & 0.560 & 0.598 & 0.091 & 0.147 & 0.187 & 0.210 \\
ProGen2-M & 0.115 & 0.182 & 0.825 & 0.835 & 0.558 & 0.595 & 0.078 & 0.133 & 0.189 & 0.206 \\
ProGen2-S & 0.091 & 0.192 & 0.815 & 0.838 & 0.546 & 0.600 & 0.069 & 0.157 & 0.174 & 0.223 \\
ProGen2-XL & 0.123 & 0.176 & 0.827 & 0.836 & 0.561 & 0.591 & 0.083 & 0.134 & 0.195 & 0.202 \\
ProGen3-112M & 0.072 & 0.177 & 0.812 & 0.833 & 0.537 & 0.594 & 0.045 & 0.137 & 0.169 & 0.209 \\
ProGen3-1B & 0.108 & 0.173 & 0.821 & 0.833 & 0.561 & 0.592 & 0.096 & 0.135 & 0.186 & 0.207 \\
ProGen3-219M & 0.092 & 0.190 & 0.816 & 0.835 & 0.548 & 0.600 & 0.073 & 0.149 & 0.183 & 0.212 \\
ProGen3-339M & 0.110 & 0.189 & 0.821 & 0.837 & 0.560 & 0.601 & 0.094 & 0.149 & 0.183 & 0.215 \\
ProGen3-3B & 0.110 & 0.170 & 0.823 & 0.835 & 0.553 & 0.588 & 0.064 & 0.120 & 0.200 & 0.208 \\
ProGen3-762M & 0.111 & 0.180 & 0.822 & 0.834 & 0.556 & 0.592 & 0.088 & 0.150 & 0.191 & 0.209 \\
RITA-L & 0.110 & 0.193 & 0.823 & 0.837 & 0.555 & 0.600 & 0.071 & 0.151 & 0.194 & 0.212 \\
RITA-M & 0.100 & 0.186 & 0.821 & 0.836 & 0.552 & 0.598 & 0.075 & 0.140 & 0.179 & 0.208 \\
RITA-S & 0.056 & 0.182 & 0.812 & 0.836 & 0.528 & 0.595 & 0.038 & 0.141 & 0.171 & 0.213 \\
RITA-XL & 0.095 & 0.160 & 0.820 & 0.831 & 0.547 & 0.585 & 0.065 & 0.130 & 0.182 & 0.208 \\
\bottomrule
\end{tabular}
\end{table}

\paragraph{\ViroBenchmarkName{}: structural models.} Tables~\ref{tab:viro_metrics_family_1} and~\ref{tab:viro_metrics_family_3} compare the five metrics for structure-aware and inverse-folding models, respectively. All scores exclude WT controls. AUC and MCC use 82 assays in 45 phenotype--protein groups, while the other metrics use all 89 assays in 52 groups.

\begin{table}[H]
\centering
\caption{\ViroBenchmarkName{} structure-aware models before (Raw) and after (\textsc{\HarnessShortName{}}) applying the harness. Metrics use mutant-only hierarchical means by phenotype and WT protein; AUC/MCC use 82 assays (45 groups), and the other metrics use 89 assays (52 groups).}
\label{tab:viro_metrics_family_1}
\fontsize{9}{10.8}\selectfont
\setlength{\tabcolsep}{0pt}
\begin{tabular}{@{}p{0.40\textwidth}*{10}{>{\centering\arraybackslash}p{0.06\textwidth}}@{}}
\toprule
Model & \multicolumn{2}{c}{Spearman} & \multicolumn{2}{c}{NDCG} & \multicolumn{2}{c}{AUC} & \multicolumn{2}{c}{MCC} & \multicolumn{2}{c}{Top-recall} \\
\cmidrule(lr){2-3}\cmidrule(lr){4-5}\cmidrule(lr){6-7}\cmidrule(lr){8-9}\cmidrule(lr){10-11}
& Raw & \textsc{\HarnessShortName{}} & Raw & \textsc{\HarnessShortName{}} & Raw & \textsc{\HarnessShortName{}} & Raw & \textsc{\HarnessShortName{}} & Raw & \textsc{\HarnessShortName{}} \\
\midrule
ESM3 & 0.137 & 0.249 & 0.690 & 0.733 & 0.573 & 0.632 & 0.103 & 0.200 & 0.124 & 0.152 \\
ProSST (K=1024) & 0.204 & 0.275 & 0.673 & 0.727 & 0.608 & 0.645 & 0.157 & 0.219 & 0.125 & 0.153 \\
ProSST (K=128) & 0.179 & 0.264 & 0.671 & 0.725 & 0.596 & 0.640 & 0.139 & 0.212 & 0.127 & 0.152 \\
ProSST (K=20) & 0.161 & 0.259 & 0.669 & 0.721 & 0.586 & 0.637 & 0.122 & 0.212 & 0.126 & 0.155 \\
ProSST (K=2048) & 0.271 & 0.310 & 0.691 & 0.730 & 0.643 & 0.664 & 0.223 & 0.252 & 0.135 & 0.157 \\
ProSST (K=4096) & 0.200 & 0.271 & 0.674 & 0.723 & 0.609 & 0.646 & 0.156 & 0.220 & 0.127 & 0.150 \\
ProSST (K=512) & 0.191 & 0.269 & 0.681 & 0.728 & 0.603 & 0.643 & 0.145 & 0.214 & 0.127 & 0.151 \\
ProtSSN & 0.233 & 0.294 & 0.716 & 0.737 & 0.622 & 0.657 & 0.181 & 0.235 & 0.150 & 0.159 \\
ProtSSN (k=10, h=1280) & 0.220 & 0.284 & 0.714 & 0.739 & 0.615 & 0.652 & 0.171 & 0.225 & 0.149 & 0.162 \\
ProtSSN (k=10, h=512) & 0.202 & 0.272 & 0.702 & 0.735 & 0.605 & 0.646 & 0.155 & 0.215 & 0.141 & 0.157 \\
ProtSSN (k=10, h=768) & 0.214 & 0.279 & 0.710 & 0.736 & 0.612 & 0.650 & 0.166 & 0.224 & 0.144 & 0.161 \\
ProtSSN (k=20, h=1280) & 0.230 & 0.289 & 0.713 & 0.736 & 0.620 & 0.654 & 0.179 & 0.232 & 0.147 & 0.158 \\
ProtSSN (k=20, h=512) & 0.233 & 0.291 & 0.714 & 0.736 & 0.622 & 0.655 & 0.179 & 0.233 & 0.145 & 0.156 \\
ProtSSN (k=20, h=768) & 0.226 & 0.287 & 0.712 & 0.737 & 0.619 & 0.654 & 0.175 & 0.230 & 0.150 & 0.161 \\
ProtSSN (k=30, h=1280) & 0.210 & 0.279 & 0.706 & 0.732 & 0.609 & 0.649 & 0.159 & 0.221 & 0.146 & 0.156 \\
ProtSSN (k=30, h=512) & 0.225 & 0.288 & 0.713 & 0.735 & 0.617 & 0.653 & 0.172 & 0.229 & 0.151 & 0.159 \\
ProtSSN (k=30, h=768) & 0.223 & 0.286 & 0.708 & 0.734 & 0.616 & 0.652 & 0.173 & 0.228 & 0.144 & 0.156 \\
S3F (mask) & 0.269 & 0.302 & 0.718 & 0.752 & 0.640 & 0.659 & 0.212 & 0.244 & 0.154 & 0.169 \\
S3F (wt) & 0.279 & 0.318 & 0.739 & 0.752 & 0.646 & 0.669 & 0.219 & 0.258 & 0.162 & 0.170 \\
SaProt (35M\_AF2, mask) & 0.149 & 0.245 & 0.660 & 0.720 & 0.579 & 0.630 & 0.112 & 0.198 & 0.133 & 0.154 \\
SaProt (35M\_AF2, wt) & 0.108 & 0.209 & 0.650 & 0.699 & 0.555 & 0.612 & 0.076 & 0.162 & 0.119 & 0.140 \\
SaProt (650M\_PDB, mask) & 0.306 & 0.320 & 0.727 & 0.753 & 0.662 & 0.670 & 0.245 & 0.259 & 0.164 & 0.171 \\
SaProt (650M\_PDB, wt) & 0.273 & 0.307 & 0.714 & 0.735 & 0.644 & 0.663 & 0.218 & 0.244 & 0.152 & 0.164 \\
SaProt (650M AF2, mask) & 0.222 & 0.281 & 0.687 & 0.739 & 0.618 & 0.649 & 0.174 & 0.226 & 0.150 & 0.168 \\
SaProt (650M AF2, wt) & 0.188 & 0.259 & 0.675 & 0.712 & 0.597 & 0.638 & 0.144 & 0.201 & 0.138 & 0.153 \\
VenusREM2 & 0.236 & 0.297 & 0.696 & 0.740 & 0.626 & 0.658 & 0.183 & 0.240 & 0.142 & 0.162 \\
\bottomrule
\end{tabular}
\end{table}

\begin{table}[H]
\centering
\caption{\ViroBenchmarkName{} inverse-folding models before (Raw) and after (\textsc{\HarnessShortName{}}) applying the harness. Metrics use mutant-only hierarchical means by phenotype and WT protein; AUC/MCC use 82 assays (45 groups), and the other metrics use 89 assays (52 groups).}
\label{tab:viro_metrics_family_3}
\fontsize{9}{10.8}\selectfont
\setlength{\tabcolsep}{0pt}
\begin{tabular}{@{}p{0.40\textwidth}*{10}{>{\centering\arraybackslash}p{0.06\textwidth}}@{}}
\toprule
Model & \multicolumn{2}{c}{Spearman} & \multicolumn{2}{c}{NDCG} & \multicolumn{2}{c}{AUC} & \multicolumn{2}{c}{MCC} & \multicolumn{2}{c}{Top-recall} \\
\cmidrule(lr){2-3}\cmidrule(lr){4-5}\cmidrule(lr){6-7}\cmidrule(lr){8-9}\cmidrule(lr){10-11}
& Raw & \textsc{\HarnessShortName{}} & Raw & \textsc{\HarnessShortName{}} & Raw & \textsc{\HarnessShortName{}} & Raw & \textsc{\HarnessShortName{}} & Raw & \textsc{\HarnessShortName{}} \\
\midrule
ESM-IF1 & 0.285 & 0.318 & 0.720 & 0.755 & 0.648 & 0.666 & 0.226 & 0.258 & 0.160 & 0.171 \\
MIF-ST & 0.288 & 0.317 & 0.736 & 0.752 & 0.653 & 0.669 & 0.234 & 0.259 & 0.166 & 0.173 \\
ProteinMPNN (v\_48\_002) & 0.269 & 0.305 & 0.729 & 0.754 & 0.647 & 0.663 & 0.229 & 0.252 & 0.170 & 0.180 \\
ProteinMPNN (v\_48\_010) & 0.271 & 0.306 & 0.728 & 0.753 & 0.647 & 0.664 & 0.225 & 0.253 & 0.169 & 0.178 \\
ProteinMPNN (v\_48\_020) & 0.270 & 0.306 & 0.722 & 0.749 & 0.647 & 0.663 & 0.225 & 0.255 & 0.164 & 0.175 \\
ProteinMPNN (v\_48\_030) & 0.266 & 0.304 & 0.718 & 0.747 & 0.644 & 0.662 & 0.221 & 0.251 & 0.161 & 0.173 \\
ProteinMPNN-Soluble (v\_48\_002) & 0.269 & 0.305 & 0.725 & 0.752 & 0.647 & 0.663 & 0.228 & 0.250 & 0.168 & 0.177 \\
ProteinMPNN-Soluble (v\_48\_010) & 0.268 & 0.304 & 0.726 & 0.754 & 0.646 & 0.662 & 0.225 & 0.251 & 0.169 & 0.177 \\
ProteinMPNN-Soluble (v\_48\_020) & 0.265 & 0.304 & 0.721 & 0.753 & 0.644 & 0.662 & 0.221 & 0.251 & 0.165 & 0.176 \\
ProteinMPNN-Soluble (v\_48\_030) & 0.256 & 0.299 & 0.714 & 0.749 & 0.640 & 0.660 & 0.215 & 0.247 & 0.163 & 0.176 \\
\bottomrule
\end{tabular}
\end{table}

\paragraph{\ViroBenchmarkName{}: sequence models.} Tables~\ref{tab:viro_metrics_family_2} and~\ref{tab:viro_metrics_family_4} compare the five metrics for masked and autoregressive models, respectively. All scores exclude WT controls. AUC and MCC use 82 assays in 45 phenotype--protein groups, while the other metrics use all 89 assays in 52 groups.

\begin{table}[H]
\centering
\caption{\ViroBenchmarkName{} masked protein LMs before (Raw) and after (\textsc{\HarnessShortName{}}) applying the harness. Metrics use mutant-only hierarchical means by phenotype and WT protein; AUC/MCC use 82 assays (45 groups), and the other metrics use 89 assays (52 groups).}
\label{tab:viro_metrics_family_2}
\fontsize{9}{10.8}\selectfont
\setlength{\tabcolsep}{0pt}
\begin{tabular}{@{}p{0.40\textwidth}*{10}{>{\centering\arraybackslash}p{0.06\textwidth}}@{}}
\toprule
Model & \multicolumn{2}{c}{Spearman} & \multicolumn{2}{c}{NDCG} & \multicolumn{2}{c}{AUC} & \multicolumn{2}{c}{MCC} & \multicolumn{2}{c}{Top-recall} \\
\cmidrule(lr){2-3}\cmidrule(lr){4-5}\cmidrule(lr){6-7}\cmidrule(lr){8-9}\cmidrule(lr){10-11}
& Raw & \textsc{\HarnessShortName{}} & Raw & \textsc{\HarnessShortName{}} & Raw & \textsc{\HarnessShortName{}} & Raw & \textsc{\HarnessShortName{}} & Raw & \textsc{\HarnessShortName{}} \\
\midrule
CARP-38M & 0.027 & 0.222 & 0.630 & 0.727 & 0.515 & 0.621 & 0.028 & 0.178 & 0.103 & 0.145 \\
CARP-600K & 0.006 & 0.210 & 0.614 & 0.722 & 0.504 & 0.615 & 0.013 & 0.170 & 0.099 & 0.143 \\
CARP-640M & 0.092 & 0.215 & 0.658 & 0.709 & 0.547 & 0.614 & 0.071 & 0.170 & 0.115 & 0.135 \\
CARP-76M & 0.050 & 0.230 & 0.637 & 0.728 & 0.527 & 0.625 & 0.040 & 0.181 & 0.109 & 0.151 \\
ESM-1b (mask) & 0.093 & 0.223 & 0.649 & 0.710 & 0.545 & 0.616 & 0.064 & 0.180 & 0.111 & 0.138 \\
ESM-1b (wt) & 0.094 & 0.248 & 0.648 & 0.730 & 0.545 & 0.632 & 0.064 & 0.192 & 0.115 & 0.154 \\
ESM-1v (mask) & 0.064 & 0.202 & 0.648 & 0.697 & 0.537 & 0.608 & 0.057 & 0.165 & 0.106 & 0.128 \\
ESM-1v (wt) & 0.069 & 0.238 & 0.651 & 0.735 & 0.539 & 0.630 & 0.061 & 0.191 & 0.106 & 0.151 \\
ESM-2 150M (mask) & 0.071 & 0.209 & 0.641 & 0.700 & 0.535 & 0.609 & 0.052 & 0.172 & 0.109 & 0.132 \\
ESM-2 150M (wt) & 0.069 & 0.235 & 0.639 & 0.720 & 0.533 & 0.626 & 0.050 & 0.185 & 0.109 & 0.145 \\
ESM-2 35M (mask) & 0.026 & 0.184 & 0.628 & 0.689 & 0.513 & 0.597 & 0.020 & 0.152 & 0.100 & 0.128 \\
ESM-2 35M (wt) & 0.026 & 0.213 & 0.624 & 0.720 & 0.514 & 0.616 & 0.023 & 0.168 & 0.099 & 0.144 \\
ESM-2 3B (mask) & 0.145 & 0.243 & 0.689 & 0.732 & 0.576 & 0.630 & 0.110 & 0.196 & 0.124 & 0.143 \\
ESM-2 3B (wt) & 0.138 & 0.260 & 0.684 & 0.736 & 0.572 & 0.639 & 0.103 & 0.203 & 0.122 & 0.153 \\
ESM-2 650M (mask) & 0.111 & 0.227 & 0.672 & 0.722 & 0.556 & 0.620 & 0.079 & 0.181 & 0.118 & 0.139 \\
ESM-2 650M (wt) & 0.114 & 0.252 & 0.673 & 0.734 & 0.557 & 0.634 & 0.082 & 0.196 & 0.119 & 0.153 \\
ESM-2 8M (mask) & 0.015 & 0.173 & 0.625 & 0.683 & 0.508 & 0.591 & 0.017 & 0.142 & 0.101 & 0.120 \\
ESM-2 8M (wt) & 0.017 & 0.214 & 0.625 & 0.724 & 0.509 & 0.616 & 0.017 & 0.170 & 0.100 & 0.144 \\
ESMC-300M & 0.097 & 0.220 & 0.668 & 0.712 & 0.550 & 0.615 & 0.073 & 0.177 & 0.126 & 0.141 \\
ESMC-600M & 0.108 & 0.228 & 0.666 & 0.717 & 0.557 & 0.621 & 0.085 & 0.187 & 0.131 & 0.141 \\
\bottomrule
\end{tabular}
\end{table}

\begin{table}[H]
\centering
\caption{\ViroBenchmarkName{} autoregressive language models before (Raw) and after (\textsc{\HarnessShortName{}}) applying the harness. Metrics use mutant-only hierarchical means by phenotype and WT protein; AUC/MCC use 82 assays (45 groups), and the other metrics use 89 assays (52 groups).}
\label{tab:viro_metrics_family_4}
\fontsize{9}{10.8}\selectfont
\setlength{\tabcolsep}{0pt}
\begin{tabular}{@{}p{0.40\textwidth}*{10}{>{\centering\arraybackslash}p{0.06\textwidth}}@{}}
\toprule
Model & \multicolumn{2}{c}{Spearman} & \multicolumn{2}{c}{NDCG} & \multicolumn{2}{c}{AUC} & \multicolumn{2}{c}{MCC} & \multicolumn{2}{c}{Top-recall} \\
\cmidrule(lr){2-3}\cmidrule(lr){4-5}\cmidrule(lr){6-7}\cmidrule(lr){8-9}\cmidrule(lr){10-11}
& Raw & \textsc{\HarnessShortName{}} & Raw & \textsc{\HarnessShortName{}} & Raw & \textsc{\HarnessShortName{}} & Raw & \textsc{\HarnessShortName{}} & Raw & \textsc{\HarnessShortName{}} \\
\midrule
ProGen2-B & 0.076 & 0.175 & 0.674 & 0.716 & 0.543 & 0.596 & 0.064 & 0.141 & 0.123 & 0.141 \\
ProGen2-L & 0.064 & 0.173 & 0.675 & 0.724 & 0.537 & 0.594 & 0.049 & 0.139 & 0.124 & 0.144 \\
ProGen2-M & 0.117 & 0.212 & 0.688 & 0.730 & 0.562 & 0.612 & 0.090 & 0.164 & 0.128 & 0.149 \\
ProGen2-S & 0.045 & 0.171 & 0.648 & 0.711 & 0.528 & 0.593 & 0.039 & 0.140 & 0.113 & 0.137 \\
ProGen2-XL & 0.162 & 0.235 & 0.715 & 0.744 & 0.588 & 0.627 & 0.128 & 0.190 & 0.146 & 0.153 \\
ProGen3-112M & 0.034 & 0.182 & 0.636 & 0.692 & 0.521 & 0.597 & 0.035 & 0.148 & 0.099 & 0.115 \\
ProGen3-1B & 0.092 & 0.207 & 0.667 & 0.712 & 0.549 & 0.609 & 0.077 & 0.162 & 0.121 & 0.136 \\
ProGen3-219M & 0.033 & 0.181 & 0.634 & 0.695 & 0.521 & 0.597 & 0.034 & 0.144 & 0.099 & 0.118 \\
ProGen3-339M & 0.058 & 0.191 & 0.647 & 0.698 & 0.533 & 0.601 & 0.056 & 0.153 & 0.105 & 0.123 \\
ProGen3-3B & 0.101 & 0.211 & 0.678 & 0.723 & 0.552 & 0.611 & 0.077 & 0.167 & 0.123 & 0.139 \\
ProGen3-762M & 0.062 & 0.192 & 0.652 & 0.705 & 0.534 & 0.602 & 0.052 & 0.151 & 0.106 & 0.127 \\
RITA-L & 0.166 & 0.247 & 0.733 & 0.751 & 0.586 & 0.630 & 0.124 & 0.191 & 0.142 & 0.154 \\
RITA-M & 0.155 & 0.237 & 0.726 & 0.749 & 0.581 & 0.626 & 0.114 & 0.183 & 0.143 & 0.156 \\
RITA-S & 0.148 & 0.236 & 0.721 & 0.748 & 0.579 & 0.625 & 0.113 & 0.186 & 0.138 & 0.156 \\
RITA-XL & 0.172 & 0.247 & 0.736 & 0.754 & 0.590 & 0.631 & 0.130 & 0.194 & 0.146 & 0.156 \\
\bottomrule
\end{tabular}
\end{table}

\endgroup

\FloatBarrier

\clearpage
\section{Usage of \HarnessName{}}
\label{app:usage}

\begin{figure}[H]
\centering
\includegraphics[width=0.86\textwidth]{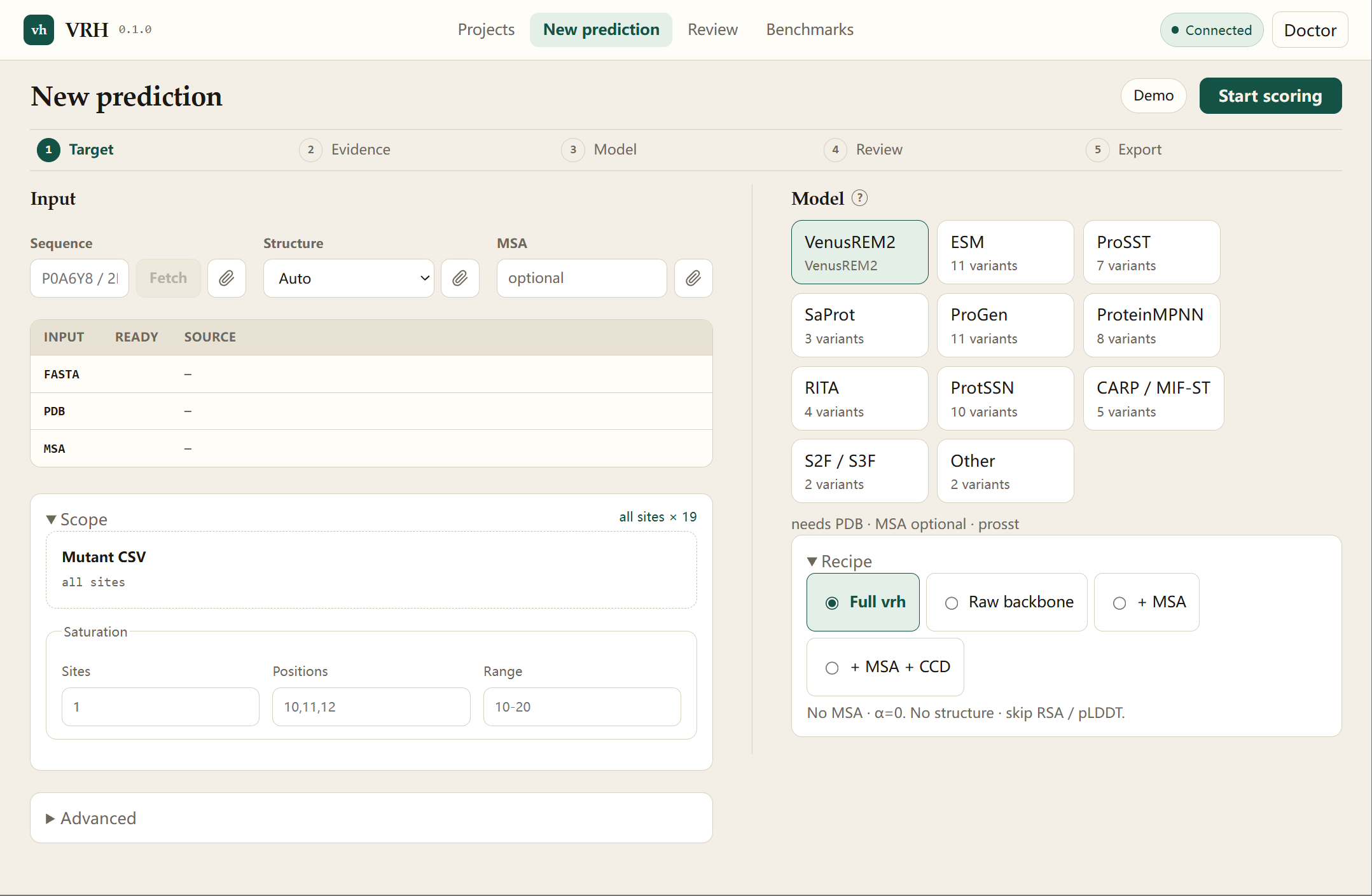}
\caption{\textbf{New prediction.} The README dashboard preview places sequence, structure, and MSA inputs on the left and model selection on the right. The input checklist and \textbf{Start scoring} button guide submission of a prediction.}
\label{fig:usage_predict}
\end{figure}

\begin{figure}[H]
\centering
\includegraphics[width=0.78\textwidth]{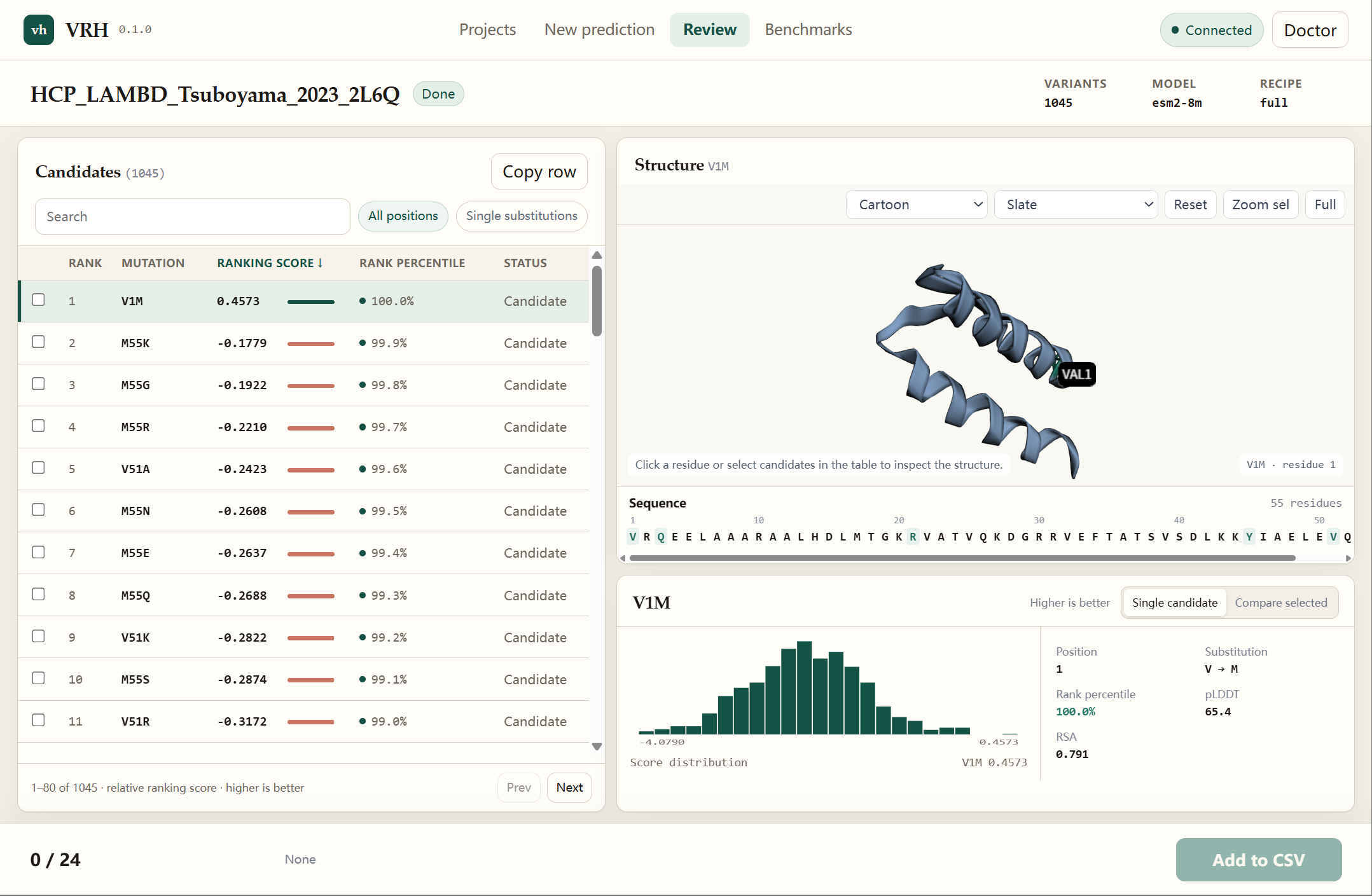}
\caption{\textbf{Review and export.} The ranked candidate table is linked to the structure viewer and information for each residue. Selected rows can be exported using \textbf{Add to CSV}. Screenshot reproduced from the README preview.}
\label{fig:usage_review}
\end{figure}

\begin{figure}[H]
\centering
\includegraphics[width=0.78\textwidth]{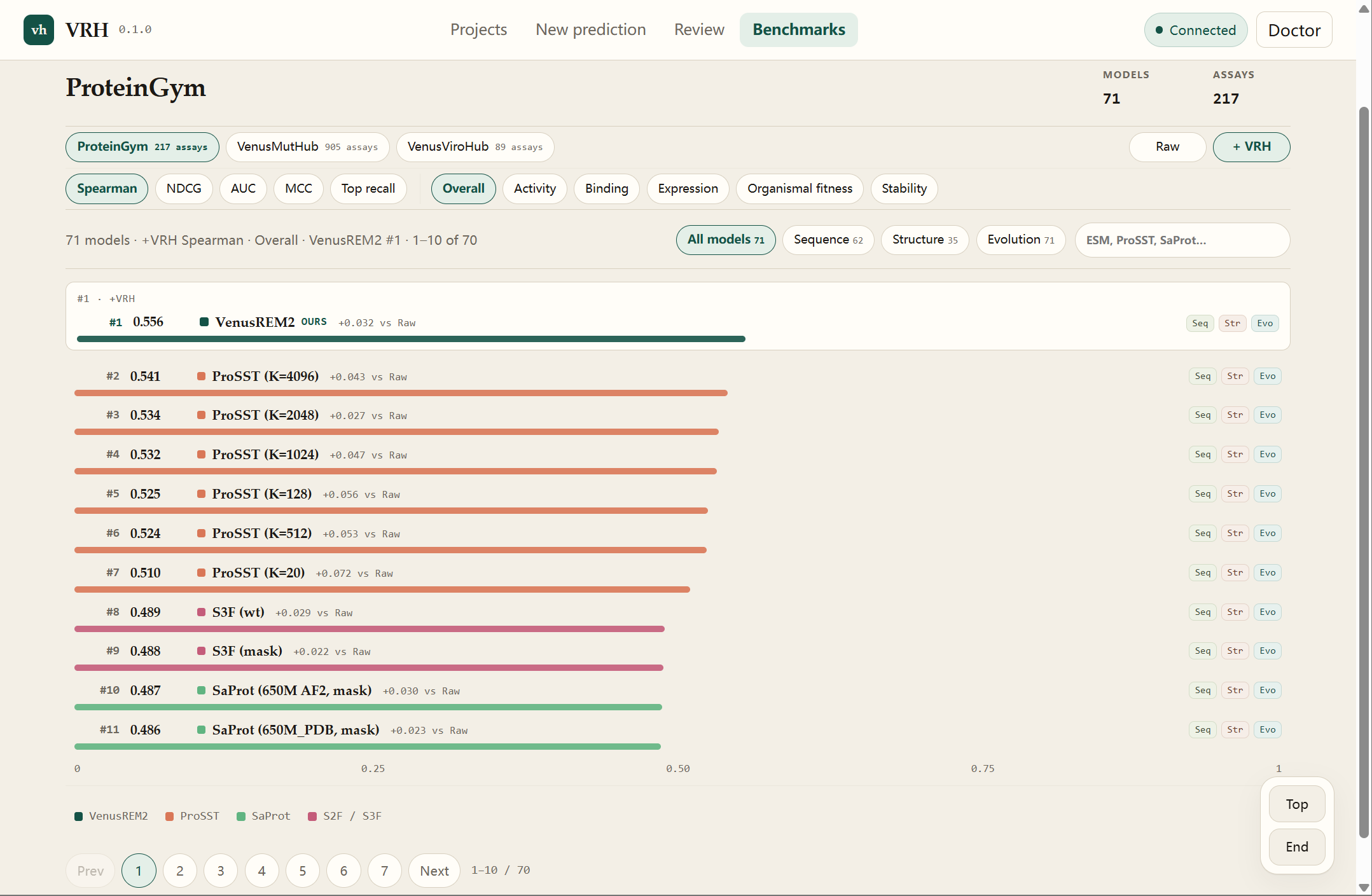}
\caption{\textbf{Benchmark browsing.} Dataset, metric, property, and input-type filters support comparison of original PFM predictions with scores after applying the harness. The README screenshot shows the 71-configuration ProteinGym comparison.}
\label{fig:usage_benchmark}
\end{figure}

The \texttt{vrh} package provides \textsc{\HarnessName{}} through a command-line interface and a local browser-based dashboard. Users prepare inputs, choose a model, inspect ranked predictions, and export selected variants. Figures~\ref{fig:usage_predict}--\ref{fig:usage_benchmark} reproduce the three README screenshots of the dashboard preview.

\paragraph{Installation and launch.}
In an environment with a compatible PyTorch installation, run the following commands from the repository root:
\begin{verbatim}
pip install -e .
vrh doctor
vrh dashboard
\end{verbatim}
Run the command-line interface with \texttt{vrh}; the \texttt{prosst} extra supplies dependencies for \textsc{VenusREM2}. The dependency check reports the installed model dependencies and cache status. Install any optional dependencies required by the chosen model; the README lists these extras. The dashboard opens locally at \url{http://127.0.0.1:8765}. The optional \texttt{vrh demo} command tests the installation on the bundled example.

\paragraph{1. Prepare inputs and start a prediction.}
Open \textbf{New prediction} (Figure~\ref{fig:usage_predict}). Upload the sequence and any required structure, or enter a UniProt/PDB identifier and use \textbf{Fetch}. Supply an MSA when available. The input checklist shows which inputs are available. Select the model family and configuration, review the variants to score and the scoring settings, and click \textbf{Start scoring}. The \textbf{Demo} button fills the form with the bundled example to try the workflow.

\clearpage
\paragraph{2. Inspect predictions and export a shortlist.}
Open \textbf{Review} after scoring (Figure~\ref{fig:usage_review}). Search or filter the ranked table and select a variant to inspect its position in the structure, its rank, and the available information for that residue. Tick the desired rows and use \textbf{Add to CSV} to export the shortlist. Scores express relative model preference, not $\Delta\Delta G$. Experimental PDB B-factors are not pLDDT; confidence coloring requires a predicted structure with confidence values.

\paragraph{3. Browse benchmark comparisons.}
Open \textbf{Benchmarks} (Figure~\ref{fig:usage_benchmark}), choose a dataset, and filter by metric, functional property, or input type. Switch between \textbf{Raw} (original PFM predictions) and the harness-adjusted view to compare results. The available datasets and model entries depend on the benchmark tables installed with the dashboard.

\clearpage
\section{\reviewchange{Comparison with published PoET-2 results}}
\label{app:poet2_comparison}

\reviewchange{Table~\ref{tab:poet2_comparison} extends Figure~\ref{fig:leaderboard} with \textsc{PoET-2} and \textsc{PoET-2} + \textsc{VenusREM} \citep{truong2025poet2}. Both studies evaluate ProteinGym DMS substitutions using Spearman correlation and report the same 17 categories. Their ensemble averages scores from two different model architectures, with weight $0.5$ selected from 21 candidate weights on eight ProteinNPT validation datasets. \textsc{VenusREM2} combines six \textsc{ProSST} checkpoints as described in Section~\ref{sec:venusrem2}.}

\reviewchange{We retain each source's reported scores. Shared baselines differ slightly: \textsc{VenusREM} has overall Spearman $0.519$ in their paper and $0.518$ in the official summary used for Figure~\ref{fig:leaderboard}. Thus, the task and metric match, but the table compares published values rather than results recomputed with an identical evaluation pipeline.}

\begin{table}[!htbp]
\centering
\caption{\reviewchange{ProteinGym Spearman comparison with the published \textsc{PoET-2} results. $\Delta$ is \textsc{VenusREM2} minus \textsc{PoET-2} + \textsc{VenusREM}. The last column gives the \textsc{VenusREM2} rank after adding both published models to the Figure~\ref{fig:leaderboard} comparison. Bold marks the largest score in each row.}}
\label{tab:poet2_comparison}
\begingroup
\ifreviewchanges\color{red}\fi
\small
\setlength{\tabcolsep}{5pt}
\renewcommand{\arraystretch}{1.06}
\begin{tabular*}{\linewidth}{@{}l@{\extracolsep{\fill}}rrrrc@{}}
\toprule
Category & \textsc{VenusREM2} & \textsc{PoET-2} & \shortstack{\textsc{PoET-2} +\\\textsc{VenusREM}} & $\Delta$ & \shortstack{\textsc{VenusREM2}\\rank} \\
\midrule
Overall & \textbf{0.556} & 0.500 & 0.543 & $+0.013$ & 1 \\
\midrule
\multicolumn{6}{@{}l}{\textit{Function}} \\
Activity & \textbf{0.541} & 0.508 & 0.538 & $+0.003$ & 1 \\
Binding & \textbf{0.495} & 0.423 & 0.475 & $+0.020$ & 1 \\
Expression & \textbf{0.557} & 0.503 & 0.552 & $+0.005$ & 1 \\
Organismal fitness & 0.494 & 0.482 & \textbf{0.505} & $-0.011$ & 2 \\
Stability & \textbf{0.691} & 0.582 & 0.644 & $+0.047$ & 1 \\
\midrule
\multicolumn{6}{@{}l}{\textit{Taxon}} \\
Human & \textbf{0.562} & 0.506 & 0.548 & $+0.014$ & 1 \\
Other eukaryotes & \textbf{0.615} & 0.569 & 0.604 & $+0.011$ & 1 \\
Prokaryotes & \textbf{0.601} & 0.507 & 0.562 & $+0.039$ & 1 \\
Virus & 0.540 & 0.528 & \textbf{0.551} & $-0.011$ & 2 \\
\midrule
\multicolumn{6}{@{}l}{\textit{MSA depth}} \\
Low & \textbf{0.553} & 0.488 & 0.528 & $+0.025$ & 1 \\
Medium & \textbf{0.557} & 0.507 & 0.550 & $+0.007$ & 1 \\
High & \textbf{0.623} & 0.555 & 0.593 & $+0.030$ & 1 \\
\midrule
\multicolumn{6}{@{}l}{\textit{Mutation depth}} \\
1 & \textbf{0.576} & 0.506 & 0.556 & $+0.020$ & 1 \\
2 & \textbf{0.473} & 0.357 & 0.402 & $+0.071$ & 1 \\
3 & \textbf{0.468} & 0.444 & 0.442 & $+0.026$ & 1 \\
4 & \textbf{0.424} & 0.419 & 0.411 & $+0.013$ & 1 \\
$5+$ & \textbf{0.473} & 0.447 & 0.441 & $+0.032$ & 1 \\
\bottomrule
\end{tabular*}
\par\smallskip
\begin{minipage}{\linewidth}
\footnotesize
\textit{Sources.} \textsc{VenusREM2}: Tables~\ref{tab:leaderboard}, \ref{tab:leaderboard_full_taxon_msa}, and~\ref{tab:leaderboard_full_mutation_depth}. \textsc{PoET-2} and its ensemble: Tables~7 and~10--13 of \citet{truong2025poet2}. We consistently use their appendix values. Their main Table~2 instead gives $0.558/0.400$ for the ensemble at mutation depths $1/2$ (appendix: $0.556/0.402$), which leaves these ranks unchanged.
\end{minipage}
\endgroup
\end{table}

\reviewchange{Including these published results, \textsc{VenusREM2} remains first overall ($0.556$ versus $0.543$) and in 15 of the 17 categories. \textsc{PoET-2} + \textsc{VenusREM} leads in organismal fitness ($0.505$ versus $0.494$) and virus ($0.551$ versus $0.540$), placing \textsc{VenusREM2} second in those two categories. Its lead across all 17 categories in Figure~\ref{fig:leaderboard} therefore refers specifically to the official leaderboard baselines shown there.}
\FloatBarrier

\end{document}